%% file: main.tex
\documentclass{article}

\usepackage{amsthm}
\makeatletter
\newtheorem*{rep@theorem}{\rep@title}
\newcommand{\newreptheorem}[2]{%
\newenvironment{rep#1}[1]{%
 \def\rep@title{#2 \ref{##1}}%
 \begin{rep@theorem}}%
 {\end{rep@theorem}}}
\makeatother
\newreptheorem{theorem}{Theorem}
\newreptheorem{corollary}{Corollary}
\newreptheorem{lemma}{Lemma}

\usepackage{microtype}
\usepackage{graphicx}
\usepackage{subfigure}
\usepackage{booktabs} 

\usepackage{hyperref}

\usepackage[accepted]{icml2023}
\usepackage{amsmath}
\usepackage{amssymb}
\usepackage{mathtools}
\usepackage{amsfonts}
\usepackage{bm}
\usepackage{bbm}
\usepackage{thmtools,thm-restate}
\usepackage{enumitem}
\usepackage{pifont}
\usepackage{array}
\usepackage{float}
\usepackage{color}
\usepackage{caption}
\usepackage{multirow}
\usepackage{booktabs}
\usepackage{colortbl}
\usepackage{multicol}

\newcommand{\R}{\mathbb R}
\newcommand{\tr}[1]{\operatorname{tr}\left(#1\right)}
\newcommand{\diag}{\mathbf{diag}}
\newcommand{\rel}{\operatorname{Rel}}
\newcommand{\red}[1]{{\color{red}#1}}
\newcommand{\nei}{\mc N}
\newcommand{\mb}{\mathbf}

\newcommand{\mc}{\mathcal}
\newcommand{\nnz}{\operatorname{nnz}}
\newcommand{\vol}[1]{\operatorname{vol}\left(#1\right)}
\newcommand{\regret}{\operatorname{Reg}}
\newcommand{\supp}[1]{\operatorname{supp}\left(#1\right)}
\newcommand{\eps}{\epsilon}

\usepackage[capitalize,noabbrev]{cleveref}

\theoremstyle{plain}
\newtheorem{theorem}{Theorem}[section]

\newtheorem{lemma}[theorem]{Lemma}
\newtheorem{corollary}[theorem]{Corollary}
\theoremstyle{definition}
\newtheorem{definition}[theorem]{Definition}

\theoremstyle{remark}
\newtheorem{remark}[theorem]{Remark}

\usepackage[textsize=tiny]{todonotes}

\icmltitlerunning{Fast Online Node Labeling for Very Large Graphs}

\begin{document}

\twocolumn[
\icmltitle{Fast Online Node Labeling for Very Large Graphs}




\begin{icmlauthorlist}
\icmlauthor{Baojian Zhou}{fudan,datalab}
\icmlauthor{Yifan Sun}{stonybrook}
\icmlauthor{Reza Babanezhad}{sait}
\end{icmlauthorlist}

\icmlaffiliation{fudan}{School of Data Science, Fudan University, Shanghai, China}
\icmlaffiliation{datalab}{the Shanghai Key Laboratory of Data Science, Fudan University}
\icmlaffiliation{stonybrook}{Department of Computer Science, Stony Brook University, Stony Brook, USA}
\icmlaffiliation{sait}{Samsung-SAIT AI lab, Montreal, Canada}

\icmlcorrespondingauthor{Baojian Zhou}{bjzhou@fudan.edu.cn}

\icmlkeywords{Online Node Labeling, Local Push, Approximate Graph Kernel}

\vskip 0.3in
]

\printAffiliationsAndNotice{} 

\begin{abstract}

This paper studies the online node classification problem under a transductive learning setting. Current methods either invert a graph kernel matrix with $\mathcal{O}(n^3)$ runtime and $\mathcal{O}(n^2)$ space complexity or sample a large volume of random spanning trees, thus are difficult to scale to large graphs. In this work, we propose an improvement based on the \textit{online relaxation} technique introduced by a series of works \citep{rakhlin2012relax,rakhlin2015hierarchies,rakhlin2017efficient}. We first prove an effective regret $\mathcal{O}(\sqrt{n^{1+\gamma}})$ when suitable parameterized graph kernels are chosen,  then propose an approximate algorithm \textsc{FastONL} enjoying $\mathcal{O}(k\sqrt{n^{1+\gamma}})$ regret based on this relaxation. The key of \textsc{FastONL} is a \textit{generalized local push} method that effectively approximates inverse matrix columns and applies to a series of popular kernels. Furthermore, the per-prediction cost  is $\mathcal{O}(\vol{\mathcal{S}}\log 1/\eps)$ locally dependent on the graph with linear memory cost. Experiments show that our scalable method enjoys a better tradeoff between local and global consistency. 
\end{abstract}

\section{Introduction}
\label{section:introduction}
This paper explores the online node labeling problem within a transductive learning framework. Specifically, we consider the scenario where multi-category node labels $\bm y_t$ enter online, and our goal is to predict future labels $\bm y_{t+1}$ under constraints imposed by an underlying graph. To illustrate this, consider the example of online product recommendations on platforms such as Amazon. At each time step $t$, the task is to recommend one of $k$ products to a user (node) based on their relationships (edges) with other users. The success of these recommendations is gauged by whether the user proceeds to purchase the recommended product. In this context, leveraging local information, such as recommendations from friends, becomes a viable strategy. However, this presents a significant challenge as we need to generate per-iteration predictions within microseconds while managing large-scale graphs comprising millions or even billions of nodes. Due to their computational complexity, traditional methods that require solving linear systems of order $\mc O\left(n^3\right)$ are ill-suited for this task where $n$ is the number of nodes. This issue is not only critical in web-spam classification \cite{herbster2008fast}, product recommendation \cite{ying2018graph}, and community detection \cite{leung2009towards}, but also permeates many other graph-based applications.

Previous studies in this area have fallen into two categories. The first involves sampling the spanning tree from the graph and then predicting labels via a weighted majority-like method 
\cite{herbster2008fast, cesa2009fast,cesa2013random}. 
 Such methods require repeated spanning tree samplings and often suffer a large variance issue. The second family of methods designs a loss function and builds feature vectors from different graph kernels \cite{herbster2005online,herbster2006prediction,herbster2009predicting,herbster2015predicting, gentile2013online,rakhlin2015hierarchies,rakhlin2016tutorial,rakhlin2017efficient}. These kernel-based methods can successfully capture the label smoothness of graphs but need to invert the graph kernel matrices, severely limiting their scalability.

Several classical and modern graph representation learning works \cite{zhu2003semi,blum2004semi,kipf2017semisupervised} suggest that graph kernel-based methods are more effective in real-world applications. 
We focus on a significant line of work based on \textit{online relaxation} \cite{rakhlin2015hierarchies,rakhlin2016bistro,rakhlin2016tutorial,rakhlin2017efficient}, which achieves linear time per-iteration cost. However, two challenges remain. First, the choice of the used graph kernel matrix affects performance, and it is unclear how to make this choice optimally and obtain an effective regret. Second, the previous relaxation method \cite{rakhlin2017efficient} assumes that the inverse of the graph kernel matrix is readily available, which is not reasonable in practice. Note that a vanilla approach to computing a matrix inverse of a graph kernel involves $\mathcal{O}(n^3)$ computational cost and $\mathcal{O}(n^2)$ space complexity. Approximate matrix inverse techniques are required, but the regret guarantee for utilizing such schemes does not yet exist. The key question, then, is whether there exists an online node labeling method which \emph{accounts for the kernel matrix inversion}, where the per-iteration cost is independent of the whole graph and the overall method is nearly-linear time. 

In this paper, we propose such a solution by extending the online relaxation method \cite{rakhlin2015hierarchies,rakhlin2016tutorial, rakhlin2017efficient,rakhlin2016bistro} via a  fast local matrix inverse approximation method. Specifically, the inversion technique is based on the Approximate PageRank (\textsc{APPR}) method \cite{andersen2006local}, which is particularly effective and efficient when the magnitudes of these kernel vectors follow a power-law distribution, often found in real-world graphs. Our proposed Fast Online Node Labeling algorithm \textsc{FastONL}  approximates the kernel matrix inverse via variants of \textsc{APPR}. Moreover, we compute an effective regret bound of $\mc{O}(k \sqrt{n^{1+\gamma}})$, which accounts for the matrix inversion steps. While we focus on static graphs, the method can naturally be extended to the dynamic graph setting. 

\textbf{Our contributions.\quad} 

\begin{itemize}
\item For the first time, we show that online relaxation-based methods with suitable graph kernel parametrization enjoy an effective regret when the graph is highly structured; specifically, the regret can be bounded by $\mathcal{O}(\sqrt{n^{1+\gamma}})$ if the graph Laplacian is regularized by $\mathcal{O}(n^\gamma)$ for some $\gamma \in (0,1)$. This is generalized to several parameterized graph kernels.

\item To overcome the $\mathcal{O}(n^3)$ time and $\mathcal{O}(n^2)$ space complexity of the large matrix inversion barrier, we consider the  \textsc{APPR} approach, which gives a per-iteration  cost of $\mc{O}(\vol{\mc S}\log(1/\epsilon))$. 
This locally linear bound is exponentially superior to the previous $\mathcal{O}(1/\epsilon)$ bound for general graphs \cite{andersen2006local}.
\item On graphs between 1000 and 1M nodes, \textsc{FastONL} shows a better empirical tradeoff between local and global consistency. For a case study on the English Wikipedia graph with 6.2M nodes and 178M edges, we obtain a low error rate with a per-prediction run time of less than a second. 
\end{itemize}
Our code and datasets have been provided as supplementary material and is publicly available at \url{https://github.com/baojian/FastONL}. All proofs have been postponed to the appendix.

\section{Related Work}

\paragraph{Online node labeling.\quad} Even binary labeling of graph nodes  in the online learning setting can be challenging. A series of works on online learning over graphs is considered \cite{herbster2005online,herbster2008exploiting,herbster2008fast,herbster2009predicting,herbster2019online,herbster2021gang}. Initially, \citet{herbster2005online} considered learning graph node labels
 using a perceptron-based algorithm, which iteratively projected a sequence of points over a closed convex set. This initial method already requires finding the pseudoinverse of the unnormalized Laplacian matrix. Moreover, the total mistakes is bounded by $4 \Phi_{\mathcal{G}}(\bm y) D_{\mathcal{G}} \operatorname{bal}(\bm y)$ where $\Phi_\mathcal{G}$ is the graph cut, $D_\mathcal{G}$ is the diameter, and $\operatorname{bal}(\bm y)$ is the label balance ratio. This mistake bound, which is distinct from the regret bound in this paper, vanishes when the label is imbalanced. Subsequent works, such as \textsc{PUNCE} \cite{herbster2008exploiting} and \textsc{Seminorm} \cite{herbster2009predicting}, also admitted mistake bounds.  To remedy this issue, following works \cite{herbster2006prediction,herbster2008online,herbster2008exploiting} proposed different methods to avoid these large bounds. However, to the best of our knowledge, their effectiveness has not been validated on large-scale graphs. Additionally, it is unclear whether these methods can be effective under multi-category label settings.

The algorithms proposed in \citet{herbster2008online,herbster2009predicting,cesa2009fast,vitale2011see,cesa2013random} accelerate per-prediction by  working on trees and paths of the graph; see also \cite{gentile2013online} for evolving graphs. However, the total time complexity of the proposed method is quadratic w.r.t the graph size. Additionally, \citet{herbster2015predicting} considered the setting of predicting a switching sequence over multiple graphs, and \citet{gu2014online} explored an online spectral learning framework. All these works fundamentally depend on the inverse of the graph Laplacian.

More generally, the problem of transductive learning on graphs has been extensively studied over past years \cite{ng2001spectral,zhou2003learning,zhu2003semi,ando2006learning,johnson2007effectiveness,el2016asymptotic,kipf2017semisupervised}. Under batch transductive learning setting, the basic assumption is that nodes with same labels are well-clustered together. In this case, the quadratic form of the graph Laplacian kernel \eqref{equ:quadform} or even $p$-Laplacian-based \citep{el2016asymptotic,fu2022p} should be small. However, different from batch settings, this paper considers online learning settings based on kernel computations.

\textbf{Personalized PageRank and approximation.\quad} Personalized PageRank (PPR) as an important graph learning tool has been used in classic graph applications \cite{jeh2003scaling,andersen2008robust} and modern graph neural networks \cite{gasteiger2018combining,bojchevski2020scaling,epasto2022differentially} due its scalable approach to matrix inversion. The \textit{local push} method has been proposed in a seminal work of \citet{andersen2006local} as an efficient and localized approach toward computing PPR vectors; it was later shown to be a variant of coordinate descent \cite{fountoulakis2019variational}, and related to Gauss-Seidel iteration \cite{sun2021worst}. This paper introduces a new variant of the \textit{local push} to approximate many other graph kernel inverses.

\section{Preliminaries}
\label{sec:preliminaries}

This section introduces notation and  problem setup and presents the online relaxation method with surrogate loss.

\subsection{Notations and problem formulation}

\textbf{Notations.\quad} We consider an undirected weighted graph $\mc{G}=\left(\mc{V},\mc{E},\bm W \right)$ where $\mc{V} \triangleq \{1,2,\ldots,n\}$ is the set of nodes, $\mc{E}\subseteq \mc{V} \times \mc{V}$ is the set of $m = |\mc{E}|$ edges, and $\bm W\in {\mathbb R}_+^{n\times n}$ is the nonnegative weighted adjacency matrix where each edge $(u,v)\in \mc{E}$ has weight $W_{u v} > 0$. The unnormalized and normalized graph Laplacian is defined as $\mathcal{{\bm L}} \triangleq \bm D - \bm W$ and $\bm L \triangleq {\bm D}^{-1/2} \mathcal{{\bm L}} {\bm D}^{-1/2}$, respectively.\footnote{ When $\mathcal{G}$ contains singletons, ${\bm D}^{-1/2}=\left( \bm D^{+} \right)^{1/2}$ where $\bm D^+$ is the Moore-Penrose inverse of $\bm D$. } The set of neighbors of $u$ is denoted as $\nei{(u)} \triangleq \{v:(u,v)\in\mc{E}\}$ and the degree $d_u = \left|\nei{(u)}\right|$. 
 The weighted degree matrix is defined as a diagonal matrix $\bm D$ where $D_{u u} = \sum_{v \in \nei(u)} W_{u v}$.\footnote{Note that $d_u = D_u$ only when $\mc{G}$ is unweighted but $d_u \ne D_{u u}$ for a weighted graph in general.} Following the work of \citet{chung1997spectral}, for $\mc{S} \subseteq \mc{V}$, the volume of $\mc{S}$ is defined as $\vol{\mc S} \triangleq \sum_{v \in \mc S} d_v$. 

Given $k$ labels, each node $v$ has a label $y_v \in \{1,2,\ldots,k,\}$. For convenience, we use the binary form ${\bm y}_t \in \{{\bm e}_1, {\bm e}_2, \ldots, {\bm e}_k\}=:\mathcal Y$ where $\bm e_i$ is the one-hot encoding vector. $\bm X_{:,i} \in \R^n$ is the $i$-th column vector of matrix $\bm X \in {\R}^{n\times n}$ and $\bm X_{i,:} \in \R ^n$ is the transpose of $i$-th row vector of $\bm X$. The support of $\bm x \in \R^n$ is $\supp{\bm x} \triangleq \{v: x_v \ne 0, v \in \mc V\}$. The trace of a square matrix $\bm M$ is defined as $\tr{\bm M} = \sum_{i=1}^n m_{ii}$ where $m_{i i}$ is the $i$-th diagonal. For a symmetric matrix $\bm M$, denote $\lambda(\bm M)$ as the eigenvalue function of $\bm M$.

\begin{table*}[htbp]
\caption{The parameterized graph kernel matrices with their basic kernel presentation}
\label{tab:graph-kernel-presentation}
\centering
 \begin{tabular}{c | c | c | c | c} 
 \toprule
 ID & $\bm K_{\beta}^{-1}$ & $\alpha$ & Basic Kernel Presentation & Paper \\ [0.1ex] 
 \midrule
1 & $\mc{L}$ & $\frac{\lambda}{n}$ & $\bm M_{\lambda,\beta} = 2\lambda \bm X_{\mc L}$ & \cite{rakhlin2017efficient} \\  [.3ex] 
2 & ${\bm L}$ & $\frac{\lambda}{n+\lambda}$ & $\bm M_{\lambda,\beta} = 
 2 n {\bm D}^{-1/2}\bm X_{\bm L} {\bm D}^{1/2}$ & \cite{rakhlin2017efficient} \\ [.3ex] 
3 & $\bm I - \beta \bm D^{-1/2} \bm W \bm D^{-1/2}$ & $\frac{n+\lambda -\beta n}{n+\lambda}$ & $\bm M_{\lambda,\beta} =  \frac{2\lambda n}{n+\lambda - \beta n} {\bm D}^{-1/2}\bm X_{\bm L} {\bm D}^{1/2}$ & \cite{zhou2003learning} \\ [.3ex]
4 & $\beta {\bm I} + {\bm S}^{-1/2} \mc{\bm L} {\bm S}^{-1/2}$ & $\frac{n \beta + \lambda}{n} $ & $ \bm M_{\lambda,\beta} = 2\lambda {\bm S}^{-1/2}\bm X_{\mc L} {\bm S}^{1/2}$ & \cite{johnson2008graph} \\ [.3ex] 
5 & ${\bm S}^{-1/2}  (\beta {\bm I} + \mc{\bm L} ) {\bm S}^{-1/2}$ & $ 2 \lambda $ & $ \bm M_{\lambda,\beta} = \big( \frac{\bm S^{1/2}}{4 n\lambda} + \frac{\beta\bm S^{-1/2}}{4\lambda^2} \big)^{-1} \bm X_{\mc L} {\bm S}^{1/2}$ & \cite{johnson2007effectiveness} \\ [.3ex] 
6 & $\mathcal{\bm L} + b\cdot \bm 1\bm 1^\top + \beta \bm I$ & $\beta + \frac{\lambda}{n} $ & $\bm M_{\lambda,\beta} = 2\lambda \bm X_{\mc L} \big( \bm I - \frac{b \bm 1\bm 1^\top}{\alpha + n b} \big) $ & \cite{herbster2005online} \\ [.3ex] 
 \bottomrule
 \end{tabular}
 \end{table*}

\textbf{Problem formulation.\quad} This paper considers the following online learning paradigm on $\mc{G}$: At each time $t=1,2,\ldots,n$, a learner picks  a node $v$ and makes a prediction $\hat{\bm y}_v \in \mathcal{Y}$. The true label $\bm y_v$ is revealed by the adversary with a corresponding 0-1 loss $\ell(\hat{\bm y}_v, \bm y_v) = 1 - {\bm y}_v^\top \hat{\bm y}_v$ back to the learner.
The goal is to design an algorithm, so the learner makes as few mistakes as possible. 
Denote a prediction of $\mc{V}$ as $\widehat{\bm Y} = \left[\hat{\bm y}_1, \hat{\bm y}_2, \ldots, \hat{\bm y}_n \right] \in \mc F$ and true label configuration as ${\bm Y} = \left[{\bm y}_1, {\bm y}_2, \ldots, {\bm y}_n \right] \in \mc F$ from the allowed label configurations $\mc F \in \left\{ \bm F \in \{0,1\}^{k\times n} : {\bm F}_{:,j}^\top
\cdot {\bm 1} = 1, \forall j \in \mc{V}\right\}$. Without further restrictions, the adversary could always select a $\bm Y$ so that the learner  makes the maximum ($n$) mistakes by always providing $\bm y_v \ne \hat{\bm y}_v$. Therefore, to have learnability, often the set   $\mc F$ is restricted to capture \emph{label smoothness} \cite{blum1998combining,blum2001learning,zhu2003semi,zhou2003learning,blum2004semi}. Formally, given an algorithm $\mc{A}$, the learner's goal is to  minimize the  regret defined as 
\begin{equation}
\mathop{\regret}_{\widehat{\bm Y} \sim \mc{A}} := \sum_{t=1}^{n} \ell(\hat{\bm y}_t, \bm y_t) - \min_{\bm F \in \mc{F}_{\lambda,\beta}} \sum_{t=1}^{n} \ell(\bm F_{:,t}, \bm y_t), \label{equ:def:regret}
\end{equation}
where  
\begin{equation}
\mc{F}_{\lambda,\beta} = \left\{ \bm F \in \mc{F}: \sum_{i=1}^k {{\bm F}_{i,:}}^\top {\bm K}_\beta^{-1} {\bm F}_{i,:} \leq \lambda \right\},
\label{equ:quadform}
\end{equation}
and $\bm K_\beta$ is a positive definite kernel parameterized by $\beta$, and $\lambda$ is a label smoothing parameter controlling the range of allowed label configurations. For example, assume $\bm K_\beta^{-1} = \bm D - \bm W$ for a unit weight graph; then if $\lambda = 1$,  $\mb y_i = \mb y_j$ whenever $(i,j)\in \mc E$ for all $\mb Y\in \mc F$; clearly in this case the labeling is learnable. On the other hand, if $\lambda > n$, then $\mb Y$ can be any labeling and is not learnable. If $\ell$ is convex with a closed convex set $\mathcal{F}$, typical online convex optimization methods such as online gradient descent or Follow-The-Regularized-Leader could provide sublinear regret \cite{shalev2012online,hazan2016introduction} for minimizing the regret \eqref{equ:def:regret}. However, when $\ell$ is the 0-1 loss, the combinatorial nature of $\mathcal{F}$ makes directly applying these methods difficult. Inspired by \citet{rakhlin2017efficient}, we propose the following convex relaxation of   $\mc{F}_{\lambda,\beta}$ to
\begin{equation}
\bar{\mc{F}}_{\lambda, \beta} =  \left\{{\bm F} \in \mathbb{R}^{k \times n}: \sum_{i=1}^{k} {\bm F}_{i,:}^{\top} \bm M_{\lambda,\beta} {\bm F}_{i,:} \leq 1\right\}, \nonumber
\end{equation}
where $\mc{F}_{\lambda,\beta} \subseteq \bar{\mc{F}}_{\lambda,\beta}$ and the regularized kernel matrix is
\begin{equation}
\bm M_{\lambda,\beta} = \left( \frac{{\bm K}_\beta^{-1}}{2\lambda} + \frac{\bm I_n}{2n} \right)^{-1}. \label{equ:regularized-kernel}
\end{equation}

\subsection{Online relaxation and surrogate loss}

\begin{algorithm}
\caption{\textsc{Relaxation}$(\mc{G}, \lambda, D)$(\citeauthor{rakhlin2017efficient})}
\begin{algorithmic}[1]
\STATE Compute $\bm M = \bm M_{\lambda, \beta}$
\STATE $T_1 = \tr{\bm M}, A_1 = 0, \bm G = [\bm 0, \ldots, \bm 0] \in \R^{k\times n}$
\FOR{$t = 1,\ldots, n$}
\STATE $\bm \psi_{t}= - \bm G \bm M_{:,t}  / \sqrt{A_t + D^2 \cdot T_t}$
\STATE Predict $\hat{\bm y}_t \sim \bm q_t(\bm \psi_t)$, $\bm \nabla_t  = \bm \nabla \phi_{\bm \psi_t}(\cdot,\bm y)$
\STATE Update $\bm G_{:,t} = \bm \nabla_t$
\STATE $A_{t+1} = A_{t} + 2 \bm \nabla_t^\top \bm G \bm M_{:,t} + m_{tt}\cdot \| \bm \nabla_t \|_2^2$
\STATE $T_{t+1} = T_t - m_{tt}$
\ENDFOR
\end{algorithmic}
\label{algo:relaxation}
\end{algorithm}
In the \textit{online relaxation framework} (Alg.\ref{algo:relaxation}), a key step of prediction node $t$ is to choose a suitable $\bm \psi_t$ strategy so that the regret defined in \eqref{equ:def:regret} can be bounded. Specifically, the prediction $\hat{\bm y}_t$ is randomly generated according to distribution $\bm q_t(\bm \psi_t)$ where the score $\bm \psi_t \in \mathbb{R}^k$ is a scaling of $-\sum_{i < t}\bm \nabla_{i} M_{i,t}$, computed in an online fashion. The distribution, $q_i = \max(\psi_i-\tau,0)$ for the choice of $\tau$ such that $\sum_{i=1}^k q_i = 1$. This technique corresponds to minimizing the surrogate convex loss 
\footnote{The method could naturally apply to other types of losses (See more candidate losses in \citet{johnson2007effectiveness}).}
\begin{equation}
\phi_{\bm \psi}(\bm g, \bm y) =
\begin{cases}  \frac{1 + \max_{r: \bm e_r \ne \bm y} \left\{\bm g^\top \bm e_r - \bm g^\top \bm y \right\}}{1+1 /|S(\bm \psi)|} &  \bm y \notin {S}(\bm \psi) \\ 1 - \bm g^{\top} \bm y + \frac{\bm g^\top \bm 1_{S(\bm \psi)}-1}{|S(\bm \psi)|} & \bm y \in S(\bm \psi)\end{cases} \label{equ:surrogate-loss}
\end{equation}
where $S\left(\bm \psi\right)$ is the support of $\bm \psi$, and $\bm \nabla_t$ is the gradient of $\phi(\cdot, \bm y)$ of the first variable. 
 Specifically, $ y_t \notin S(\bm \psi)$ means the learner receives loss $\phi_{\bm \psi}(\bm g, \bm y) \geq 1$. Note that the per-iteration cost of Alg.\ref{algo:relaxation} is $\mathcal{O}(k n)$ once the $\bm \psi_t$ is computed. 

We now define an \emph{admissible relaxation function}.

\begin{definition}[Admissible function \cite{rakhlin2012relax}]
Let $\bm \nabla_i\in \R^k$, $\|\bm \nabla_i\|_2 \leq D$ for some $D>0$.
A real-valued function $\rel({\bm \nabla}_{1:t})$ is said to be admissible if, for all $t\in \mathcal{V}$, it satisfies recursive inequalities
\begin{equation*}
\inf_{\bm \psi_t \in \R^k} \sup_{\|\bm \nabla_t\|_2 \leq D} \left\{ \bm \nabla_t^\top \bm \psi_t + \rel({\bm \nabla}_{1:t}) \right\} \leq \rel({\bm \nabla}_{1:t-1}),
\end{equation*}
\label{def:admissibility}
with $\rel(\bm \nabla_{1:n}) \geq - \inf_{\bm F \in \mathcal{F}_{\lambda,\beta}} \sum_{t=1}^n\bm \nabla_t^\top \bm F_{:,t}$.
\end{definition}

It was shown in \citet{rakhlin2012relax} (and later \cite{rakhlin2015hierarchies,rakhlin2016tutorial,rakhlin2017efficient}) that if there exists an admissible function $\rel$ for some $\bm \psi_t$, then the regret of Alg.\ref{algo:relaxation} is upper bounded by $\rel(\emptyset) = \sqrt{D\cdot\tr{\bm M}}$, providing an upper bound of the regret. 
Here $\bm M$ is either $ (\tfrac{\mathcal{L}}{2\lambda} + \tfrac{\bm I}{2 n})^{-1}$ or $(\tfrac{\bm L}{2\lambda} + \tfrac{\bm I}{2 n})^{-1}$ for the binary case. Note that in both cases, since $\lambda_{\max}(\bm M)\leq n$, then in the worst case, the regret could be $\mathcal{O}(n)$ (vanishing in general). Thus, two questions remain.
\begin{enumerate}
\item Does there exist $\bm K_\beta^{-1}$ that not only captures label smoothness but also has regret \emph{smaller} than $\mc O(n)$?
\item How do we reconcile the kernel computation overhead $\mc O(n^3)$ but still provide an effective regret bound?
\end{enumerate}
These two main problems motivate us to study this online relaxation framework further. Sec. \ref{sect:local-approx-kernel} answers the second question by showing that solving many popular kernel matrices is equivalent to solving two basic kernel matrices, and we explore local approximate methods for both. We then answer the first question in Sec. \ref{section:fast-onl}  by proving effective bounds when the parameterized kernel matrix is computed exactly or approximated.

\section{Local approximation of  kernel $\bm M_{\lambda,\beta}$}
\label{sect:local-approx-kernel}

Section \ref{sect:local-approx-kernel:graph-kernel} presents how popular kernels can be evaluated from simple transformations of the inverse approximations computed via \textsc{FIFOPush}, whose convergence is described in  Section \ref{sect:local-approx-kernel:local-convergence}.

\subsection{Basic kernel presentations of $\bm M_{\lambda,\beta}$}
\label{sect:local-approx-kernel:graph-kernel}
The regularized kernel matrix is defined in \eqref{equ:regularized-kernel} for various instances of $\bm K_\beta^{-1}$ as listed in Tab. \ref{tab:graph-kernel-presentation}.
As shown in the table, a key observation is that several existing online labeling methods involve the inverse of two basic kernel forms. We present this in Thm. \ref{thm:basic-kernel-representation}.
\begin{theorem}
Let $\bm K_\beta^{-1}$ be the inverse of the symmetric positive definite kernel matrix defined in Tab. \ref{tab:graph-kernel-presentation}. Then $\bm M_{\lambda,\beta}$ can be decomposed into $\bm M_{\lambda,\beta} = a \bm A^{-1}\bm X \bm B$, which is easily computed once $\bm X$ available. $\bm X$ represents two basic kernels
\begin{align*}
\bm X_{\mc L} = \left(\alpha \bm I + \mc L\right)^{-1},\quad \bm X_{\bm L} = \alpha \left(\bm I - (1-\alpha) \bm W \bm D^{-1}\right)^{-1}
\end{align*}
corresponding to the inverse of variant matrices of $\mc L$ and $\bm L$, respectively.\footnote{Note  $\bm L = \alpha \bm D^{1/2}(\bm I - (1-\alpha)\bm D^{-1/2}\bm W \bm D^{-1/2})^{-1}\bm D^{-1/2}$.}
\label{thm:basic-kernel-representation}
\end{theorem}
In Tab. \ref{tab:graph-kernel-presentation}, the column ``Basic Kernel Presentation" shows how $\bm M_{\lambda,\beta}$ can then be efficiently computed from either $\bm X_{\bm L}$ or $\bm X_{\mc L}$, using minimal post-processing overhead. As in the online relaxation framework, for any time $t$ of node $v_t$, it requires to access the $v_t$-th column of $\bm M_{\lambda,\beta}$. Therefore, we need to solve the following two basic equations
\begin{align}
\text{Type-}\mc L\text{:\quad} {\bm x}_{v_t} &= \bm X_{\mc L}{\bm e}_{v_t}, \label{equ:kernel-1}\\
\text{Type-}\bm L\text{:\quad} {\bm x}_{v_t} &= \bm X_{\bm L}{\bm e}_{v_t}. \label{equ:kernel-2}
\end{align}
For the second case, note ${\bm x}_{v_t} = \bm X_{\bm L}{\bm e}_{v_t}$ gives the Personalized PageRank Vector (PPV) \cite{page1999PageRank,jeh2003scaling}. For example, using $\alpha = \tfrac{n}{n+\kappa}$, we compute $\bm M_{\lambda,\beta} = 2 n {\bm D}^{-1/2} \bm X_{\bm L} {\bm D}^{1/2}$  where   $\bm X_{\bm L}$ is the Personalized PageRank matrix (See the second row of Tab. \ref{tab:graph-kernel-presentation}). We now discuss the inversion for computing $\bm X_{\bm L}$ and $\bm X_{\mc L}$.

\begin{algorithm}
\caption{$\textsc{FIFOPush}(\mathcal{G},\eps,\alpha, s)$}
\begin{algorithmic}[1]
\STATE $\bm r= \begin{cases}
\text{Type-}{\mc L}: \tfrac{\bm e_s}{\alpha} \\
\text{Type-}{\bm L}: \bm e_s
\end{cases}$
\STATE $\bm x = \bm 0, \mc{Q}=[s]$
\WHILE{$\mc{Q}\ne \emptyset$}
\STATE $u = \mc{Q}\text{.pop()}$
\IF{$r_{u} < \eps \cdot d_{u}$}
\STATE \textbf{continue}
\ENDIF
\STATE $x_{u} = \begin{cases} 
\text{Type-}{\mc L}\text{: } x_u + \tfrac{\alpha r_u}{\alpha + D_u}\\
\text{Type-}{\bm L}\text{: } x_{u} + \alpha r_{u}
\end{cases}$ 
\FOR{$v \in \nei(u)$}
\STATE $r_{v} = \begin{cases} \text{Type-${\mc L}$: } r_v + \tfrac{r_u}{\alpha + D_u} \cdot w_{u v} \\
\text{Type-${\bm L}$: } r_{v} + \frac{(1-\alpha) r_{u}}{D_{u}} \cdot w_{u v}
\end{cases}$
\IF{$v \notin \mathcal{Q}$}
\STATE $\mathcal{Q}\text{.push}(v)$
\ENDIF
\ENDFOR
\STATE $r_{u} = 0$
\ENDWHILE
\STATE \textbf{Return} $\bm x,\bm r$
\end{algorithmic}
\label{algo:fifo-push}
\end{algorithm}

Before introducing the \textsc{FIFOPush} inversion method, let us consider the more commonly used power iteration for matrix inversion. $\bm M_{\lambda,\beta}$ can be approximated by a series of matrix multiplications. Take $\bm K_{\beta}^{-1} = \bm I - \bm D^{-1/2}\bm W\bm D^{-1/2}$ as an example. Then a truncated power iteration gives
\begin{align}
\bm M_{\lambda,\beta} 
&\approx \frac{2n\lambda}{n + \lambda} \sum_{i=0}^p\Big( \frac{ n}{n+\lambda} \bm D^{-1/2}\bm W\bm D^{-1/2} \Big)^i \label{inequ:approximate-method}
\end{align}
assuming that $\|\tfrac{ n}{n+\lambda} \bm D^{-1/2}\bm W\bm D^{-1/2}\|_2 < 1$ (See Lemma 2.3.3 of \citet{golub2013matrix}). When $p$ is small, this method often produces a reasonable and efficient approximation of $\bm M_{\lambda,\beta}$, and is especially efficient if $\bm W$ is sparse.  However, as  $p$  gets large, the intermediate iterates quickly become dense matrices, creating challenges for online learning algorithms where the per-kernel vector operator is preferred ( See also Fig. \ref{fig:nnz-rate} on real-world graphs ). Thus, the situation is even worse when we only need to access one column at each time $t$ under the online learning setting. 

\begin{figure}[!htbp]
\centering
\includegraphics[width=.45\textwidth]{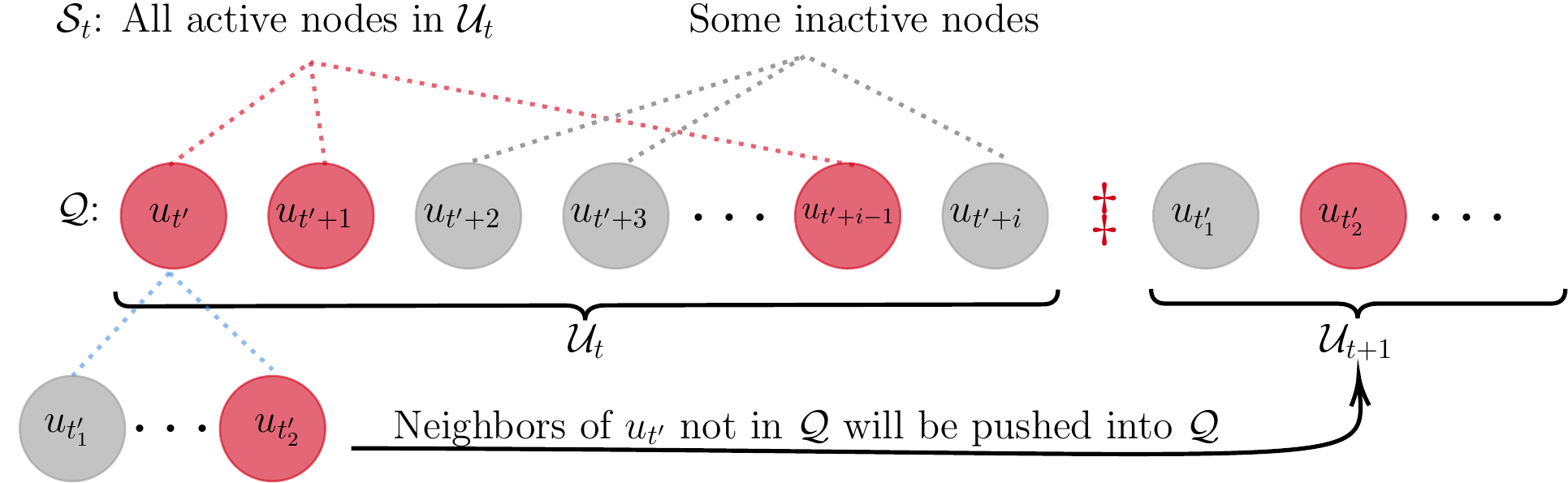}
\caption{The illustration of $t$-th epoch of \textsc{FIFOPush}. At the initial of $t$-th epoch, $\mc{Q} = [u_{t'},\ldots, u_{t^\prime+i}]$ contains all active nodes (red) $\mc{S}_t$ and part of inactive nodes (gray). After the first active node $u_{t'}$ has been processed, the neighbors of $u_{t'}$ that are not in $\mc{Q}$ will be pushed into $\mc{Q}$ for next $(t+1)$-th epoch. $\red{\ddag}$ is a dummy node to mark the end of an epoch.}
\label{fig:active-queue-nodes}
\end{figure}

For these reasons, we introduce \textsc{FIFOPush} (Alg. \ref{algo:fifo-push}), which reduces to the well-known \textsc{APPR} algorithm \cite{andersen2006local} when the goal is to approximate $\bm X_{\bm L}$. Specifically, it is a \textit{local push} method for solving either \eqref{equ:kernel-1} or \eqref{equ:kernel-2} based on First-In-First-Out (FIFO) queue. Each node $u_t \in \mc S_t$ is either \textit{active}, i.e., $r_{u_t} \geq \eps d_{u_t}$, or \textit{inactive} otherwise. As illustrated in Fig. \ref{fig:active-queue-nodes}, at a higher level, it maintains a set of nonzero residual nodes $\mc U_t$ and active nodes $\mc S_t \subseteq \mc U_t$ in each epoch $t$. \textsc{FIFOPush} updates the solution column $\bm x$ and residual $\bm r$ (corresponding to the ``gradient") by shifting mass from a high residual node (an active node) to its neighboring nodes in $\bm x$ and $\bm r$. This continues until all residual nodes are smaller than a tolerance $\epsilon$.
This method essentially retains the linear invariant property introduced in Appendix \ref{sect:appendix:local-convergence-proof}.

\subsection{Local linear convergence of \textsc{FIFOPush}}
\label{sect:local-approx-kernel:local-convergence}

\begin{figure*}
\centering 
\subfigure[Labeled Karate]{\label{fig:type-ii-a}\includegraphics[width=50mm]{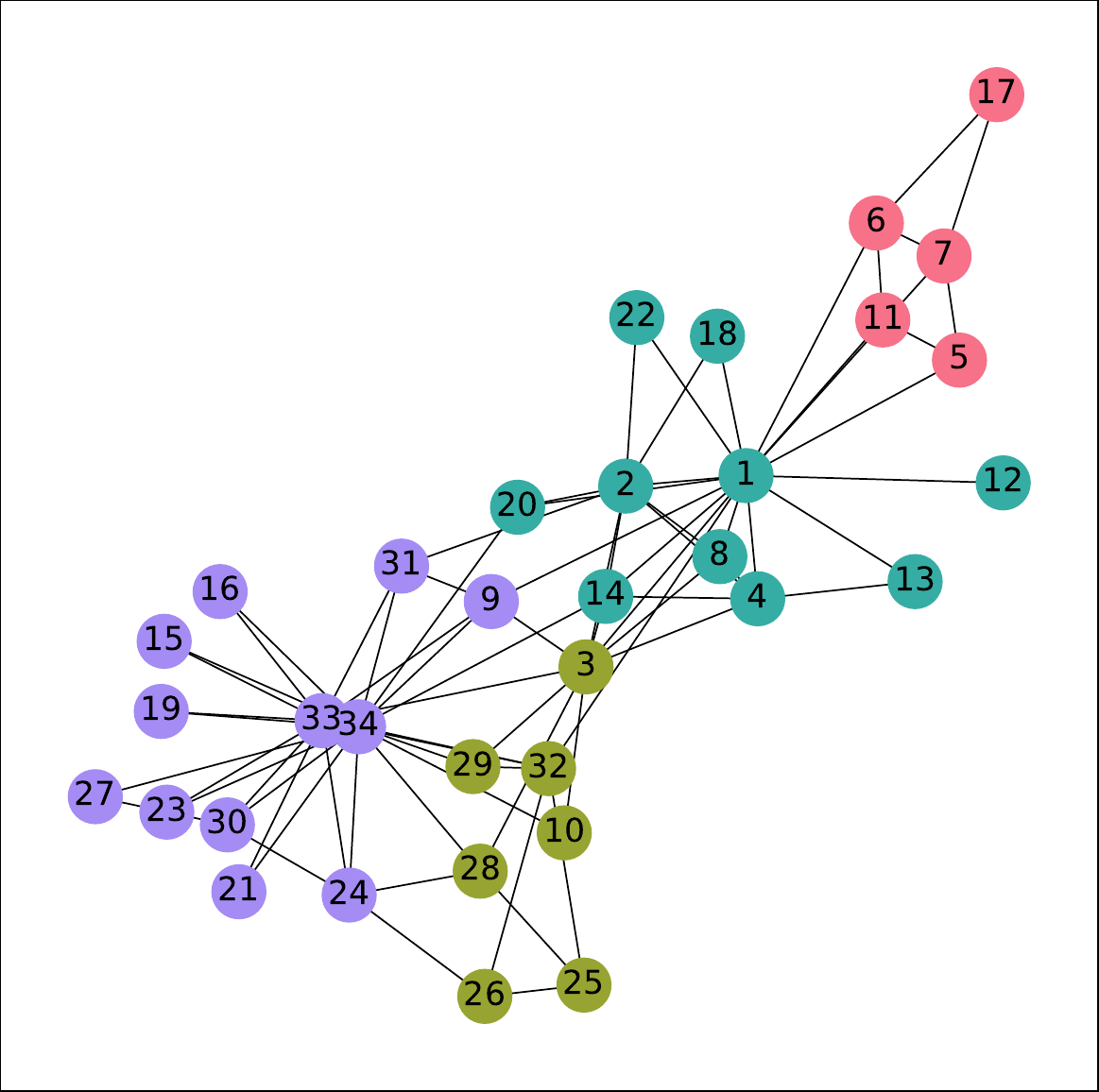}}
\subfigure[$\bm x_{22,\eps}$ on $\mathcal{G}$]{\label{fig:type-ii-b}\includegraphics[width=50mm]{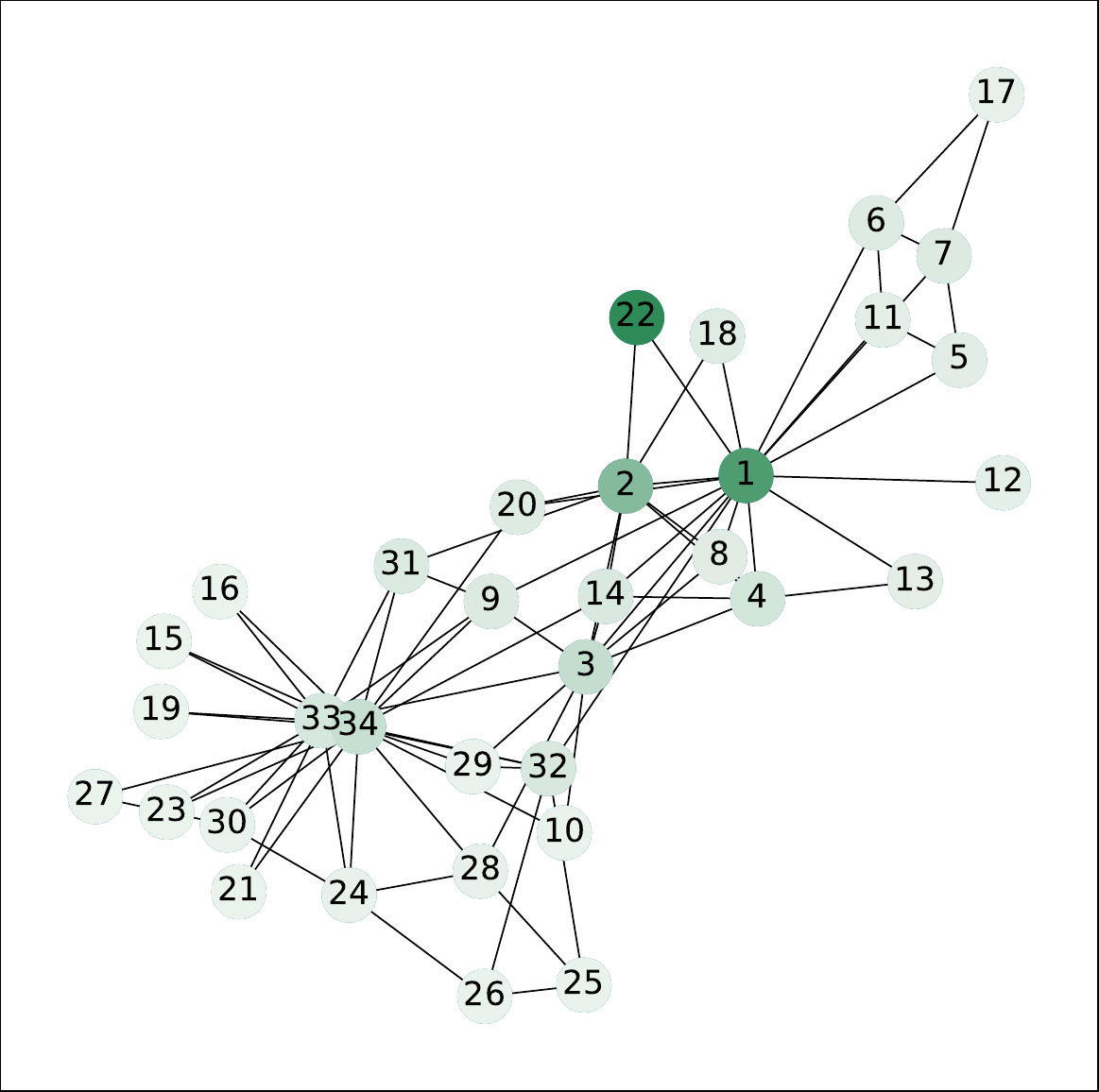}}
\subfigure[Power law of $\bm X_{\eps}$]{\label{fig:fig-type-ii-c}\includegraphics[width=57mm]{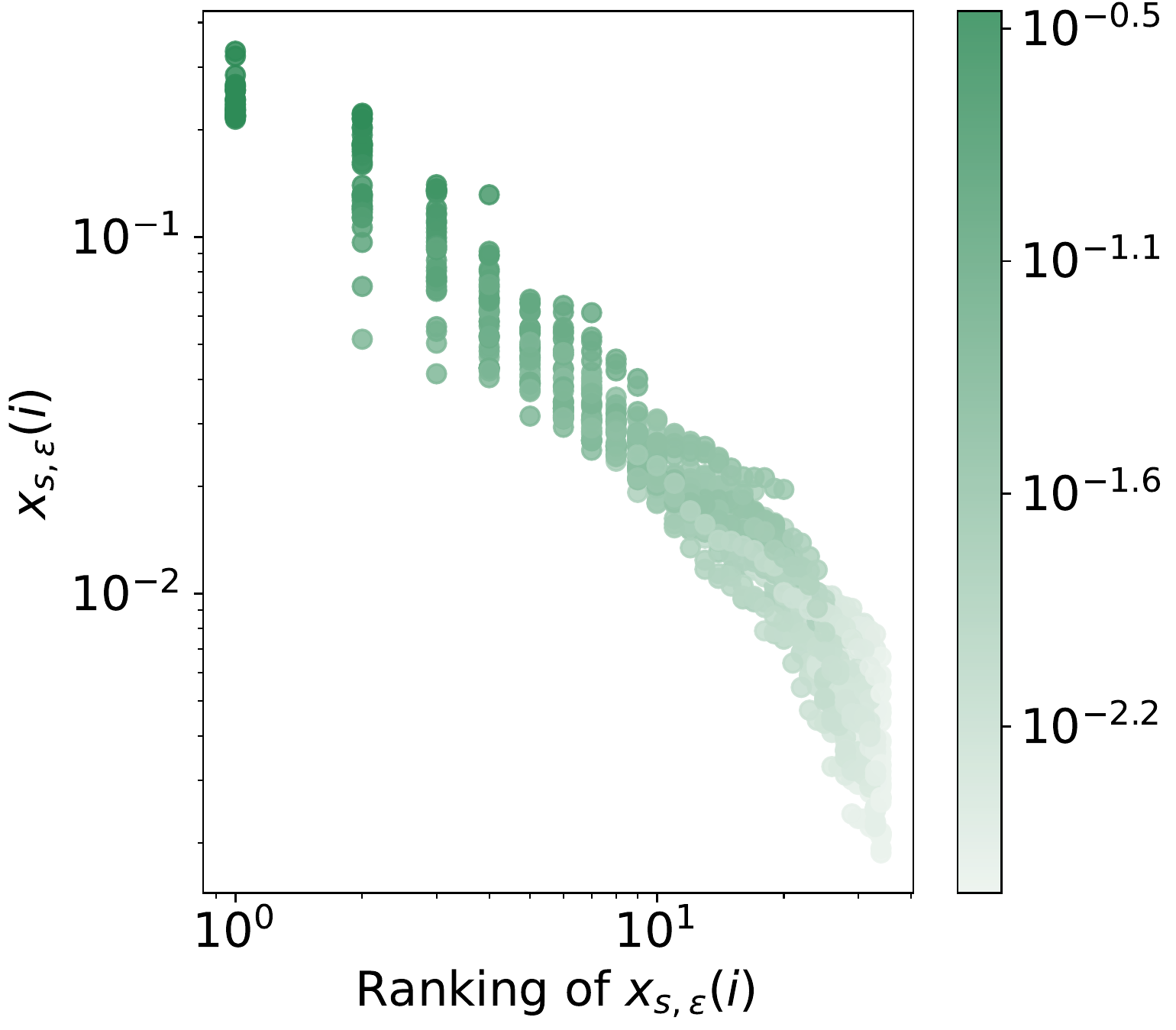}}
\caption{Power law distribution of $|(\bm X_{\bm L}\bm e_s)_i|$ on Karate graph \cite{girvan2002community}. (a) The Karate contains 34 nodes and 78 edges forming four communities ($w_{u v}$ is the 2-dimensional Euclidean distance.); (b) We run \textsc{FIFOPush}$(\mc G, \eps,\alpha,s)$ on $s=22$ with $\alpha = 0.2$ and $\eps = 10^{-12}$. Neighbors of $s=22$ have large magnitudes of $\bm x_{s}$; (c) The power law distribution of entries of $\bm x_s$ for all 34 nodes.}
\label{fig:power-law-of-x-s-epsilon}
\end{figure*}

For calculating $\bm X_{\bm L}$, \citeauthor{andersen2006local} showed that \textsc{FIFOPush} gives a time complexity $\mathcal{O}(\tfrac{1}{\alpha \eps})$ for precision level $\eps>0$. \footnote{This algorithm was also  proposed in \citet{berkhin2006bookmark}, namely Book Coloring algorithm.}
This bound is \textit{local sublinear}, meaning the bound is locally dependent of $\mathcal{G}$ and sublinear to the precision $\eps$. Moreover, the rate's independence on $\mathcal{G}$ is a key advantage of \textsc{FIFOPush} over other numerical methods such as Power Iteration, which is similar to \textsc{FIFOPush} \cite{wu2021unifying} when $\eps < \mathcal{O}(m^{-1})$ (recall $m = |\mc E|$). Specifically, the Power Iteration typically needs $\mathcal{O}(\tfrac{m}{\alpha}\log\tfrac{1}{\eps})$ operations. However, when $m$ is large, and $\eps$ is very small, the advantage of the local sublinear bound is lost, and the time complexity bound is not optimal. 

It is natural to ask whether any method achieves a locally dependent and logarithmically related bound to $\eps$.
We answer this question positively. Specifically, we notice that in most real-world sparse graphs, the columns of $\bm X$ have magnitudes following a power law distribution (See Karate graph in Fig. \ref{fig:power-law-of-x-s-epsilon}, real-world graphs shown in Fig. \ref{fig:power-low-normalized} and \ref{fig:power-low-unnormalized} for $\bm X_{\bm L}$ and $\bm X_{\mc L}$, respectively.), suggesting that local approximations are sufficient in computing high fidelity approximate inverses. Note that this greatly improves computational complexity and preserves memory locality, reducing graph access time for large-scale graphs.

 We now provide our \textit{local linear} and \emph{graph independent} complexity bound.
 Denote the estimation matrix $\bm X_{\eps} = \left[\bm x_{1,\eps}, \ldots, \bm x_{n,\eps}\right]$ and the residual matrix $\bm R_{\eps} = \left[\bm r_{1,\eps}, \ldots, \bm r_{n,\eps}\right]$ where $(\bm x_{s,\eps}, \bm r_{s,\eps}) = \textsc{FIFOPush}(\mc{G},\eps,\alpha,s)$ for all $s\in \mc{V}$. Denote $\mc I_t = \supp{\bm r_t}$ as the support of residual after $t$-th epoch, and $\mc S_t$ as the set of \textit{active} nodes.
 
\begin{theorem}[Local linear convergence of \textsc{FIFOPush} for $\bm X_{\bm L}$] Let  $\bm x_s = \bm X_{\bm L} \bm e_s$. Denote $T$ as the total epochs executed by \textsc{FIFOPush},  and $\mc{S}_t : = \{v: r_t(v) \geq \eps \cdot d_{v}, v \in \mc{I}_t\}$ as the set of active nodes in $t$-th epoch. Then, the total operations of $ \textsc{FIFOPush}(\mc{G},\eps,\alpha,s)$ is dominated by 
\begin{equation}
R_T := \sum_{t=1}^T\sum_{u_t\in {\mc S}_t} d_{u_t} \leq \frac{\vol{\mc{S}_{1:T}}}{\alpha \cdot \eta_{1:T}} \log\left( \frac{C_{\alpha,T}}{\eps} \right), \label{inequ:local-linear-bound}
\end{equation}
where $\vol{\mc{S}_{1:T}} = \sum_{t=1}^{T}\vol{\mc{S}_t} / T$ is the average volume of $\mc{S}_t$. Additionally, $\eta_{1:T} = \sum_{t=1}^{T} \eta_t / T$ is the average of local convergence factors $\eta_t \triangleq \sum_{u \in \mc{S}_t} d_u / \sum_{v \in \mc{I}_{t}} d_v$, and $C_{\alpha,T} = 1/(\sum_{v\in \mc{I}_T} (1-\alpha)d_u w_{u v} / D_u)$. For $s, i \in \mc{V}$, we have $\bm x_s = \bm x_{s,\eps} + \bm X_{\bm L} \bm r_{s,\eps}, r_{s,\eps}(i) \leq [0, \eps d_i)$.
\label{thm:local-convergence}
\end{theorem}
Thm. \ref{thm:local-convergence} provides a \textit{local linear convergence} of \textsc{FIFOPush} where both $\vol{\mc{S}_{1:T}}$ and ${\eta}_{1:T}$ are locally dependent on $\mathcal{G}$, $\alpha$, and $\eps$. For unweighted $\mathcal{G}$, the bound in \eqref{inequ:local-linear-bound} can be simplified as $\tfrac{\vol{\mc{S}_{1:T}}}{\alpha \cdot \eta_{1:T}} \log\tfrac{1}{\eps(1-\alpha)|\mc{I}_T|}$. The key of Thm. \ref{thm:local-convergence} is to evaluate $\vol{\mc S_{1:T}}$ and $\eta_{1:T}$. To estimate $\vol{\mc S_{1:T}}$, since each active node appears at most $T$ times and at least once in all $T$ epochs, after \textsc{FIFOPush} terminates, we have
\begin{equation}
\frac{\vol{\supp{\bm x_{s,\eps}}}}{T} \leq \vol{\mc S_{1:T}} \leq \vol{\supp{\bm x_{s,\eps}}},\nonumber
\end{equation}
where for $\alpha$ and $\eps$ such that $\mathcal{O}(|\supp{\bm x_{s,\eps}}|) \ll n$ means $\mathcal{O}(\vol{\supp{\bm x_{s,\eps}}}) \ll m$ in expectation. More importantly,
compared with $\mc{O}(1/\alpha\eps)$ of \citet{andersen2006local}, Thm. \ref{thm:local-convergence} provides a better bound when $\epsilon \leq \mc{O}(m^{-1})$. The work of \citet{fountoulakis2019variational} shows \textsc{APPR} is equivalent to the coordinate descent method, and the total time complexity is comparable to $\tilde{\mathcal{O}}(\tfrac{1}{\alpha\eps})$. 

The average of the linear convergence factor $\eta_{1:T}$ is always $>0$ by noticing that at least one active node is processed in each epoch. One can find more quantitative discussion in Appendix \ref{sect:appendix:local-convergence-proof}. The above theorem is a refinement of \cite{wu2021unifying} where $\mathcal{O}(m\log\frac{1}{\eps})$ is obtained (only effective when $\eps < 1/2m$). However, our proof shows that obtaining $m$ independent bound is possible by showing that local magnitudes are reduced by a constant factor. The above theorem gives a way to approximate $\bm M_{\lambda,\beta}$, and we will build an approximate online algorithm based on \textsc{FIFOPush}. We close this section by introducing our \textit{local linear convergence} for $\bm X_{\mc L} \bm e_s$ as the following.

\begin{theorem}[Local convergence of \textsc{FIFOPush} for $\bm X_{\mc L}$] Let $\bm x_s = \bm X_{\mc L} \bm e_s$ and run Aglo. \ref{algo:fifo-push} for $\bm X_{\mc L}$. For $s, i \in \mc{V}$, we have $ \bm x_s = \bm x_{s,\eps} + \alpha \bm X_{\mc L} \bm r_{s,\eps}, \text{ with } r_{s,\eps}(i) \leq [0,\eps d_i), \forall i\in \mathcal{V}$. The main operations of $ \textsc{FIFOPush}$ for $\bm X_{\mc L}$ is bounded as
\begin{equation}
R_T \leq \frac{\vol{\mc S_{1:T}}(\alpha + D_{\max})}{\alpha \cdot \eta_{1:T}} \log\left( \frac{C_{\alpha,T}}{\eps} \right), \label{equ:time-complexity-2}
\end{equation}
where $\vol{ \mc S_{1:T}}$ and $\eta_{1:T}$ are the same as in Thm. \ref{thm:local-convergence}, 
$\eta_t \triangleq \tfrac{\sum_{u \in \mc{S}_t} d_u/(\alpha + D_u)}{\sum_{v \in \mc{I}_{t}} d_v/(\alpha + D_v)}$, 
$C_{\alpha,T} = 1 / \sum_{v\in \mc{I}_{T}} \frac{d_u w_{u v}}{\alpha + D_u}$, and $D_{\max} = \max_{v\in \supp{\bm x_{s,\eps}}} D_v$. 
\label{thm:local-convergence-2}
\end{theorem}
\begin{remark}
We obtain a similar \textit{local linear convergence} for solving $\bm X_{\mc L}$ by \textsc{FIFOPush}. The additional factor $(\alpha + D_{\max})$ appears in Equ. \eqref{equ:time-complexity-2} due to the upper bound of the maximal eigenvalue of $\mc L$.
\end{remark}

\begin{figure}
\centering    \includegraphics[width=.45\textwidth]{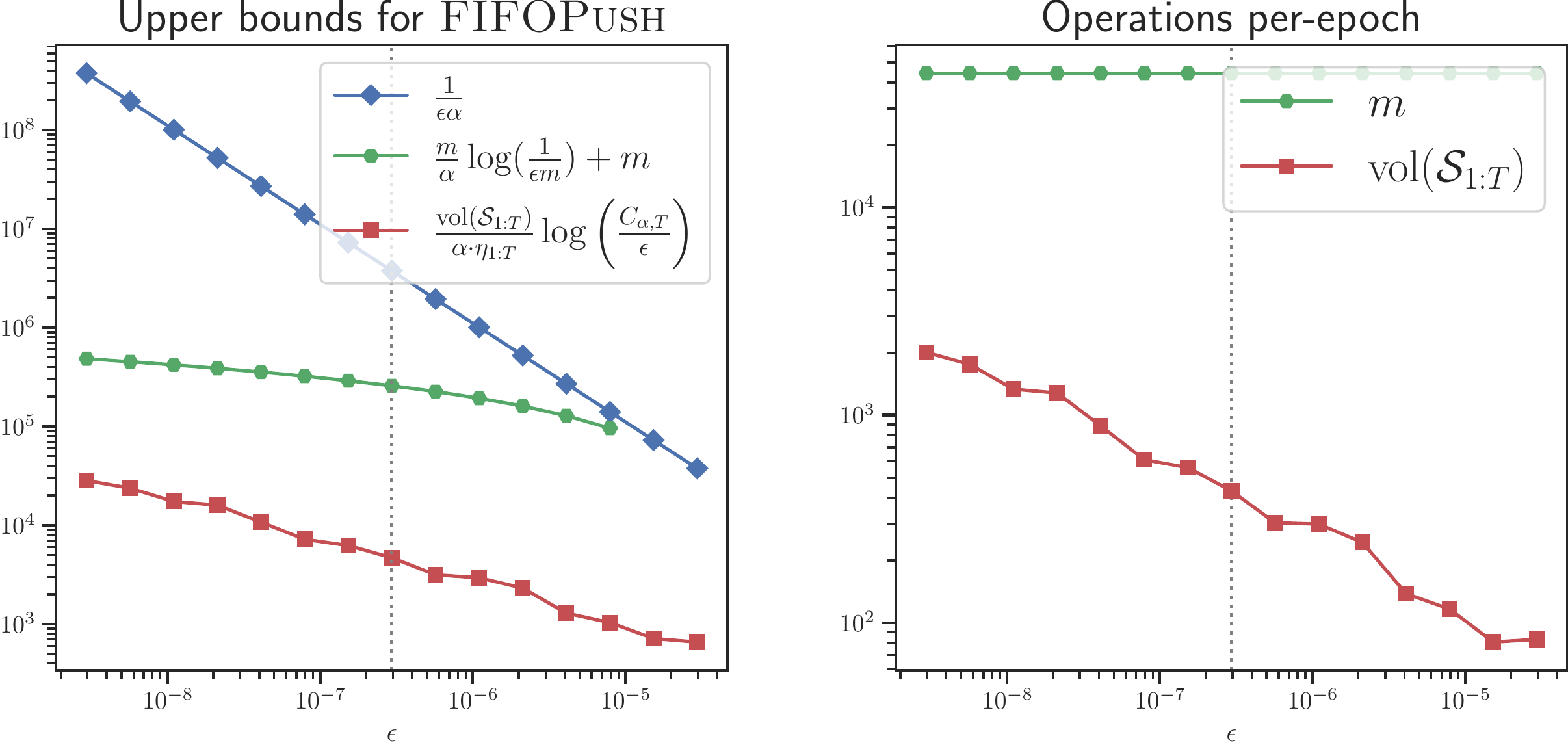}
\caption{The bounds comparison of $R_T$. To see if there is a true advantage of our bound, we compare two bounds of \textsc{FIFOPush}, $\mc{O}(\tfrac{1}{\alpha\eps})$ \cite{andersen2006local} and $\mc{O}(m\frac{m}{\alpha}\log\frac{1}{\eps m} + m)$ \cite{wu2021unifying} with Ours. Each vertical line with its left part is when $\eps$ satisfies Cor. \ref{corollary:total-time-complexity}\vspace{-5mm}}
\label{fig:theoretical-bound-pubmed-alpha-0.9}
\end{figure}
 
\section{Fast Online Node Labelling: \textsc{FastONL}}
\label{section:fast-onl}
This section shows how we can obtain a meaningful regret using parameterized graph kernels for the original online relaxation method. We then design approximated methods based on \textsc{FIFOPush}.

\subsection{Regret analysis for online relaxation method}

As previously discussed, simply applying $\bm M_{\lambda,\beta}^{-1} = \mc L$ or $\bm L$ will not yield  a meaningful regret as the maximal eigenvalue of $\bm M_{\lambda,\beta}$ depends on the minimal eigenvalue of $\bm M_{\lambda,\beta}^{-1}$. In particular, for a connected graph $\mc{G}$, the second smallest eigenvalue of $\bm L$ is lower bounded by $\operatorname{diam}(\mc{G}) \geq \frac{4}{\lambda_2 n}$ \cite{mohar1991eigenvalues}, and  is tight for a chain graph; this yields a $\mc O(n^2)$ bound which is non-optimal. Instead, our key idea of producing a method with improved bounds is to ``normalize'' the kernel matrix $\bm K_{\beta}^{-1}$ so that  $\tr{\bm M_{\lambda,\beta}} \ll \mc O(n^2)$, yielding a more meaningful bound. We state the regret bound as in the following theorem.

\begin{theorem}[Regret of \textsc{Relaxation} with parameterized $\bm M_{\beta}^{-1}$]
Let $\hat{\bm Y}$ be the prediction matrix returned by \textsc{Relaxation}, if the true label sequences $\bm Y \in \mathcal{F}_{\lambda,\beta}$ with parameter $\lambda = n^\gamma$ and $\gamma \in (0, 1)$. Then choosing $\beta = n^{\gamma -1}$ for kernel $\bm K_{\beta}^{-1} = \bm I - \beta \bm D^{-1/2}\bm W {\bm D}^{-1/2}$ and $\beta = 1-\frac{\lambda}{n}$ for $\bm K_\beta^{-1} = \beta \bm I + \bm S^{-1/2} \mathcal{L} \bm S^{-1/2}$, we have the following regret
\begin{equation}
\regret = \mathop{\mathbb{E}}_{\widehat{\bm Y} \sim \mathcal{A}} \sum_{t=1}^{n} \ell(\hat{\bm y}_t, \bm y_t) - \frac{2k-1}{k} \min_{\bm F \in \mc{F}_{\lambda,\beta}} \sum_{t=1}^{n} \ell(\bm f_t, \bm y_t),\nonumber
\end{equation}
\label{thm:regret-analysis-relaxation}
which is bounded i.e., $\regret \leq D \sqrt{2 n^{1+\gamma}}$. \footnote{Note that, for binary case, $k=2$, $\regret$ exactly recover $\mathop{\mathbb{E}}_{\widehat{\bm Y} \sim \mathcal{A}} \mathop{\regret}$ for binary case defined in Equ. \eqref{equ:def:regret}.}
\end{theorem}
\begin{remark}
The constant $D$ involved in the bound is the assumption of the bounded gradient of $\bm \nabla_t$, which is always $\leq 2$ for the loss chosen in \eqref{equ:surrogate-loss}. The above Thm. \ref{thm:regret-analysis-relaxation} is an improvement upon the regret given in \citet{rakhlin2017efficient} of $\mc O(n)$. Note that this rate does not take into account the run time $\mc O(n^3)$ required to invert $\bm M_{\lambda,\beta}$ in \textsc{Relaxation}. In the rest, we give the regret of \textsc{FastONL}, which implements \textsc{Relaxation} using  \textsc{FIFOPush}, and show that the regret is still small. 
\end{remark}

\subsection{Fast approximation algorithm \textsc{FastONL}}
We describe the approximated method, \textsc{FastONL} in Alg. \ref{algo:fast-onl} as follows, and recall that $m_{t t} = (M_\epsilon)_{t:t}$.

\begin{algorithm}
\caption{\textsc{FastONL}($\mc{G},\epsilon,\bm K_{\beta}^{-1},\lambda$)}
\begin{algorithmic}[1]
\STATE $\bm G = [\bm 0, \bm 0, \ldots, \bm 0] \in \mathbb{R}^{k\times n}$
\STATE $A_1 = 0$
\STATE $\bm M_\eps$ is obtained via \textsc{FIFOPush}$(\mathcal{G},\eps,\alpha,s)$ $\forall s\in\mc V$
\STATE $T_1 = \sum_{t=1}^n m_{t,t}$
\FOR{$t = 1,2,\ldots, n$}
\STATE $\bm v = \bm G (\bm M_\eps)_{:,t} + \bm G (\bm M_\eps)_{t,:}$
\STATE $\bm \psi_{t}= - \bm v / \sqrt{A_t + k \cdot T_t} $
\STATE Update gradient $\bm G_{:,t} = \bm \nabla_t = \bm \nabla \phi(\cdot,\bm y)$
\STATE $A_{t+1} = A_{t} + \bm \nabla_t^\top \bm v  + m_{t t}\cdot \| \bm \nabla_t \|_2^2$
\STATE $T_{t+1} = T_{t} - m_{t t}$
\ENDFOR
\end{algorithmic}
\label{algo:fast-onl}
\end{algorithm}

\begin{theorem}[Regret analysis of \textsc{FastONL} with approximated parameterized kernel] Consider the similar residual matrix $\Tilde{\bm R}_\eps = \bm D^{-1/2} \bm R_\eps \bm D^{1/2} $. Given $\lambda=n^\gamma$ for $\gamma\in(0,1)$, picking $\epsilon$ so that $\| \tilde{\bm R}_\eps \|_2  \leq \frac{1}{\alpha }$ yields
\[
\regret \leq D\sqrt{(1+k^2) n^{1+\gamma}},
\]
where the restriction on $\eps$ is due to maintaining the positive semidefiniteness of $\left(\bm M_\eps + \bm M_\eps^\top\right)/2$. 
\label{thm:regret-fastonl}
\end{theorem}
Based on Thm. \ref{thm:regret-fastonl}, we have the following runtime requirement for \textsc{FastONL}.
\begin{corollary}[Per-iteration complexity of \textsc{FIFOPush}]
Based on the conditions of Thm. \ref{thm:regret-fastonl}, the number of operations required in one iteration of \textsc{FastONL} is bounded by
\begin{equation}
\mathcal{O}\left( \frac{{\mc S}_{1:T}}{\alpha \cdot \eta_{1:T}} \log^{3/2}\left( n\right) \right).
\vspace{-5mm}
\end{equation}    
\label{corollary:total-time-complexity}
\end{corollary}
Fig. \ref{fig:theoretical-bound-pubmed-alpha-0.9} illustrates the advantage of our local bound by plotting all constants for the PubMed graph (Similar trends are observed in other graphs in Appendix \ref{sect:appendix:local-convergence-proof}). In practice, we observe that  $\eps \sim \mathcal{O}(n^{-1}) \gg \mathcal{O}(n^{-3/2})$, given from  a pessimistic estimation of $\|\bm D^{-1/2}\bm R_\eps\bm D^{1/2}\|_2$.
In particular, we notice a significant improvement of our bound over the previous ones when  $\alpha$ is large.

\begin{figure*}[t]
\centering
\includegraphics[width=.99\textwidth]{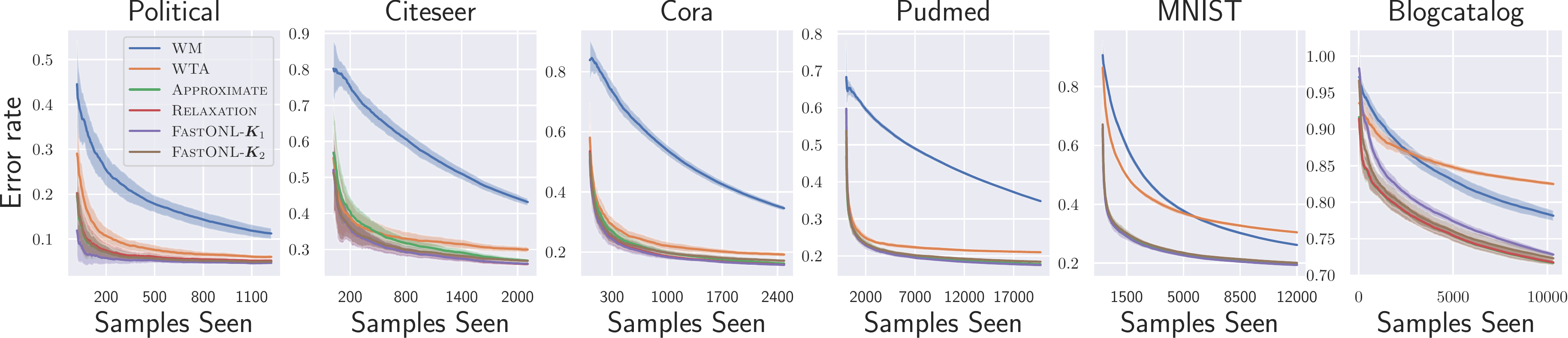}
\caption{Error rate as the function of samples seen on six small-scale graphs.\vspace{-5mm}}
\label{fig:fig-error-rate-small-graphs}
\end{figure*}

\paragraph{Practical implementation.\quad} A caveat of approximate inversion in \textsc{FastONL} is that $\bm M_\epsilon$ is not in general symmetric; therefore, for analysis, we require $\bm \psi_t$ to be computed using the symmetrized $\bm M_\eps[:,t]+\bm M_\eps[t,:]$, which requires row and column access at time $t$; effectively, this requires that $\bm M_\eps$ is fully pre-computed. While this does not affect our overall bounds, the memory requirements may be burdensome. However, when $\bm M_\epsilon \approx \bm M$ (which \emph{is} symmetric), then $(\bm M_\eps)_{:,t} \approx (\bm M_{\eps})_{t,:}$ and in practice, we use the column to represent the row; our experiments show that this does not incur noticeable performance drop.  To avoid pre-computing diagonal elements of $\bm M_\epsilon$, we estimate $\sum_{t=1}^n m_{t t} \approx k n ^2$; experiments show this works well in practice.

\paragraph{Dynamic setting.\quad}  An extension of our current setting is the dynamic setting, in which newly labeled nodes and their edges are dynamically added or deleted. As is, \textsc{FastONL} is well-suited to this extension;  the key idea is to use an efficient method to keep updating \textsc{FIFOPush}, which can quickly keep track of these kernel vectors \citep[e.g.]{zhang2016approximate}. The regret analysis of the dynamic setting is more challenging, and we will consider it as future work.

\section{Experiments}
\label{section:experiments}

In this section, we conduct extensive experiments on the online node classification for different-sized graphs and compare \textsc{FastONL} with baselines. We address the following questions: \textit{1) Do these parameterized kernels work and capture label smoothness?; 2) How does \textsc{FastONL} compare in terms of classification accuracy and run time with baselines?}

\textbf{Experimental setup.\quad} We collect ten graph datasets where nodes have true labels (Tab. \ref{tab:datasets}) and create one large-scale Wikipedia graph where chronologically-order node labels are from ten categories of English Wikipedia articles.  We consider four baselines, including 1) Weighted Majority (\textsc{WM}), where we predict each node $u$ by its previously labeled neighbors (a purely local but strong baseline described in Appendix \ref{sect:appendix:experiments}); 2) \textsc{Relaxation} \cite{rakhlin2017efficient}, a globally guaranteed method; 3) Weighted Tree Algorithm (WTA) \cite{cesa2013random}, a representative method based on sampling random spanning trees;\footnote{We note that the performance of WTA is competitive to, sometimes outperforms  \textsc{Perceptron}-based methods \cite{herbster2005online}.} and 4) \textsc{Approximate}, the power iteration-based method defined by Equ. \eqref{inequ:approximate-method}. We implemented these baselines using Python. For \textsc{FastONL}, we chose the first two kernels defined in Tab. \ref{tab:graph-kernel-presentation} and named them as \textsc{FastONL}-$\bm K_1$ and \textsc{FastONL}-$\bm K_2$, respectively.  All experimental setups, including parameter tuning, are further discussed in Appendix \ref{sect:appendix:experiments}.

\begin{figure}[H]
\centering
\includegraphics[width=7.5cm]{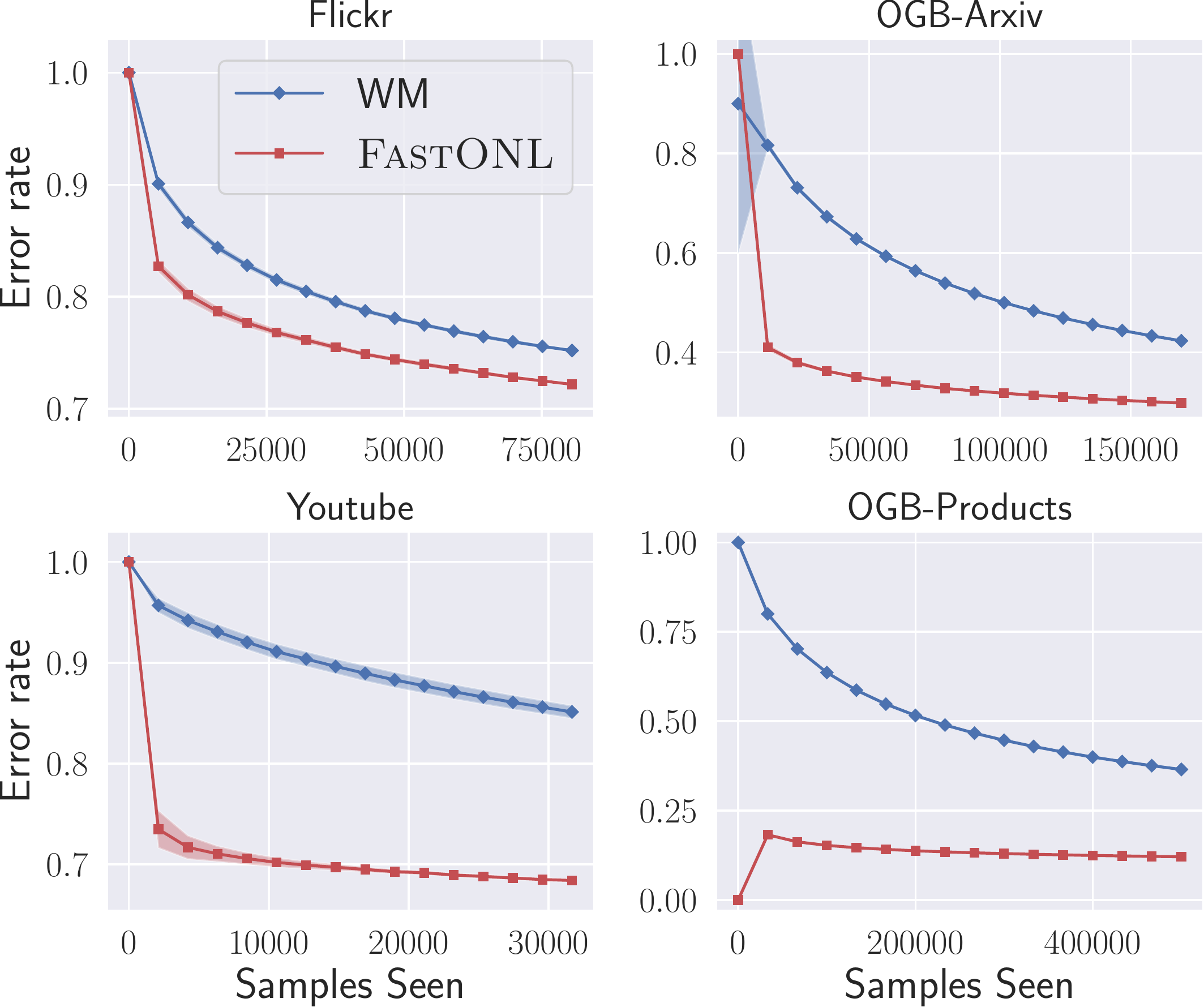}
\caption{The comparison of error rate of \textsc{FastONL} and \textsc{WM} on middle-scale graphs.}
\label{fig:middle-scale-graph}
\end{figure}

\begin{table*}[t!]
\caption{Run time of online node labeling methods over six graphs (seconds) averaged over 10 trials.}
\begin{center}
\begin{tabular}{p{2.7cm} p{1.5cm}p{1.5cm}p{1.4cm}p{2cm}p{2cm}p{2cm}}
 \toprule
 & Political & Citeseer & Cora & Pubmed & MNIST & Blogcatalog \\ [0.5ex] 
 \hline
 \textsc{WM} & 0.01 & 0.01 & 0.01 & 0.08 & 0.09 & 0.05\\
\textsc{WTA} & 66.61 & 146.97 & 213.00 & 2177.49 & 10726.67 & 5108.45 \\
\textsc{Approximate} & 1.47 & 0.66 & 0.97 & 159.48 & 43.83 & 68.52 \\
\textsc{Relaxation} &  0.78 & 1.66 & 2.94 & 122.45 & 976.69 & 154.32\\
\textsc{FastONL}-$\bm K_1$ & 1.12 & 1.10 & 1.73 & 4.86 & 22.42 & 22.14\\
\textsc{FastONL}-$\bm K_2$ & 1.21 & 1.12 & 2.57 & 7.27 & 33.00 & 12.03\\
 \bottomrule
\end{tabular}
\label{tab:run-time}
\end{center}
\end{table*}

\begin{figure}[H]
\centering
\includegraphics[width=7cm]{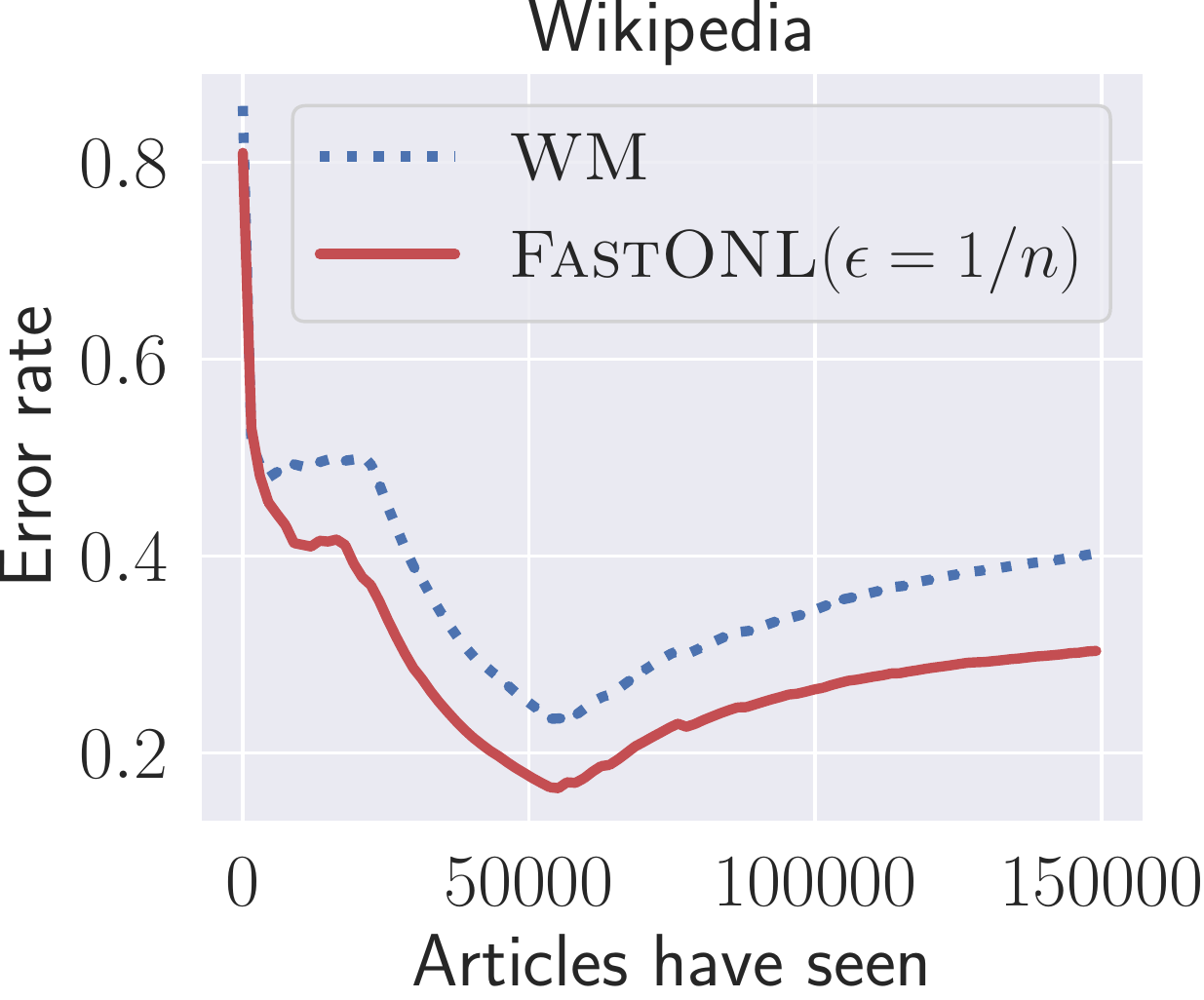}
\caption{The comparison of error rate of \textsc{FastONL} and \textsc{WM} on one large-scale graph.}
\label{fig:en-wikipedia}
\end{figure}

\textbf{Online node labeling performance.\quad} The online labeling error rates over 10 trials on small-scale graphs are presented in Fig. \ref{fig:fig-error-rate-small-graphs} where we pick $\eps=10^{-5}$. As we can see from Fig. \ref{fig:fig-error-rate-small-graphs}, the approximated performance is almost the same as of \textsc{Relaxation} but with great improvements on runtime as shown in Tab. \ref{tab:run-time}. The \textsc{WM} can be treated as a strong baseline. For middle and large-scale graphs, matrix inversion is infeasible, and these baselines are unavailable. We compare \textsc{FastONL} with the local method \textsc{WM}. Fig. \ref{fig:middle-scale-graph} and \ref{fig:en-wikipedia} present the error rate as a function of the number of seen nodes. \textsc{FastONL} outperforms the local \textsc{WM} by a large margin. This indicates that \textsc{FastONL} has a better tradeoff between local and global consistency.  The average run time of these methods is presented in Tab. \ref{tab:run-time}. The local method \textsc{WM} has the lowest run time per iteration. Our method is between \textsc{Relaxation} and \textsc{WTA}.

\begin{figure}[H]
\centering
\includegraphics[width=.47\textwidth]{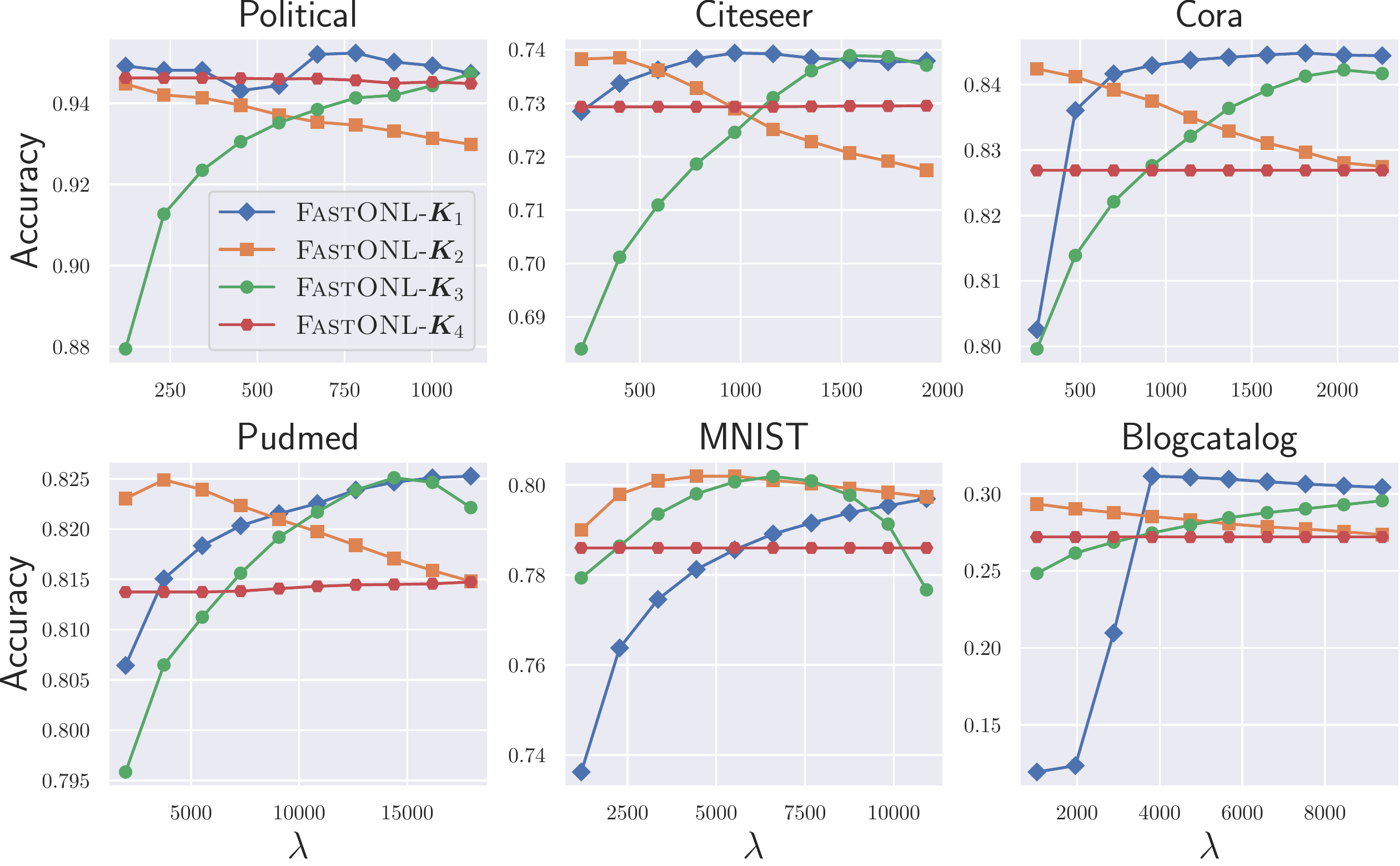}
\caption{The accuracy of applying the first four kernel matrices for \textsc{FastONL} on six small graphs.\vspace{-8mm}}
\label{fig:parameter-tuning}
\end{figure}

\textbf{Performance of parameterized kernels.\quad} In our theory, we showed what the effect of parameters $(\lambda, \beta)$ is on the regret (see Thm.\ref{thm:regret-analysis-relaxation}). The parameter $\lambda$ is a label smoothing parameter controlling the range of allowed label configurations while $\beta$ is the kernel parameter. We tested the first four kernels where kernels $\bm{K}_1$ and $\bm{K}_2$ solely depend on $\lambda$, while kernels $\bm{K}_3$ and $\bm{K}_4$ involve both $\lambda$ and $\beta$. However, for $\bm{K}_3, \beta$ is defined as $\lambda / n$, and for $\bm{K}_4$, it is defined as $\beta=1-\lambda / n$, as established in Thm.\ref{thm:regret-analysis-relaxation}. By defining $\beta$ this way, our theorem ensures an effective regret.
We experimented with various values of $\lambda$, selecting from $0.1 \cdot n, 0.2 \cdot n, \ldots, 0.9 \cdot n, n$. Fig. \ref{fig:parameter-tuning} shows how different kernels perform over different graphs.  All of the kernels successfully captured label smoothing but exhibited differing performances with varying $\lambda$. We consider the first four kernels as listed in Tab. \ref{tab:graph-kernel-presentation}, sweeping $\lambda$. To answer our first question, we find that \textit{all kernels can capture the label smoothing well but perform differently with different $\lambda$.} Overall, the normalized kernel of $\bm K_2$ enjoys a large range of $\lambda$, while $\bm K_1$ and $\bm K_3$ tend to prefer big $\lambda$.

\textbf{Case study of labeling Wikipedia articles.\quad} We apply our method to a real-world Wikipedia graph, which contains 6,216,199 nodes where corresponding labels appear chronically and unweighted 177,862,656 edges (edges are hyperlinks among these Wikipedia articles). Each node may have a label (downloaded from \citet{dbpedialink}, about 50\% percentage of nodes have labels, we use the first 150,000 labeled nodes) belonging to ten categories describing the Wikipedia articles, such as people, places, etc. Fig. \ref{fig:en-wikipedia} presents our results on this large-scale graph. Compared with the strong baseline \textsc{WM}, our \textsc{FastONL} truly outperforms it by a large margin with only about 0.3 seconds for each article.

\section{Conclusion}

We study the online relaxation framework for the node labeling problem. We propose, for the first time, a fast approximate method that yields effective online regret bounds, filling a significant gap in the theoretical analysis of this online learning problem. We then design a general \textsc{FIFOPush} algorithm to quickly compute an approximate column of the kernel matrix in an online fashion that does not require large local memory storage. Therefore, the actual computational complexity per-iteration is truly local and competitive to other baseline methods. The local analysis of \textsc{FIFOPush} is challenging when the acceleration is added to the algorithm. It is interesting to see if there is any local analysis for accelerated algorithms (See the open question \cite{fountoulakis2022open}). It is also interesting to see whether our work can be extended to directed or dynamic graph settings.

\clearpage
\bibliography{references}
\bibliographystyle{icml2023}

\input{appendix.tex}

\end{document}

%% file: appendix.tex
\newpage
\appendix
\onecolumn
The appendix is organized as follows: 
Section \ref{sect:appendix:proofs} presents all missing proofs for graph kernel matrices computation and approximation. Section \ref{sect:appendix:eig-evaluation} proves the bounds of eigenvalues of $\bm M_\eps$ and $\bm R_\eps$. Section \ref{sect:appendix:regret-bound}  provides the regret analysis. Section \ref{sect:appendix:experiments} presents more experimental details and results.

\section{Proofs}
\label{sect:appendix:proofs}
\subsection{Graph kernel matrices and their equivalence: The proof of Thm. \ref{thm:basic-kernel-representation}}

Recall that our goal is to compute the following augmented kernel matrix
\begin{equation}
\bm M_{\lambda,\beta} = \left( \frac{{\bm K}_\beta^{-1}}{2\lambda} + \frac{\bm I_n}{2n} \right)^{-1}  \nonumber
\end{equation}
for various instances of $\bm K_\beta^{-1}$ as listed in Table \ref{tab:graph-kernel-presentation}. The
unnormalized Laplacian is defined as $\mc{\bm L} \triangleq \bm D - \bm W$ where $\bm W$ is the nonnegative symmetric weighted adjacency matrix, and $\bm D$ is the corresponding weighted degree matrix defined as $\diag{(\bm W \bm 1)}$. The normalized graph Laplacian is $\bm L \triangleq {\bm D}^{-1/2} \mc{\bm L} {\bm D}^{-1/2}$. Using \textsc{FIFOPush}, we can approximate the following two basic matrices
\begin{align}
\text{Type-}\bm L: \quad \bm X_{\bm L} =  \alpha \left(\bm I_n - (1-\alpha) \bm W \bm D^{-1}\right)^{-1}, \quad\quad\quad \text{Type-}\mc L: \quad \bm X_{\mc{L}} = \left(\alpha \bm I_n + \bm D - \bm W\right)^{-1}.
\end{align}
We repeat our theorem in the following
\begin{reptheorem}{thm:basic-kernel-representation}
Let $\bm K_\beta^{-1}$ be the inverse of the symmetric positive definite kernel matrix defined in Table \ref{tab:graph-kernel-presentation}. Then $\bm M_{\lambda,\beta}$ can be decomposed into $\bm M_{\lambda,\beta} = a \bm A^{-1}\bm X \bm B$, which is easily computed once $\bm X$ available. $\bm X$ represents two basic kernel inverse
\begin{align*}
\bm X_{\mc L} = \left(\alpha \bm I + \mc L\right)^{-1},\quad \bm X_{\bm L} = \alpha \left(\bm I - (1-\alpha) \bm W \bm D^{-1}\right)^{-1}
\end{align*}
corresponding to inverse of regularized $\bm L$ and $\mc L$, respectively.
\end{reptheorem}
\begin{proof}
We now show how each of these kernels $\bm K$ can be efficiently computed given either $\bm X_{\bm L}$ or $\bm X_{\mc{L}}$ as follows:
\paragraph{Instance 1.\quad} For the kernel $\bm K_\beta^{-1} = \mc{\bm L}$, we have
\begin{align*}
\left(\frac{\bm K_\beta^{-1}}{2\lambda} + \frac{\bm I_n}{2 n} \right)^{-1} &= \left(\frac{\mc{\bm L}}{2\lambda} + \frac{\bm I_n}{2 n} \right)^{-1} \\
&= \left(\frac{\bm D - \bm W}{2\lambda} + \frac{\bm I_n}{2 n} \right)^{-1} \\
&= 2 \lambda \left( \alpha \bm I_n + \bm D - \bm W  \right)^{-1} \\
&= 2\lambda \bm X_{\mc {\bm L}},
\end{align*}
where we let $\alpha = \tfrac{\lambda}{n}$.

\paragraph{Instance 2.\quad} For the kernel $\bm K_{\beta}^{-1} = \bm I - \bm D^{-1/2}\bm W {\bm D}^{-1/2}$, we have
\begin{align*}
\left(\frac{\bm K_\beta^{-1}}{2\lambda} + \frac{\bm I_n}{2 n} \right)^{-1} &= \left(\frac{\bm I_n - {\bm D}^{-1/2} {\bm W} {\bm D}^{-1/2} }{2\lambda} + \frac{\bm I_n}{2 n} \right)^{-1} \\
&= \left( \frac{1}{2 \lambda} + \frac{1}{2 n} \right)^{-1}\cdot \left( \bm I_n - \frac{\left(\frac{1}{2 \lambda} + \frac{1}{2 n}\right)^{-1}}{2\lambda} {\bm D}^{-1/2} {\bm W} {\bm D}^{-1/2}  \right)^{-1} \\
&= \frac{2n\lambda}{n + \lambda }{\bm D}^{-1/2}\left(\bm I_n - \frac{n}{n + \lambda}  \bm W {\bm D}^{-1}\right)^{-1} {\bm D}^{1/2} \\
&= 2n \alpha {\bm D}^{-1/2} \left(\bm I_n - (1-\alpha) \bm W {\bm D}^{-1}\right)^{-1} {\bm D}^{1/2} \\
&= 2n {\bm D}^{-1/2} \bm X_{\bm L} {\bm D}^{1/2},
\end{align*}
where $\alpha = \frac{\lambda}{n + \lambda }$.

\paragraph{Instance 3.\quad} The kernel $\bm K_{\beta}^{-1} = \bm I - \beta \bm D^{-1/2}\bm W {\bm D}^{-1/2}$. Different from \textbf{Instance 2}, the kernel is now parameterized. Specifically, we have
\begin{align*}
\left(\frac{\bm K_\beta^{-1}}{2\lambda} + \frac{\bm I_n}{2 n} \right)^{-1} &= \left(\frac{\bm I_n - \beta {\bm D}^{-1/2} {\bm W} {\bm D}^{-1/2} }{2\lambda} + \frac{\bm I_n}{2 n} \right)^{-1} \\
&= \left( \frac{1}{2 \lambda} + \frac{1}{2 n} \right)^{-1}\cdot \left( \bm I_n - \frac{\beta\left(\frac{1}{2 \lambda} + \frac{1}{2 n}\right)^{-1}}{2\lambda} {\bm D}^{-1/2} {\bm W} {\bm D}^{-1/2}  \right)^{-1} \\
&= \frac{2n\lambda}{n + \lambda }{\bm D}^{-1/2}\left(\bm I_n - \frac{\beta n}{n + \lambda}  \bm W {\bm D}^{-1}\right)^{-1} {\bm D}^{1/2} \\
&= \frac{2 n \lambda}{(n+\lambda)\alpha}  {\bm D}^{-1/2} \alpha\left(\bm I_n - (1-\alpha) \bm W {\bm D}^{-1}\right)^{-1} {\bm D}^{1/2} = \frac{2\lambda n}{ n + \lambda - n \beta } {\bm D}^{-1/2} \bm X_{\bm L} {\bm D}^{1/2},
\end{align*}
where $1-\alpha = \frac{n \beta}{n+\lambda}$, and the parameter $\beta \in (0,\frac{n+\lambda}{n})$. 
\paragraph{Instance 4.\quad} The kernel $\bm K_\beta^{-1} = \beta {\bm I} + {\bm S}^{-1/2} \mc{\bm L} {\bm S}^{-1/2}$ was initially considered in \citet{johnson2008graph} where $\bm S$ is a positive diagonal matrix. Typical examples of $\bm S$ could be $\bm D$, $\bm I$, etc. Note that
\begin{align*}
\left(\frac{\bm K_\beta^{-1}}{2\lambda} + \frac{\bm I_n}{2 n} \right)^{-1} &= \left(\frac{\beta {\bm I} + {\bm S}^{-1/2} {\mc L} {\bm S}^{-1/2}}{2\lambda} + \frac{\bm I_n}{2 n} \right)^{-1} \\ 
&= \left(\frac{1}{b}\left(a b {\bm I} + {\bm S}^{-1/2} (\bm D - \bm W) {\bm S}^{-1/2}\right)\right)^{-1} \text{\quad\quad // let $a = \left( \frac{\beta}{2\lambda} + \frac{1}{2 n}\right), b = 2 \lambda$}\\
&= b {\bm S}^{1/2}\left( a b {\bm S} + \bm D - \bm W \right)^{-1}{\bm S}^{1/2}  \\
&= b {\bm S}^{-1/2}\left( a b {\bm I} + \left(\bm D - \bm W\right)\bm S^{-1} \right)^{-1}{\bm S}^{1/2} \\
&= b {\bm S}^{-1/2}\left( a b {\bm I} + \left(\bm D - \bm W\right)\bm S^{-1} \right)^{-1}{\bm S}^{1/2} \\
&= 2\lambda {\bm S}^{-1/2}\bm X_{\mc L}{\bm S}^{1/2},
\end{align*}
where $\alpha = \frac{\beta n + \lambda}{n}$. Note $\left(\bm D - \bm W\right)\bm S^{-1}$ is a positive semidefinite matrix as $\bm D - \bm W$ is positive semidefinite and $\bm S^{-1}$ is positive definite, then applying $\bm D^{'} = \bm D \bm S^{-1}$ and $\bm W^{'} = \bm W \bm S^{-1}$. Therefore, it is essential to solve $(\alpha \bm I + \bm D' -\bm W')^{-1}$ because $\bm D' - \bm W'$ has nonnegative eigenvalues and $D'_u =\sum_{v \in \nei{(u)}} W'_{u v}$ for all $u \in \mathcal{V}$. Therefore, it belongs to Type-II. Here, we abuse notations where we let $\bm D = \bm D'$ and $\bm W = \bm W'$.

\paragraph{Instance 5.\quad} The normalized kernel matrix $\bm K_\beta^{-1} = {\bm S}^{-1/2} {\left(\beta {\bm I} + \mc L  \right)} {\bm S}^{-1/2}$, the $\bm S$ is the normalization matrix \citep{ando2006learning,johnson2007effectiveness}. Just like \textbf{Instance 4}, we could have different choices for $\bm S$: $\bm I$, $\bm D$, or $(\beta \bm I + \mc L )^{-1}$ (for the last case, the kernel is then normalized with unit diagonal), etc. Note
\begin{align*}
\left(\frac{\bm K_\beta^{-1}}{2\lambda} + \frac{\bm I_n}{2 n} \right)^{-1} &= \left(\frac{{\bm S}^{-1/2} ( \beta {\bm I} + \mc L ) {\bm S}^{-1/2}}{2\lambda} + \frac{\bm I_n}{2 n} \right)^{-1} \\
&= {\bm S}^{1/2} \left(\Tilde{\bm S} + \frac{\mc L }{2\lambda}\right)^{-1} {\bm S}^{1/2} \text{\quad\quad\quad // let $\Tilde{\bm S} = \frac{\beta \bm I}{2\lambda} + \frac{\bm S}{2n}$}\\
&= 2\lambda {\bm S}^{1/2} \left( 2\lambda \Tilde{\bm S} + \mc L \right)^{-1} {\bm S}^{1/2} \\
&= 2\lambda {\bm S}^{1/2} \Tilde{\bm S}^{-1} \left( 2\lambda \bm I + (\bm D - \bm W)\Tilde{\bm S}^{-1}\right)^{-1} {\bm S}^{1/2},
\end{align*}
where $(\bm D - \bm W)\Tilde{\bm S}^{-1}$ is a transformed version of $\mc{L}$. We continue to have
\begin{align*}
\left(\frac{\bm K_\beta^{-1}}{2\lambda} + \frac{\bm I_n}{2 n} \right)^{-1} &= 2\lambda {\bm S}^{1/2} \Tilde{\bm S}^{-1} \left( 2\lambda \bm I + (\bm D - \bm W)\Tilde{\bm S}^{-1}\right)^{-1} {\bm S}^{1/2} \\
&= 2\lambda {\bm S}^{1/2} {\left(\frac{\beta \bm I}{2\lambda} + \frac{\bm S}{2n}\right)}^{-1} \left( 2\lambda \bm I + (\bm D - \bm W)\Tilde{\bm S}^{-1}\right)^{-1} {\bm S}^{1/2} \\
&= \left( \frac{\bm S^{1/2}}{4 n\lambda} + \frac{\beta\bm S^{-1/2}}{4\lambda^2} \right)^{-1} \bm X_{\mc L} {\bm S}^{1/2},
\end{align*}
where $\alpha = 2\lambda$. Similar to \textbf{Instance 4}, $(\bm D - \bm W)\Tilde{\bm S}^{-1}$ is transformed Laplacian matrix. Let $\bm D' = \bm D \Tilde{\bm S}^{-1}$ and $\bm W' = \bm W \Tilde{\bm S}^{-1}$. We abuse of notations of $\bm D, \bm W$ and let $\bm D = \bm D'$ and $\bm W = \bm W'$.

\textbf{Instance 6.\quad} The augmented kernel $\bm K_{\lambda} = \mathcal{\bm L} + b\cdot \bm 1\bm 1^\top + \beta \bm I$ \cite{herbster2005online} can be reformulated as 
\begin{align*}
\left(\frac{\bm K_\beta^{-1}}{2\lambda} + \frac{\bm I_n}{2 n} \right)^{-1} &= \left( \frac{\mc{\bm L} + b\cdot \bm 1 \bm 1^\top + \beta \bm I}{2\lambda} + \frac{\bm I}{2 n} \right)^{-1} \\
&= \left( \frac{\mc{\bm L} + \beta \bm I}{2\lambda} + \frac{\bm I}{2 n}  + \frac{ b\cdot \bm 1 \bm 1^\top}{2\lambda}\right)^{-1} \\
&= 2\lambda \bm X_{\mc L}\left( \bm I - \frac{ b \bm 1\bm 1^\top \bm X_{\mc L}}{1 + b \bm 1^\top \bm X_{\mc{L}} \bm 1}\right),
\end{align*}
where $\alpha = \beta + \frac{\lambda}{n}$ and we use the similar method as in \textbf{Instance 4} but plus a rank one matrix $\bm 1 \bm 1^\top$. The last equality is from the fact that, for any invertible matrix $\bm X$, by the Sherman–Morrison formula, we have
\begin{equation}
(\bm X + \bm 1 \bm 1^\top)^{-1} = \bm X^{-1} - \frac{\bm X^{-1} \bm 1 \bm 1^\top \bm X^{-1}}{1+ \bm 1^\top \bm X^{-1} \bm 1}. \nonumber
\end{equation}
Furthermore, note the summation of each column of $(\alpha \bm I + \mathcal{L})^{-1}$ is a constant $1/\alpha$, then $\bm 1^\top \bm X_{\mc{L}} \bm 1 = n/\alpha$. Then, we continue to have
\begin{align*}
\left(\frac{\bm K_\beta^{-1}}{2\lambda} + \frac{\bm I_n}{2 n} \right)^{-1} &= 2\lambda \bm X_{\mc L}\left( \bm I - \frac{ b \bm 1\bm 1^\top \bm X_{\mc L}}{1 + b \bm 1^\top \bm X_{\mc{L}} \bm 1}\right) \\
&= 2\lambda \bm X_{\mc L} \left( \bm I - \frac{ \alpha b \bm 1\bm 1^\top \bm X_{\mc L}}{\alpha + n b} \right) \\
&= 2\lambda \bm X_{\mc L} \left( \bm I - \frac{b \bm 1\bm 1^\top}{\alpha + n b} \right),
\end{align*}
where note that $\bm 1\bm 1^\top \bm X_{\mc L} = \bm 1\bm 1^\top/\alpha$.
\end{proof}

\subsection{Local linear convergence of \textsc{FIFOPush} for $\alpha\left(\bm I - (1-\alpha) \bm W \bm D^{-1}\right)^{-1}\bm e_s$: The proof of Thm. \ref{thm:local-convergence}}
\label{sect:appendix:local-convergence-proof}

Given any $\alpha \in (0,1), \epsilon > 0$, Algorithm \ref{algo:dummy-fifo-push-1} is to approximate $\alpha\left(\bm I - (1-\alpha) \bm W \bm D^{-1}\right)^{-1}\bm e_s$. Before the proof of Theorem \ref{thm:local-convergence}, we provide an equivalent version of \textsc{FIFOPush} as presented in Algorithm \ref{algo:dummy-fifo-push-1} where time index $t$ of $\bm r, \bm x$ and time index $t^\prime$ of processed nodes $u$ are added. Our proof is based on this equivalent version. Compared with Algorithm \ref{algo:fifo-push}, the only difference is that we added a dummy node $\red{\ddag}$. Still, Algorithm \ref{algo:dummy-fifo-push-1} is essentially the same as Algorithm \ref{algo:fifo-push}. The chronological order of processed nodes $u_{t'}$ by \textsc{FIFOPush} can then be represented as the following order

\begin{algorithm}
\caption{$\textsc{FIFOPush}(\mc{G},\eps,\alpha, s)$\cite{andersen2006local} with a dummy node $\red{\ddag}$}
\begin{algorithmic}[1]
\STATE \textbf{Initialize}: $\bm r_t = \bm e_s, \ \ \bm x_t = \bm 0, \ \ t= 1, \ \ t^\prime = 1$
\STATE $\mc{Q}=[s, \red{\ddag}]$ \quad\quad\quad\quad\quad\quad\quad\ \ // At the initial stage, $\mc{Q}$ contains $s$ and a dummy node $\red{\ddag}$.
\WHILE{$\mc{Q}\text{.size}()\ne 1$}
\STATE $u_{t'} = \mc{Q}\text{.pop()}$
\IF{$u_{t'} == \red{\ddag}$}
\STATE $t = t + 1$ \quad\quad\quad\quad\quad\quad // Nodes in $\mc{U}_t$ has been processed. Go to the next epoch.
\STATE $\mc{Q}\text{.push}(u_{t'})$
\STATE \textbf{continue}
\ENDIF
\IF{$r_{u_{t'}} < \eps \cdot d_{u_{t'}}$}
\STATE $t^\prime = t^\prime + 1$ \quad\quad\quad\quad\quad\ \ // $u_{t'}$ is an ``\textit{inactive}'' node
\STATE \textbf{continue}
\ENDIF
\STATE $x_{u_{t'}} = x_{u_{t'}} + \alpha r_{u_{t'}}$ 
  \quad\quad\quad // $u_{t'}$ is an ``\textit{active}'' node
\FOR{$v \in \nei(u_{t'})$}
\STATE $r_{v} = r_{v} + \frac{(1-\alpha) r_{u_{t'}}}{D_{u_{t'}}} \cdot w_{u_{t'} v}$
\IF{$v \notin \mc{Q}$}
\STATE $\mc{Q}\text{.push}(v)$
\ENDIF
\ENDFOR
\STATE $r_{u_{t'}} = 0$ 
\STATE $t^\prime = t^\prime + 1$
\ENDWHILE
\STATE \textbf{Return} $(\bm x_t,\bm r_t)$
\end{algorithmic}
\label{algo:dummy-fifo-push-1}
\end{algorithm}

\begin{equation}
\ \red{\ddag} \ \underbrace{u_1}_{\mc{U}_1} \ \red{\ddag} \ \underbrace{u_2, u_3, \ldots}_{\mc{U}_2} \ \red{\ddag} \ \cdots \ \red{\ddag} \ \underbrace{u_{t^{\prime}}, u_{t^{\prime}+1}, u_{t^{\prime}+2}, \ldots, u_{t^{\prime}+i}}_{\mc{U}_t} \ \red{\ddag} \ \cdots
\label{equ:super-epoch-fifo-push},
\end{equation}

where $\mc{U}_t = \{u_{t'},u_{t^\prime+1},\ldots,u_{t^\prime+i}\}$ is the set of nodes processed in $t$-th epoch. Hence, \eqref{equ:super-epoch-fifo-push} defines super epochs indexed by $t$ where $\mc{U}_t$ will be processed. From epoch $t$, new nodes will be added into $\mc{Q}$ for the next epoch as illustrated in Fig. \ref{fig:active-queue-nodes}. $\mc{U}_t$ contains: 1) a set of \textit{active} nodes $\mc{S}_t \triangleq \left\{ u_{t'} : r_{u_{t'}} \geq \eps \cdot d_{u_{t'}}, u_{t'} \in \mc{U}_t\right\}$; and 2) a set of \textit{inactive} nodes $\left\{ u_{t'} : 0 < r_{u_{t'}} < \eps \cdot d_{u_{t'}}, u_{t'} \in \mc{U}_t\right\} = \mc{U}_t \backslash \mc{S}_t$.\footnote{We say a node $u_{t'}$ is \textit{active} if $r_{u_{t'}} \geq \eps d_{u_{t'}}$ and \textit{inactive} if $0\leq r_{u_{t'}} < \eps d_{u_{t'}}$.} \textsc{FIFOPush} terminates only when $\mc{Q}$ contains the dummy node $\red{\ddag}$.

Define $\bm X = \alpha\left(\bm I - (1-\alpha) \bm W \bm D^{-1}\right)^{-1}$. Denote the estimation matrix $\bm X_{\eps} = \left[\bm x_{1,\eps}, \ldots, \bm x_{n,\eps}\right]$ and the residual matrix $\bm R_{\eps} = \left[\bm r_{1,\eps}, \ldots, \bm r_{n,\eps}\right]$ where $(\bm x_{s,\eps}, \bm r_{s,\eps}) = \textsc{FIFOPush}(\mc{G},\eps,\alpha,s)$ for all $s\in \mc{V}$. The next lemma shows that $\bm X_\eps$ is a good approximation of $\bm X$ from the bottom when $\eps$ is small.

\begin{lemma}
Let $\bm X = \alpha\left(\bm I - (1-\alpha) \bm W \bm D^{-1}\right)^{-1}$ and denote $s$-th column of $\bm X$  as $\bm x_s = \alpha\left(\bm I - (1-\alpha) \bm W \bm D^{-1}\right)^{-1} {\bm e}_s$. Let $({\bm x}_{s,\eps}, \bm r_{s,\eps}) = \textsc{FIFOPush}(\mc{G},\eps,\alpha,s)$ be the pair of vectors returned by Alg. \ref{algo:dummy-fifo-push-1} where $\bm x_{s,\eps}$ is an estimate of $\bm x_s$ and $\bm r_{s,\eps}$ is the corresponding residual vector. Denote the estimation matrix $\bm X_{\eps} = \left[\bm x_{1,\eps}, \ldots, \bm x_{n,\eps}\right]$ and the residual matrix $\bm R_{\eps} = \left[\bm r_{1,\eps}, \ldots, \bm r_{n,\eps}\right]$ by calling \textsc{FIFOPush} for all $s\in \mc{V}$. For any $\eps > 0$, we have
\begin{equation}
\bm X = \bm X_\eps + \bm X \bm R_\eps, \nonumber
\end{equation}
where $\bm R_\eps$ satisfies $\bm 0_{n\times n} \leq \bm R_\eps < \eps \cdot \diag{(d_1,\ldots,d_n)} \cdot \bm 1 \bm 1^\top$.
\label{lemma:A-1}
\end{lemma}
\begin{proof}
Let us assume $t = t'$ at the beginning of $t$-th epoch. During the $t$-th epoch, \textsc{FIFOPush} updates $t$ to $t+1$ and updates $t'$  from $t'= t_0', t_1',t_2',\ldots$ to $t_{|\mc{S}_t|}' = t+1$ where $t_i'$ is the time after the update of node $u_{t_i'}$. Recall $\mc{S}_t$ is the set of processed active nodes (at the beginning of $t$-th epoch, we do not know how many nodes in $\mc{S}_t$ since some inactive nodes could be active after some push operations).
For each active node $u_{t_i'}$ ($i=1,2,\ldots,|\mathcal{S}_t|$) at $t$-th epoch, we denote $\bm x_{u_{t_i'}}$ and $\bm r_{u_{t_i'}}$ as the updated vectors of $\bm x_t$ and $\bm r_t$, respectively. After all active nodes $u_{t_i'} \in \mc{S}_t$ have been processed, $\bm x$ is updated from $\bm x_t$ to $\bm x_{t+1}$ and $\bm r$ from $\bm r_t$ to $\bm r_{t+1}$ as the following
\begin{align}
\bm x_t &= \bm x_{u_{t_0'}} \xrightarrow[]{u_{t_1'}} {\bm x}_{t_1'}  \xrightarrow[]{u_{t_2'}} {\bm x}_{t_2'} \quad \cdots \quad \xrightarrow[] {u_{t_{|\mc{S}_t|}'}} {\bm x}_{t_{|\mc{S}_t|}'} = \bm x_{t+1} \nonumber\\
\bm r_t &= \bm r_{u_{t_0'}} \xrightarrow[]{u_{t_1'}} {\bm r}_{ u_{t_1'} }  \xrightarrow[]{u_{t_2'}} {\bm r}_{u_{t_2'}} \quad \cdots \quad \xrightarrow[]{u_{t_{|\mc{S}_t|}'}} {\bm r}_{u_{{|\mc{S}_t|}}'} = \bm r_{t+1}. \nonumber
\end{align}
For each $i$-th active node $u_{t_i'}$, the updates are from Line 14 to Line 21 of Alg. \ref{algo:dummy-fifo-push-1} give us the following iterations
\begin{align}
{\bm x}_{t_i'} &= \bm x_{t_{i-1}'}  
\underbrace{ + \alpha r_{u_{t_{i-1}'}} \cdot \bm e_{u_{t_{i-1}'}}}_{\text{Line 14}}\label{equ:forward-push-xx}\\
{\bm r}_{t_i'} &= \bm r_{t_{i-1}'}  \underbrace{+ (1-\alpha) r_{u_{t_{i-1}'}} \cdot \bm W \bm D^{-1} \bm e_{u_{t_{i-1}'}}}_{\text{Line 15,16}} \ \underbrace{- r_{u_{t_{i-1}'}}\cdot \bm e_{u_{t_{i-1}'}}}_{\text{Line 21}}. \label{equ:forward-push-rr}
\end{align}
These two iterations \eqref{equ:forward-push-xx} and \eqref{equ:forward-push-rr} essentially moves residual $r_{u_{t_{i-1}'}}$ out of node $u_{t_{i-1}'}$ to its estimate vector $\bm x$ and residual entries of its neighbors. Specifically, the first iteration \eqref{equ:forward-push-xx} moves $\alpha$ times magnitude of $r_{u_{t_{i-1}'}}$ to $\bm x$ and the second iteration \eqref{equ:forward-push-rr} moves $(1-\alpha) $ times of $r_{u_{t_{i-1}'}}$ to neighbors spread the magnitude by the distribution vector $\bm W \bm D^{-1} \bm e_{u_{t_{i-1}'}}$. The last term $- r_{u_{t_{i-1}'}}\cdot \bm e_{u_{t_{i-1}'}}$ is to remove $ r_{u_{t_{i-1}'}}$ from node $u_{t_{i-1}'}$. 
Rearrange \eqref{equ:forward-push-rr}, we have
\begin{align}
{\bm r}_{t_i'} &= \bm r_{t_{i-1}'} + (1-\alpha) r_{u_{t_{i-1}'}} \cdot \bm W \bm D^{-1} \bm e_{u_{t_{i-1}'}} - r_{u_{t_{i-1}'}}\cdot \bm e_{u_{t_{i-1}'}} \nonumber\\
{\bm r}_{t_i'} &= \bm r_{t_{i-1}'} - \left(\bm I - (1-\alpha) \bm W \bm D^{-1} \right) r_{u_{t_{i-1}'}} \cdot \bm e_{u_{t_{i-1}'}} \nonumber\\
\alpha r_{u_{t_{i-1}'}} \cdot \bm e_{u_{t_{i-1}'}} &= \alpha \left(\bm I - (1-\alpha) \bm W \bm D^{-1} \right)^{-1}\left( \bm r_{t_{i-1}'} - {\bm r}_{t_i'} \right). \label{equ:telescope}
\end{align}
Use \eqref{equ:telescope} and \eqref{equ:forward-push-xx}, we have
\begin{align*}
{\bm x}_{t_i'} &= \bm x_{t_{i-1}'}  
 + \alpha r_{u_{t_{i-1}'}} \cdot \bm e_{u_{t_{i-1}'}} \\
 &= \bm x_{t_{i-1}'}  
 + \alpha \left(\bm I - (1-\alpha) \bm W \bm D^{-1} \right)^{-1}\left( \bm r_{t_{i-1}'} - {\bm r}_{t_i'} \right)
\end{align*}
Since $i=1,2,\ldots, |\mathcal{S}_t|$, we sum above equation over all active nodes $\mathcal{S}_t$, we have
\begin{align*}
{\bm x}_{t+1} &= {\bm x}_{t_{|\mathcal{S}_t|'}} \\
&= \bm x_{t_{{|\mathcal{S}_t|-1}'}}  
 + \alpha \left(\bm I - (1-\alpha) \bm W \bm D^{-1} \right)^{-1}\left( \bm r_{t_{|\mathcal{S}_t|-1}'} - {\bm r}_{t_{|\mathcal{S}_t|}'} \right) \\
&\vdots \\
&= \bm x_{t_{0'}}  
 + \alpha \left(\bm I - (1-\alpha) \bm W \bm D^{-1} \right)^{-1} \sum_{i=1}^{|\mathcal{S}_t|}\left( \bm r_{t_{i-1}'} - {\bm r}_{t_{i}'} \right) \\
 &= \bm x_{t_{0'}}  
 + \alpha \left(\bm I - (1-\alpha) \bm W \bm D^{-1} \right)^{-1} \left( \bm r_{t_{0}'} - {\bm r}_{t_{|\mathcal{S}_t|}'} \right) \\
&= \bm x_{t}  
 + \alpha \left(\bm I - (1-\alpha) \bm W \bm D^{-1} \right)^{-1} \left( \bm r_{t} - {\bm r}_{t+1} \right)
\end{align*}

On the other hand, for all epochs, we continue to use the last equation to have
\begin{align}    
\bm x_{t+1} &= \bm x_t + \alpha \left(\bm I - (1-\alpha) \bm W \bm D^{-1}\right)^{-1} \left(\bm r_t - \bm r_{t+1}\right) \nonumber\\
&= \alpha\left(\bm I - (1-\alpha) \bm W \bm D^{-1}\right)^{-1} \sum_{i=1}^t\left(\bm r_i - \bm r_{i+1}\right) \nonumber\\
&=\alpha\left(\bm I - (1-\alpha) \bm W \bm D^{-1}\right)^{-1} \left(\bm r_1 - \bm r_{t+1}\right) \nonumber\\
&= \bm X (\bm e_s - \bm r_{t+1}). \label{equ:x-r-identity}
\end{align}
Since $\bm x_{s,\eps}$ is an estimate of $\bm x_s$ returned by \textsc{FIFOPush}, for any node $s\in \mc{V}$ and by \eqref{equ:x-r-identity}, we have
\begin{equation}
\bm X \bm e_s = {\bm x}_{s,\eps} + \bm X \bm r_{s,\eps}, \quad \forall s\in \mc{V}. \nonumber
\end{equation}
Write the above equation as a matrix form; we obtain $\bm X = \bm X_\eps + \bm X \bm R_\eps$. Notice that each $u$-th element of $\bm r_{s,\eps}$ satisfies $0 \leq r_{s,\eps}(u) < \eps d_u$. Hence we have ${\bm 0}_{n\times n} \leq \bm R_\eps < \eps \cdot \diag{(d_1,\ldots,d_n)}\cdot \bm 1\bm 1^\top$.
\end{proof}

\begin{remark}
The above lemma is essentially the linear invariant property \cite{andersen2006local}. Here, we show a relation between the estimation and residual vectors. Recall that for any subset of nodes $\mc{S}\subseteq \mc{V}$, the volume of $\mc{S}$ is defined $\vol{\mc{S}} = \sum_{v\in \mc{S}} d_v$. We are ready to prove Thm. \ref{thm:local-convergence}.
\end{remark}

\begin{reptheorem}{thm:local-convergence}[Local linear convergence of \textsc{FIFOPush} for $\bm X_{\bm L}$] Let  $\bm x_s = \bm X_{\bm L} \bm e_s$. Denote $T$ as the total epochs executed by \textsc{FIFOPush},  and $\mc{S}_t : = \{v: r_t(v) \geq \eps \cdot d_{v}, v \in \mc{I}_t\}$ as the set of active nodes in $t$-th epoch. Then, the total operations of $ \textsc{FIFOPush}(\mc{G},\eps,\alpha,s)$ is dominated by 
\begin{equation}
R_T := \sum_{t=1}^T\sum_{u_t\in {\mc S}_t} d_{u_t} \leq \frac{\vol{\mc{S}_{1:T}}}{\alpha \cdot \eta_{1:T}} \log\left( \frac{C_{\alpha,T}}{\eps} \right), \label{inequ:local-linear-bound-repeat}
\end{equation}
where $\vol{\mc{S}_{1:T}} = \sum_{t=1}^{T}\vol{\mc{S}_t} / T$ is the average volume of $\mc{S}_t$. Additionally, $\eta_{1:T} = \sum_{t=1}^{T} \eta_t / T$ is the average of local convergence factors $\eta_t \triangleq \sum_{u \in \mc{S}_t} d_u / \sum_{v \in \mc{I}_{t}} d_v$, and $C_{\alpha,T} = 1/(\sum_{v\in \mc{I}_T} (1-\alpha)d_u w_{u v} / D_u)$. For $s, i \in \mc{V}$, we have $\bm x_s = \bm x_{s,\eps} + \bm X_{\bm L} \bm r_{s,\eps}, r_{s,\eps}(i) \leq [0, \eps d_i)$.
\label{thm:local-convergence-repeat}
\end{reptheorem}

\begin{proof}[Proof of Theorem \ref{thm:local-convergence}]
From Lemma \ref{lemma:A-1}, we know that for $s, i \in \mc{V}$, we have $\bm x_s = \bm x_{s,\eps} + \bm X_{\bm L} \bm r_{s,\eps}, r_{s,\eps}(i) \leq [0, \eps d_i)$. At epoch $t\geq 1$, recall $\mc{Q}$ contains a set of \textit{active} nodes $\mc{S}_t$ and a set of \textit{inactive} nodes $\mc{U}_t \backslash \mc{S}_t$. After \textsc{FIFOPush} processed the last node $u_{t^\prime+i}$ in $\mc{U}_{t}$, it is easy to see that the total operations of $t$-th epoch are dominated by the volume of $\mc{S}_t$, i.e., $\vol{\mc S_t}$ (from Line 13 to Line 16). Hence, the total time complexity is dominated by $R_T = \sum_{t=1}^{T} \vol{\mc S_t}$. In the rest, we shall provide two upper bounds of $R_T$. 
\begin{enumerate}
\item The first is to prove an upper bound $1 / \alpha \eps$, which is directly followed from \citet{andersen2006local}. We repeat the main idea here. For each active iteration of Algo. \ref{algo:dummy-fifo-push-1}, we have $r_{u_{t'}} \geq \eps \cdot d_{u_{t'}}$, which indicates $r_{u_{t'}}$ was at least $\eps \cdot d_{u_{t'}}$; hence $\|\bm r_t\|_1$ decreased by at least $\alpha \eps\cdot d_{u_{t'}}$ with total $d_{u_{t'}}$ operations. Hence, overall $u_{t'} \in \mc{S}_t$, we have 
\[
\alpha \eps \sum_{u_{t'} \in \mc{S}_t} d_{u_{t'}} \leq \alpha \sum_{u_{t'} \in \mc{S}_t} r_{u_{t'}} = \|\bm r_t\|_1 - \|\bm r_{t+1}\|_1. 
\]
Summing the above inequality over $t$, we have the total operations of \textsc{FIFOPush} bounded by 
\begin{equation}
R_T = \sum_{t=1}^{T} \vol{\mc S_t} = \sum_{t=1}^{T} \sum_{u \in \mc{S}_{t}} d_u \leq \frac{1}{\alpha \eps} \sum_{t=1}^{T} \left( \|\bm r_t\|_1 - \|\bm r_{t+1}\|_1 \right)  = \frac{1-\|\bm r_{T+1}\|_1}{\alpha \eps} \leq \frac{1}{\alpha \eps}, \label{inequ:upper-bound-1}
\end{equation}
where note that $\|\bm r_1\|_1 = 1$. 

\item The sublinear bound $\mc{O}(\tfrac{1}{\alpha\eps})$ in \eqref{inequ:upper-bound-1} is independent of the graph size which is the key advantage of \textsc{FIFOPush} over other numerical methods such as Power Iteration where $\mc{O}(\tfrac{m}{\alpha}\log(\tfrac{1}{\eps}))$ operations needed, where $m$ is the number of edges in the graph. However, when $\eps$ becomes small, $\mc{O}(\tfrac{1}{\alpha\eps})$ is too pessimistic. It is natural to ask whether there exists any bound that takes advantage of both \textsc{FIFOPush} and \textsc{PowerIter}. We answer this question positively by providing a local linear convergence. There are two key components in our proof: 1) residuals left in $\bm r_T$ are relatively significant so that total epochs $T$ can be bound by $\mc{O}(\log \tfrac{1}{\eps})$; 2) the average operations of $t$ epochs is equal to $\vol{\mc S_{1:T}}$, which is independent of $m$.

After the $T$-th epoch finished, the set of all inactive nodes is exactly $\mc{I}_{T+1}$, i.e., $\mc{I}_{T+1} = \{v: 0< r_{T+1}(v) < \eps \cdot d_v, v\in \mc{V}\}$.  For each $v \in \mc{I}_{T+1}$, note that there exists at least one of its neighbor $u_{t'} \in \nei{(v)}$ such that $r_{u_{t'}} \geq \eps \cdot d_{u_{t'}}$ had happened in a previous active iteration. Combine with Line 12 of Algorithm \ref{algo:dummy-fifo-push-1}, we have
\begin{equation}
\forall v \in \mc{I}_{T+1},\quad \frac{(1-\alpha) r_u}{D_u} w_{u v} \geq \frac{(1-\alpha)\eps d_u}{D_u} w_{u v} \triangleq \tilde{r}_v, \label{equ:never-pushed-out-residual} \footnote{We ignore the time index $t^\prime$, which is unrelated with our analysis.}
\end{equation}
where $\tilde{r}_v$ is the residual pushed into $v$ but never popped out. From \eqref{equ:never-pushed-out-residual}, note that $\sum_{v\in \mc{I}_{T+1}} \tilde{r}_v$ is an estimate of $\|\bm r_{T+1}\|_1$ from bottom. That is
\begin{equation}
\sum_{v\in \mc{I}_{T+1}} \tilde{r}_v = \sum_{v\in \mc{I}_{T+1}} \frac{(1-\alpha)\eps d_u}{D_u} w_{u v} \leq \|\bm r_{T+1}\|_1. \label{inequ:lower-bound-r}    
\end{equation}
Next, we show a significant amount of residual that has been pushed out from $\bm r_{t}$ to $\bm r_{t + 1}$. For $t$-th epoch, the total amount of residual that had been pushed out is $\alpha \sum_{u_{t'} \in \mc{S}_t} r_{u_{t'}}$ (Line 10). That is,
\begin{equation}
\|\bm r_t\|_1 - \|\bm r_{t+1}\|_1 \geq \alpha \sum_{ u_{t'} \in \mc{S}_t} r_{u_{t'}}. \label{inequ:48}
\end{equation}
On the other hand, by the activation condition, we have
\begin{align*}
\forall u_{t'} \in \mc{S}_{t}, r_{u_{t'}} \geq \eps \cdot d_{u_{t'}}, \quad \quad \forall v \in \mc{I}_t \backslash \mc{S}_t, 0 < r_t(v) < \eps \cdot d_v.
\end{align*}
Summation above inequalities over all active nodes $u_{t'}$ and inactive nodes $v$, we have
\begin{equation}
\frac{\sum_{u_{t'} \in \mc{S}_{t}} r_{u_{t'}} }{\sum_{u_{t'} \in \mc{S}_{t}} d_{u_{t'}} } \geq \eps > \frac{\sum_{v \in \mc{I}_t \backslash \mc{S}_t} r_v}{\sum_{v \in \mc{I}_t \backslash \mc{S}_t} d_v }, \nonumber
\end{equation}
which indicates
\begin{equation}
\frac{\sum_{u_{t'} \in \mc{S}_t} r_{u_{t'}} }{\sum_{u_{t'} \in \mc{S}_t} d_{u_{t'}} } > \frac{\sum_{u_{t'} \in \mc{S}_t} r_{u_{t'}} + \sum_{v \in \mc{I}_t \backslash \mc{S}_t} r_v}{\sum_{u_{t'} \in \mc{S}_t} d_{u_{t'}} + \sum_{v \in \mc{I}_t \backslash \mc{S}_t} d_v } = \frac{\sum_{v \in  \mc{I}_t} r_v}{\sum_{v \in \mc{I}_t}  d_v} = \frac{\|\bm r_{t}\|_1}{\sum_{v \in  \mc{I}_t  } d_v} \label{inequ:50},
\end{equation}
where the last equality is due to the fact that $\mc{I}_t$ indexes all nonzero entries of $\bm r_t$, i.e., $\|\bm r_{t}\|_1 = \sum_{v \in \mc{I}_t } r_t(v)$.
Combine \eqref{inequ:48} and \eqref{inequ:50}, for $t = 1,2, \ldots, T$, we have 
\begin{equation}
\|\bm r_{t + 1}\|_1 < \left(1 - 
 \frac{\alpha \sum_{u \in \mc{S}_{t}} d_u}{\sum_{v \in \mc{I}_{t}} d_v} \right) \|\bm r_{t}\|_1. \label{inequ:34}
\end{equation}
Notice that $\| \bm r_{T+1}\|_1$ is lower bounded by \eqref{inequ:lower-bound-r}. Use \eqref{inequ:34} from $t=1$ to $t=T$, we obtain
\begin{equation}
\eps (1-\alpha) \Tilde{C}_T  \leq \|\bm r_{T+1}\|_1 \leq \prod_{t = 1}^{T} \left(1 - 
 \frac{\alpha \sum_{u \in \mc{S}_{t}} d_u}{\sum_{v \in \mc{I}_t} d_v} \right) \label{inequ:36},
\end{equation}
where $\Tilde{C}_T = \sum_{v\in \mc{I}_{T+1}} \frac{ d_u w_{u v}}{D_u}$. Take the logarithm on both sides of \eqref{inequ:36} and use the fact that $ \log(1-x) \leq -x, \forall x\geq 0$. We reach 
\begin{equation}
\alpha \sum_{t=1}^{T} \frac{\sum_{u \in \mc{S}_{t}} d_u}{\sum_{v \in \mc{I}_t} d_v} \leq \log\left(\frac{1}{\eps(1-\alpha)\Tilde{C}_T}\right). 
\label{inequ:37}
\end{equation}
Let $\eta_t = \tfrac{\sum_{u \in \mc{S}_{t}} d_u}{\sum_{v \in \mc{I}_t} d_v}$ be the \textit{active} ratio at $t$-th epoch.  The average of all $\eta_t$ is then defined as ${\eta}_{1:T} = \tfrac{1}{T}\sum_{t=1}^{T} \eta_t$. Note both $\eta_t$ and ${\eta}_{1:T}$ are in $(0, 1]$. Then by \eqref{inequ:37}, the total number of epochs can be bounded by
\begin{align*}
T \leq \frac{1}{\alpha \cdot {\eta}_{1:T}} \log\left(\frac{1}{\eps(1-\alpha)\Tilde{C}_T}\right).     
\end{align*}
The total operations for processing \textit{active} nodes is $R_T$, which can be represented as
\begin{equation}
R_T = \sum_{t = 1}^{T} \sum_{u \in \mc{S}_t} d_u = \sum_{t = 1}^{T} \vol{\mc S_t} = T \cdot \vol{\mc S_{1:T}} \leq \frac{\vol{\mc S_{1:T}}}{\alpha \cdot {\eta}_{1:T}} \log\left(\frac{1}{\eps(1-\alpha)\Tilde{C}_T}\right)
\label{inequ:upper-bound-2}
\end{equation}
\end{enumerate}
Combine two bounds in \eqref{inequ:upper-bound-1} and \eqref{inequ:upper-bound-2}, we have
\begin{equation}
R_T \leq \min\left\{ \frac{1}{\alpha \eps}, \frac{\vol{\mc S_{1:T}}}{\alpha \cdot {\eta}_{1:T}} \log\left(\frac{1}{\eps(1-\alpha)\Tilde{C}_T}\right) \right\}. \nonumber
\end{equation}
Let $C_{\alpha,T} = 1/ ((1-\alpha)\Tilde{C}_T )$, we finish the proof.
\end{proof}

\begin{remark}
One of the key components in our proof is the local convergence factor $\eta_t$ in \eqref{inequ:34}, which is inspired by a critical observation in Lemma 4.4 of \citet{wu2021unifying}. The authors show that \textsc{FIFOPush} is similar to a variant of Power Iteration when $\eps < \frac{1}{2 m}$ with admitted time complexity $\mc{O}(\frac{m}{\alpha}\log(\frac{1}{\eps m}))$. There is no bound for $\eps > \frac{1}{2 m}$. However, our provided bound works for all $\eps > 0$. We first show that there is a relatively significant amount of residual left in $\bm r_t$, which makes us bound the total epochs $T$ by $\mc{O}(\tfrac{1}{\alpha\cdot {\eta}_{1:T}}\log(\tfrac{1}{\eps(1-\alpha)})$. The other critical component is that we show the number of operations of each epoch mainly depends on $\mc{O}(\vol{\mc S_{t}})$ instead of $\mc{O}(m)$.
\end{remark}

\begin{figure}[!htbp]
\centering
\includegraphics[width=.95\textwidth]{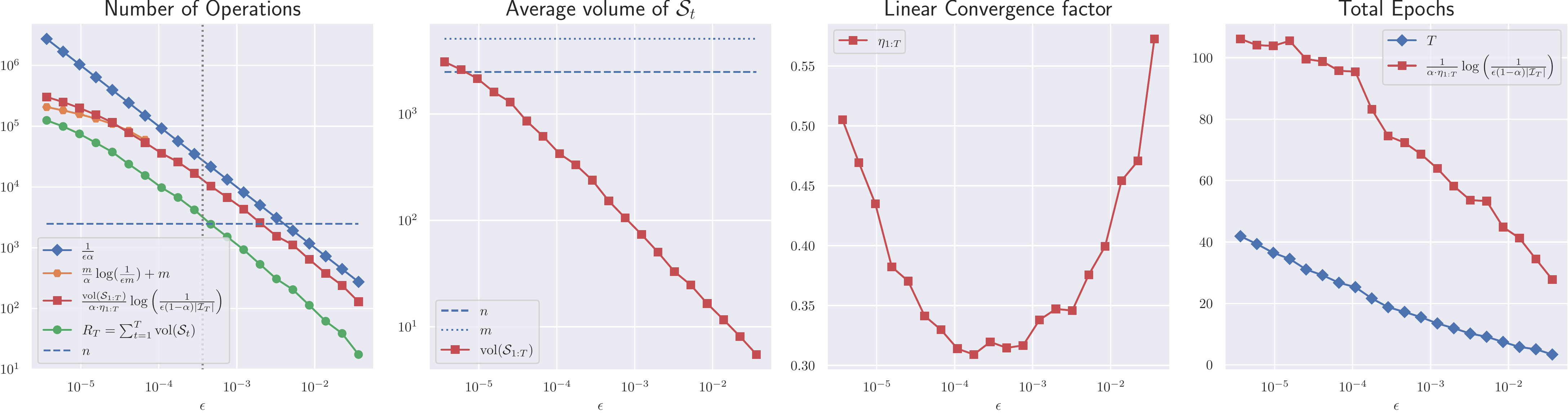}
\caption{The illustration of operations on the \textit{Cora} dataset. We run \textsc{FIFOPush}$(\mc{S},\eps,\alpha, s)$ for $\alpha =0.1$ and $\eps \in \left[10^{-2}, 10^2\right]*\sqrt{\frac{1-\alpha}{1+\alpha}}/n$. Compared with the linear bound $1/\alpha\eps$ in \citet{andersen2006local} and power-iteration bound $\frac{m}{\alpha} \log\left(\frac{1}{\eps m}\right)+m$ provided in \citet{wu2021unifying}, our bound is better and shows the ``locality'' property of \textsc{FIFOPush}. Note that the bound $\frac{m}{\alpha} \log\left(\frac{1}{\eps m}\right)+m$ only works when $\eps < 1/2m$.}
\label{fig:empirical-upper-bound-cora-alpha-0.1}
\end{figure}

\begin{figure}[!htbp]
\centering
\includegraphics[width=.95\textwidth]{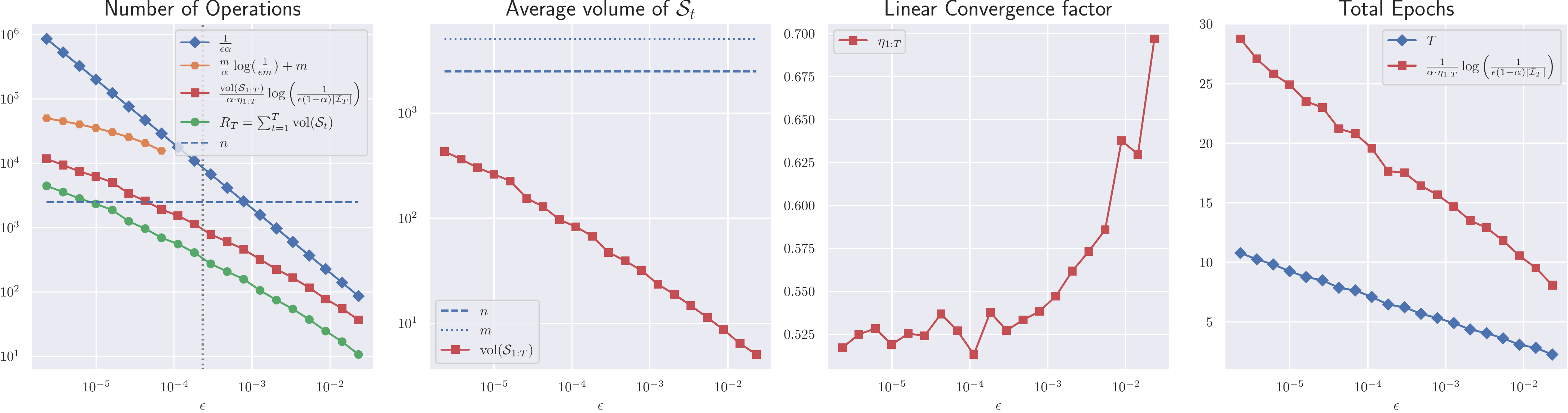}
\caption{The illustration of our bound and parameters on the \textit{Cora} dataset. We run \textsc{FIFOPush}$(\mc{S},\eps,\alpha, s)$ with $\alpha =0.5$.}
\label{fig:empirical-upper-bound-cora-alpha-0.5}
\end{figure}

\begin{figure}[!htbp]
\centering
\includegraphics[width=.95\textwidth]{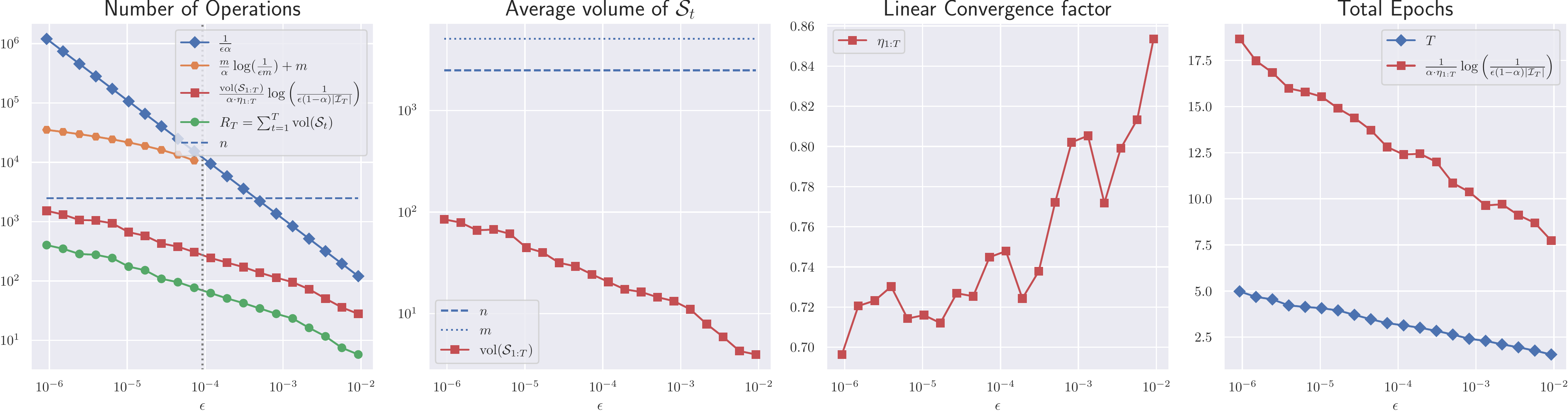}
\caption{The illustration of our bound and parameters on the \textit{Cora} dataset. We run \textsc{FIFOPush}$(\mc{S},\eps,\alpha, s)$ with $\alpha =0.9$.}
\label{fig:empirical-upper-bound-cora-alpha-0.9}
\end{figure}

\begin{figure}[!htbp]
\centering
\includegraphics[width=.95\textwidth]{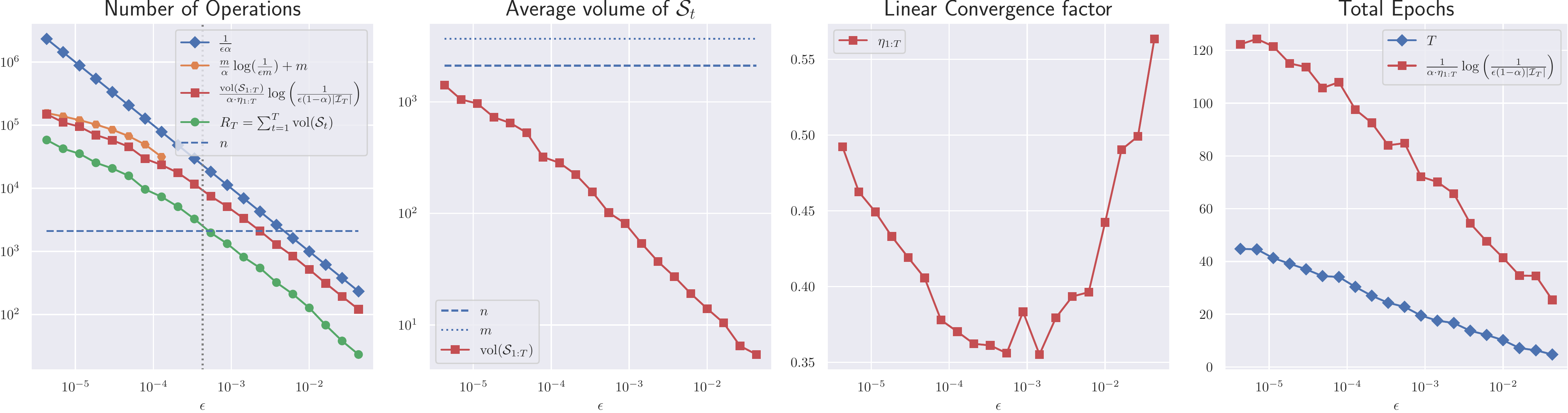}
\caption{The illustration of operations on the \textit{Citeseer} dataset. We run \textsc{FIFOPush}$(\mc{S},\eps,\alpha, s)$ for $\alpha =0.1$ and $\eps \in \left[10^{-2}, 10^2\right]*\sqrt{\frac{1-\alpha}{1+\alpha}}/n$. Compared with the linear bound $1/\alpha\eps$ in \citet{andersen2006local} and power-iteration bound $\frac{m}{\alpha} \log\left(\frac{1}{\eps m}\right)+m$ provided in \citet{wu2021unifying}, our bound is better and shows the ``locality'' property of \textsc{FIFOPush}. Note that the bound $\frac{m}{\alpha} \log\left(\frac{1}{\eps m}\right)+m$ only works when $\eps < 1/2m$.}
\label{fig:empirical-upper-bound-citeseer-alpha-0.1}
\end{figure}

\begin{figure}[!htbp]
\centering
\includegraphics[width=.95\textwidth]{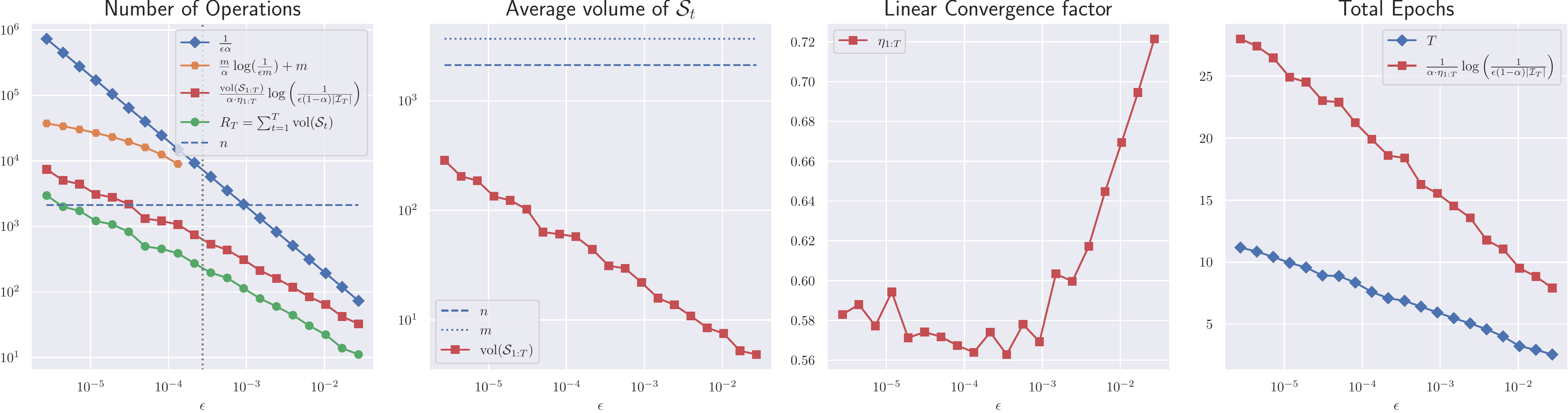}
\caption{The illustration of our bound on the \textit{Citeseer} dataset. We run \textsc{FIFOPush}$(\mc{S},\eps,\alpha, s)$ with $\alpha =0.5$.}
\label{fig:empirical-upper-bound-citeseer-alpha-0.5}
\end{figure}

\begin{figure}[!htbp]
\centering
\includegraphics[width=.95\textwidth]{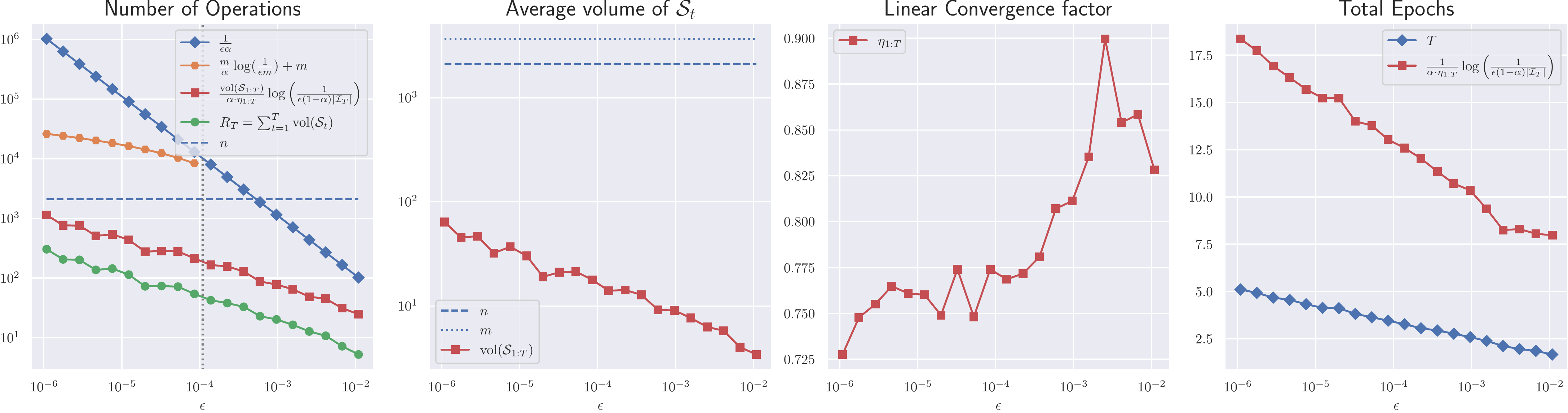}
\caption{The illustration of our bound on the \textit{Citeseer} dataset. We run \textsc{FIFOPush}$(\mc{S},\eps,\alpha, s)$ with $\alpha =0.9$.}
\label{fig:empirical-upper-bound-citeseer-alpha-0.9}
\end{figure}

To see the effectiveness of the local linear convergence bound, we apply \textsc{FIFOPush} with $\alpha = 0.1, 0.5, 0.9$ where the results as illustrated in Fig. \ref{fig:empirical-upper-bound-cora-alpha-0.1}, \ref{fig:empirical-upper-bound-cora-alpha-0.5}, and \ref{fig:empirical-upper-bound-cora-alpha-0.9} of Cora dataset. We also include the results of \textit{Citeseer} dataset as shown in Fig. \ref{fig:empirical-upper-bound-citeseer-alpha-0.1}, \ref{fig:empirical-upper-bound-citeseer-alpha-0.5}, and \ref{fig:empirical-upper-bound-citeseer-alpha-0.9}. Our bound is much better, especially when $\alpha$ is large. We find similar patterns on other graph datasets. 

\subsection{Local linear convergence of \textsc{FIFOPush} for $(\alpha \bm I + \bm D - \bm W)^{-1}$: The proof of Thm. \ref{thm:local-convergence-2}}

Given any $\alpha > 0, \epsilon > 0$, Algo. \ref{algo:dummy-fifo-push-2} is to approximate $(\alpha \bm I + \bm D - \bm W)^{-1}$

\begin{algorithm}[H]
\caption{$\textsc{FIFOPush}(\mc{G},\eps,\alpha, s)$ with a dummy node $\red{\ddag}$}
\begin{algorithmic}[1]
\STATE \textbf{Initialize}: $\bm r_t = \tfrac{\bm e_s}{\alpha}, \ \ \bm x_t = \bm 0, \ \ t= 1, \ \ t^\prime = 1$
\STATE $\mc{Q}=[s, \red{\ddag}]$ \quad\quad\quad\quad\quad\quad\quad\ \ // At the initial stage, $\mc{Q}$ contains $s$ and a dummy node $\red{\ddag}$.
\WHILE{$\mc{Q}\text{.size}()\ne 1$}
\STATE $u_{t'} = \mc{Q}\text{.pop()}$
\IF{$u_{t'} == \red{\ddag}$}
\STATE $t = t + 1$ \quad\quad\quad\quad\quad\quad // Nodes in $\mc{U}_t$ has been processed. Go to the next epoch.
\STATE $\mc{Q}\text{.push}(u_{t'})$
\STATE \textbf{continue}
\ENDIF
\IF{$r_{u_{t'}} < \eps \cdot d_{u_{t'}}$}
\STATE $t^\prime = t^\prime + 1$ \quad\quad\quad\quad\quad\ \ // $u_{t'}$ is an ``\textit{inactive}'' node
\STATE \textbf{continue}
\ENDIF
\STATE $x_{u_{t'}} = x_{u_{t'}} + \frac{\alpha r_{u_{t'}}}{\alpha + D_{u_{t'}}}$ 
  \quad\quad\quad // $u_{t'}$ is an ``\textit{active}'' node
\FOR{$v \in \nei(u_{t'})$}
\STATE $r_{v} = r_{v} + \frac{r_{u_{t'}}w_{v u_{t'}}}{\alpha + D_{u_{t'}}}$
\IF{$v \notin \mc{Q}$}
\STATE $\mc{Q}\text{.push}(v)$
\ENDIF
\ENDFOR
\STATE $r_{u_{t'}} = 0$ 
\STATE $t' = t' + 1$
\ENDWHILE
\STATE \textbf{Return} $(\bm x_t,\bm r_t)$
\end{algorithmic}
\label{algo:dummy-fifo-push-2}
\end{algorithm}

\begin{reptheorem}{thm:local-convergence-2}[Local convergence of \textsc{FIFOPush} for $\bm X_{\mc L}$] Let $\bm x_s = \bm X_{\mc L} \bm e_s$ and run Aglo. \ref{algo:fifo-push} for $\bm X_{\mc L}$. For $s, i \in \mc{V}$, we have $ \bm x_s = \bm x_{s,\eps} + \alpha \bm X_{\mc L} \bm r_{s,\eps}, \text{ with } r_{s,\eps}(i) \leq [0,\eps d_i), \forall i\in \mathcal{V}$. The main operations of $ \textsc{FIFOPush}$ for $\bm X_{\mc L}$ is bounded as
\begin{equation}
R_T \leq \frac{\vol{\mc S_{1:T}}(\alpha + D_{\max})}{\alpha \cdot \eta_{1:T}} \log\left( \frac{C_{\alpha,T}}{\eps} \right), 
\end{equation}
where $\vol{ \mc S_{1:T}}$ and $\eta_{1:T}$ are the same as in Thm. \ref{thm:local-convergence}, 
$\eta_t \triangleq \tfrac{\sum_{u \in \mc{S}_t} d_u/(\alpha + D_u)}{\sum_{v \in \mc{I}_{t}} d_v/(\alpha + D_v)}$, 
$C_{\alpha,T} = 1 / \sum_{v\in \mc{I}_{T}} \frac{d_u w_{u v}}{\alpha + D_u}$, and $D_{\max} = \max_{v\in \supp{\bm x_{s,\eps}}} D_v$. 
\label{thm:local-convergence-2-repeat}
\end{reptheorem}

\begin{proof}
The key of Alg. \ref{algo:dummy-fifo-push-2} is to maintain $\bm x_t$ and $\bm r_t$ so that $1/\alpha$ magnitudes will move from $\bm r$ to $\bm x$. For each active node $u_t$, $\bm x_t$ updates to $\tilde{{\bm x}}_{t+1}$ and $\bm r_t$ updates to $\tilde{{\bm r}}_{t+1}$ as the following
\begin{align*}
\Tilde{\bm x}_{t+1} &= \bm x_t + \frac{\alpha r_u}{\alpha + D_u} \bm e_u \\
\Tilde{\bm r}_{t+1} &= \bm r_t + \frac{r_u}{\alpha + D_u} \bm W \bm e_u  - r_u \bm e_u = \bm r_t  - \left( \bm I + \frac{\bm D - \bm W}{\alpha} \right) \frac{\alpha r_u}{\alpha + D_u} \bm e_u \\
\frac{\alpha r_u}{\alpha + D_u} \bm e_u &= \left( \bm I + \frac{\bm D - \bm W}{\alpha} \right)^{-1}\left( \bm r_t  - \Tilde{\bm r}_{t+1} \right)
\end{align*}
Bring back $t'$ for $u$. After all active nodes in $t$-th epoch have been updated, we have
\begin{align*}
{\bm x}_{t+1} &= \bm x_t + \sum_{u_{t'} \in \mathcal{S}_t} \frac{\alpha r_{u_{t'}}}{\alpha + D_{u_{t'}}} \bm e_{u_{t'}} \\
&= \bm x_t + \left( \bm I + \frac{\bm D - \bm W}{\alpha} \right)^{-1} \left( \bm r_t  - {\bm r}_{t+1} \right) = \bm x_1 + \left( \bm I + \frac{\bm D - \bm W}{\alpha} \right)^{-1} \sum_{i=1}^t\left( \bm r_i  - {\bm r}_{i+1} \right)\\
&= \left( \bm I + \frac{\bm D - \bm W}{\alpha} \right)^{-1}\left( \frac{\bm e_s}{\alpha} - {\bm r}_{t+1} \right),
\end{align*}
where $\bm r_t = \tilde{\bm r}_{t} \rightarrow \tilde{\bm r}_{t'+1} \rightarrow \tilde{\bm r}_{t'+2} \rightarrow \cdots \tilde{\bm r}_{t'+|\mc{S}_t|} = {\bm r}_{t+1}$.
Denote $\bm x_{t+1}$ as $\bm x_{s,\eps}$ and $\bm r_{t+1}$ as $\bm r_{s,\eps}$, then we have 
\begin{equation}
\bm x_s = \bm x_{s,\eps} + \alpha \bm X \bm r_{s,\eps}.
\end{equation}

After the $T$-th epoch finished, $\mc{I}_{T+1} = \supp{\bm r_{s,\eps}}$.  For each $v \in \mc{I}_{T+1}$, note that there exists at least one of its neighbor $u \in \nei{(v)}$ such that $r_{u} \geq \eps \cdot d_{u}$ had happened in a previous active iteration. \begin{equation}
\forall v \in \mc{I}_{T+1},\quad \frac{r_u w_{u v}}{\alpha + D_u}  \geq \frac{\eps d_u w_{u v}}{\alpha + D_u}  \triangleq \tilde{r}_v, \nonumber
\end{equation}
where $\tilde{r}_v$ is the residual pushed into $v$ but never popped out; hence $\sum_{v\in \mc{I}_{T+1}} \tilde{r}_v$ is an estimate of $\|\bm r_{T+1}\|_1$ from bottom. That is
\begin{equation}
\sum_{v\in \mc{I}_{T+1}} \tilde{r}_v = \sum_{v\in \mc{I}_{T+1}} \frac{\eps d_u w_{u v}}{\alpha + D_u}  \leq \|\bm r_{T+1}\|_1. \label{inequ:lower-bound-r-2}    
\end{equation}

The operation bounds are similar to that of \ref{thm:local-convergence}. Note that for each epoch, the updates of $\bm r$ satisfies
\begin{equation}
\|\bm r_t\|_1 - \|\bm r_{t+1}\|_1 \geq \sum_{u_{t'} \in \mc{S}_t}\frac{\alpha r_{u_{t'}} }{ \alpha + D_{u_{t'}} }. \label{equ:54}
\end{equation}
By the condition, we have $\forall u_{t'} \in \mc{S}_{t}, r_{u_{t'}} \geq \eps \cdot d_{u_{t'}}, \forall v \in \mc{I}_t \backslash \mc{S}_t, 0 < r_t(v) < \eps \cdot d_v$. Summation above inequalities over all active $u_{t'}$ and inactive $v$, multiply $\tfrac{\alpha}{\alpha + D_{u}}$ on both sides, we have
\begin{equation}
\frac{\sum_{u_{t'} \in \mc{S}_{t}} \frac{\alpha r_{u_{t'}}}{\alpha + D_{u_{t'}}} }{\sum_{u_{t'} \in \mc{S}_{t}} \frac{ \alpha d_{u_{t'}} }{ \alpha + D_{u_{t'}} } } \geq \eps > \frac{\sum_{v \in \mc{I}_t \backslash \mc{S}_t} \frac{ \alpha r_v}{\alpha + D_v} }{\sum_{v \in \mc{I}_t \backslash \mc{S}_t} \frac{\alpha d_v}{\alpha + D_v} }, \nonumber
\end{equation}
which indicates
\begin{equation}
\frac{\sum_{u_{t'} \in \mc{S}_{t}} \frac{\alpha r_{u_{t'}}}{\alpha + D_{u_{t'}}} }{\sum_{u_{t'} \in \mc{S}_{t}} \frac{ \alpha d_{u_{t'}} }{ \alpha + D_{u_{t'}} } } \geq \frac{\sum_{u_{t'} \in \mc{S}_{t}} \frac{\alpha r_{u_{t'}}}{\alpha + D_{u_{t'}}} + \sum_{v \in \mc{I}_t \backslash \mc{S}_t} \frac{ \alpha r_v}{\alpha + D_v} }{\sum_{u_{t'} \in \mc{S}_{t}} \frac{ \alpha d_{u_{t'}} }{ \alpha + D_{u_{t'}} } + \sum_{v \in \mc{I}_t \backslash \mc{S}_t} \frac{\alpha d_v}{\alpha + D_v} } \geq \frac{\frac{\alpha}{\alpha + D_{\max}}\| \bm r_t\|_1}{\sum_{v \in \mc{I}_t }\frac{\alpha d_v}{\alpha + D_v} } \label{equ:55}
\end{equation}
where the last equality is due to the fact that $\mc{I}_t$ indexes all nonzero entries of $\bm r_t$, i.e., $\|\bm r_{t}\|_1 = \sum_{v \in \mc{I}_t } r_t(v)$.
Combine \eqref{equ:54} and \eqref{equ:55}, for $t = 1,2, \ldots, T$, we have 
\begin{equation}
\|\bm r_{t + 1}\|_1 < \left(1 - 
 \frac{\alpha}{\alpha + D_{\max}} \frac{ \sum_{u \in \mc{S}_{t}} d_u /(\alpha + D_{u})}{ \sum_{v \in \mc{I}_{t}} d_v/(\alpha + D_{v})} \right) \|\bm r_{t}\|_1. \nonumber
\end{equation}
From $t=1$ to $t=T$, we obtain
\begin{equation}
\sum_{v\in \mc{I}_{T+1}} \frac{ \eps d_u w_{u v}}{\alpha + D_u} \leq \|\bm r_{T+1}\|_1 \leq \prod_{t = 1}^{T} \left(1 - 
 \frac{\alpha}{\alpha + D_{\max}} \frac{ \sum_{u \in \mc{S}_{t}} d_u /(\alpha + D_{u})}{ \sum_{v \in \mc{I}_{t}} d_v/(\alpha + D_{v})} \right) \nonumber.
\end{equation}
Denote $C_T = 1 / \sum_{v\in \mc{I}_{T+1}} \frac{d_u w_{u v}}{\alpha + D_u}$. Take the logarithm on both sides of the above and we reach 
\begin{equation}
 \frac{\alpha}{\alpha + D_{\max}} \sum_{t=1}^T \frac{ \sum_{u \in \mc{S}_{t}} d_u /(\alpha + D_{u})}{ \sum_{v \in \mc{I}_{t}} d_v/(\alpha + D_{v})} \leq \log\left(\frac{C_T}{\eps}\right). 
\label{inequ:58}
\end{equation}
Then by \eqref{inequ:58}, the total number of epochs can be bounded by \(T \leq \tfrac{\alpha + D_{\max}}{\alpha \cdot {\eta}_{1:T}} \log\left(\tfrac{C_T}{\eps}\right)\). The total operations for processing \textit{active} nodes is \(R_T = \sum_{t = 1}^{T} \vol{\mc{S}_t}\), which is bounded as
\begin{equation}
R_T \leq \frac{(\alpha + D_{\max})\vol{\mc S_{1:T}}}{\alpha \cdot {\eta}_{1:T}} \log\left(\frac{C_T}{\eps}\right). \nonumber
\end{equation}
\end{proof}

\begin{figure}
\centering     
\subfigure[Labeled Karate]{\label{fig:type-i-a}\includegraphics[width=50mm]{figs/fig-karate-graph}}\quad\quad
\subfigure[$\bm x_{22,\eps}$ on $\mathcal{G}$]{\label{fig:type-i-b}\includegraphics[width=50mm]{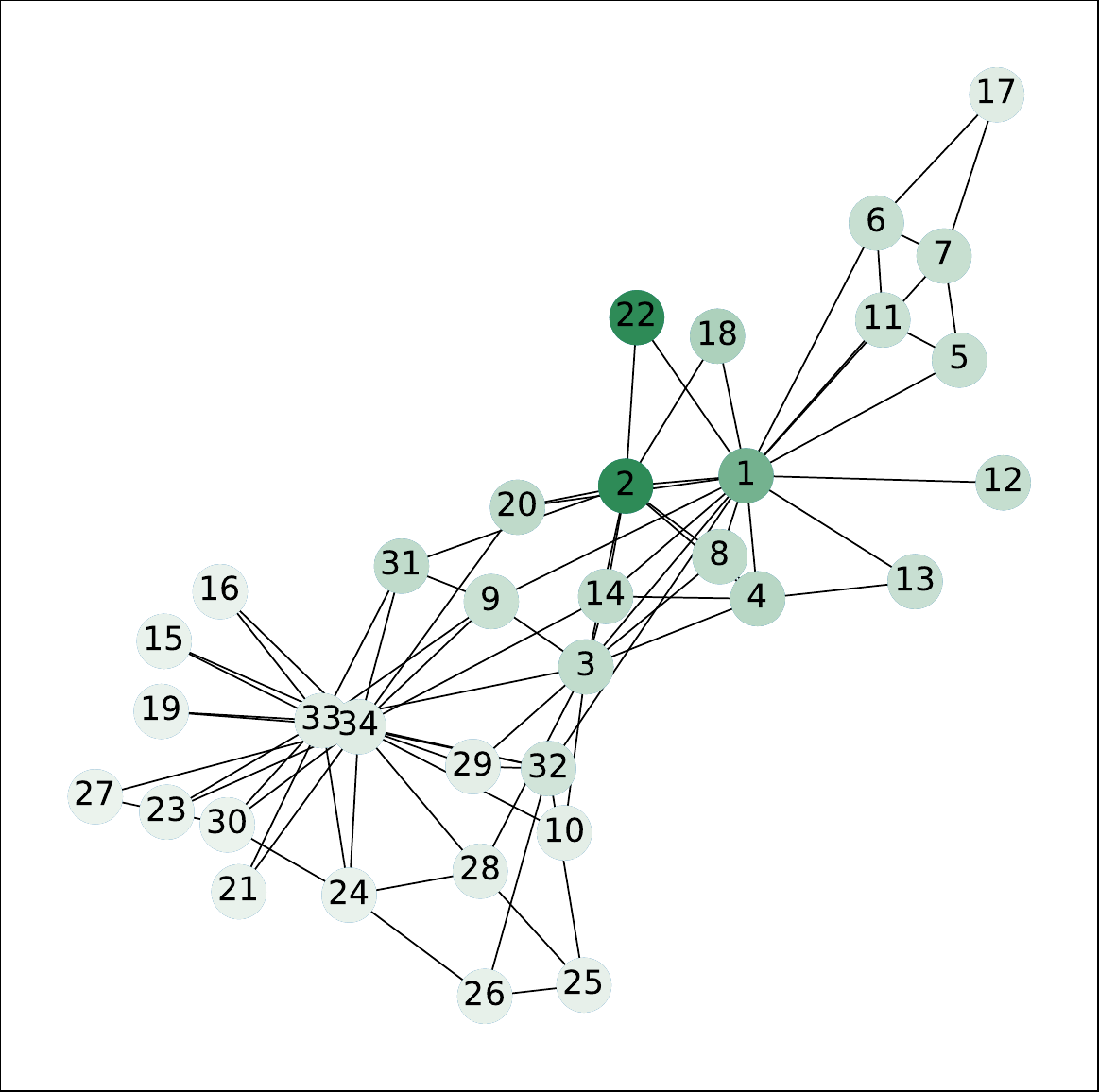}}\quad\quad
\subfigure[Power law of $\bm X_{\eps}$]{\label{fig:type-i-c}\includegraphics[width=55mm]{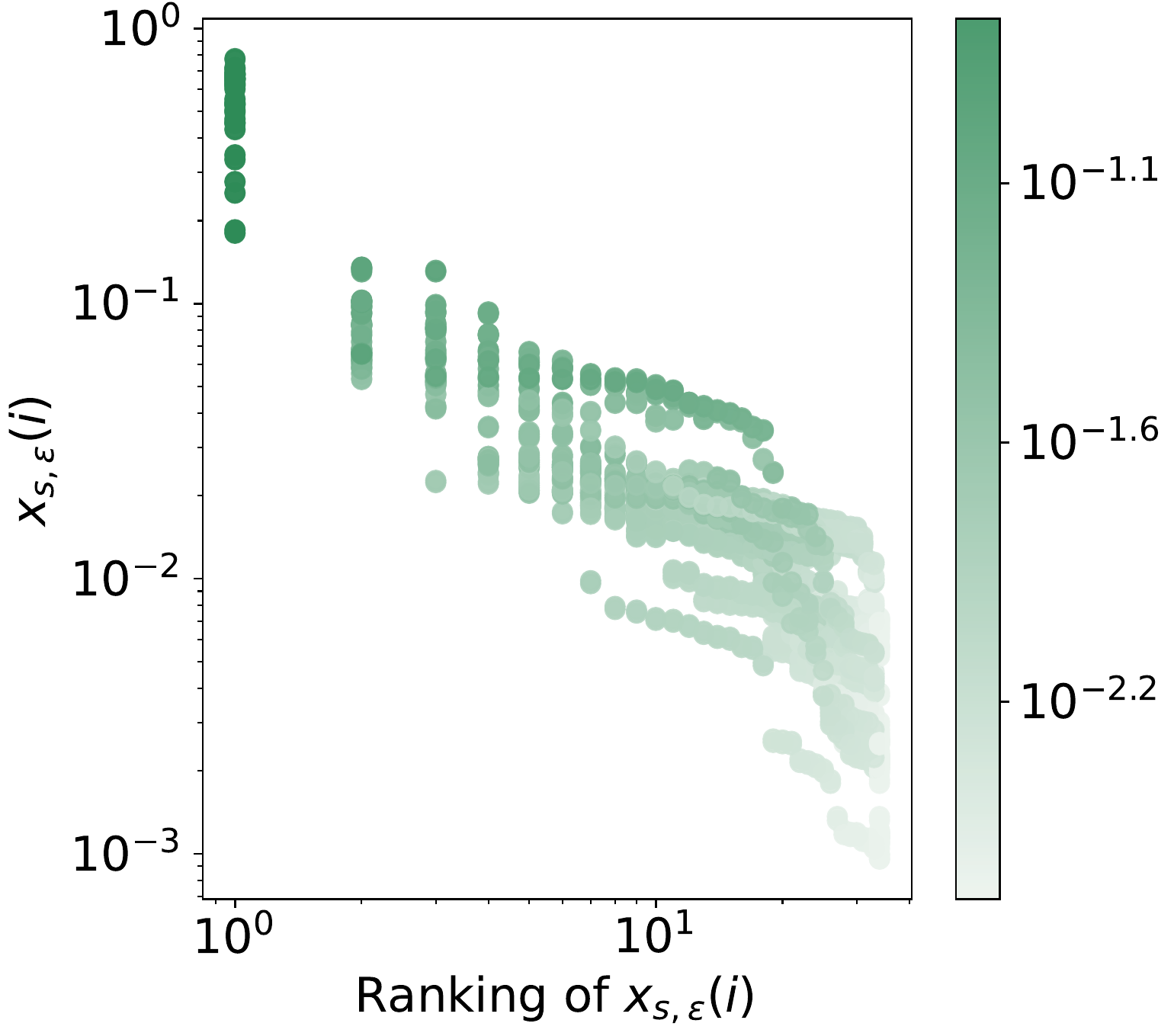}}
\caption{Power law distribution of $\bm X_{\mc L}\bm e_s$ on Karate graph \cite{girvan2002community}. (a) The Karate contains 34 nodes and 78 edges forming four communities ($w_{u v}$ is the 2-dimensional Euclidean distance.); (b) We run \textsc{FIFOPush}$(\mc G, \eps,\alpha,s)$ on $s=22$ with $\alpha = 0.85$ and $\eps = 10^{-12}$. Neighbors of $s=22$ have large magnitudes of $\bm x_{s}$; (We changed the magnitude of $\bm x_{s}(22)$ to the largest magnitude of rest for better visibility.) (c) The power law distribution of entries of $\bm x_s$ for all 34 nodes.}
\label{fig:power-law-of-x-s-epsilon-type-i}
\end{figure}

\subsection{Approximation kernels and their residuals}

\begin{table}[h!]
\caption{The parameterized graph kernel matrices with their kernel approximation}
\centering
\small
\begin{tabular}{p{.2cm}| p{1cm}|p{5.3cm}|p{3.9cm}|p{4.3cm}}
 \toprule
 ID & $\alpha$ & Basic Kernel Presentation & Approximation $\bm M_\eps$ & Residual Matrix $\bm E_\eps$ \\ [0.5ex] 
 \midrule
1 & $\frac{\lambda}{n}$ & $\bm M_{\lambda,\beta} = 2\lambda \bm X_{\mc L}$ & $2\lambda \bm X_\eps$ & $2\lambda \alpha \bm X_{\mc{\bm L}} \bm R_\eps$ \\  [1.5ex] 
2 & $\frac{\lambda}{n+\lambda}$ & $\bm M_{\lambda,\beta} = 
 2 n {\bm D}^{-1/2}\bm X_{\bm L} {\bm D}^{1/2}$ &  $2n \bm D^{-1/2} {\bm X}_\eps \bm D^{1/2}$ & $2n \bm D^{-1/2} \bm X_{\bm L} \bm R_\eps \bm D^{1/2}$ \\ [1.5ex] 
3 & $\frac{n+\lambda -\beta n}{n+\lambda}$ & $\bm M_{\lambda,\beta} =  \frac{2\lambda n}{n+\lambda - \beta n} {\bm D}^{-1/2}\bm X_{\bm L} {\bm D}^{1/2}$ & $\frac{2 \lambda n}{n+\lambda -\beta n} \bm D^{-1/2} {\bm X}_\eps \bm D^{1/2}$ & $\frac{2 \lambda n}{n+\lambda -\beta n} \bm D^{-1/2} \bm X_{\bm L} \bm R_\eps\bm D^{1/2}$ \\ [1.5ex]
4 & $\frac{n \beta +\lambda}{n}$ & $ \bm M_{\lambda,\beta} = 2\lambda {\bm S}^{-1/2}\bm X_{\mc L} {\bm S}^{1/2}$ & $2\lambda\bm S^{-1/2}\bm X_\eps\bm S^{1/2}$ & $2\lambda\bm S^{-1/2} \alpha \bm X_{\mc{\bm L}} \bm R_\eps\bm S^{1/2}$ \\ [1.5ex] 
5 & $ 2 \lambda $ & $ \bm M_{\lambda,\beta} = \left( \frac{\bm S^{1/2}}{4 n\lambda} + \frac{\beta\bm S^{-1/2}}{4\lambda^2} \right)^{-1} \bm X_{\mc L} {\bm S}^{1/2}$ & $\left( \frac{\bm S^{1/2}}{4 n\lambda} + \frac{\beta\bm S^{-1/2}}{4\lambda^2} \right)^{-1} \bm X_\eps {\bm S}^{1/2}$ & $\alpha \left( \frac{\bm S^{1/2}}{4 n\lambda} + \frac{\beta\bm S^{-1/2}}{4\lambda^2} \right)^{-1} \bm X_{\mc L}\bm R_\eps {\bm S}^{1/2}$ \\ [1.5ex] 
6 & $\beta + \frac{\lambda}{n} $ & $\bm M_{\lambda,\beta} = 2\lambda \bm X_{\mc L} \left( \bm I - \frac{b \bm 1\bm 1^\top}{\alpha + n b} \right) $ & $\bm M_{\lambda,\beta} = 2\lambda \bm X_\eps \left( \bm I - \frac{b \bm 1\bm 1^\top}{\alpha + n b} \right)$ & $ 2\lambda \alpha \bm X_{\mc L} \bm R_\eps \left( \bm I - \frac{b \bm 1\bm 1^\top}{\alpha + n b} \right)$ \\ [1.5ex] 
 \bottomrule
 \end{tabular}
 \label{tab:graph-kernel-approximation}
 \end{table}

Recall two types of matrix presentation and their approximations 
\begin{align*}
\bm X_{\bm L} &= \alpha (\bm I - (1-\alpha) \bm W \bm D^{-1})^{-1} = \bm X_\eps + \bm X_{\bm L} \bm R_\eps, \\
\bm X_{\mc{\bm L}} & = (\alpha \bm I + \bm D - \bm W)^{-1} = \bm X_\eps + \alpha \bm X_{\mc{\bm L}} \bm R_\eps.
\end{align*}

Based on the above lemmas, we list the approximation kernels in Tab. \ref{tab:graph-kernel-approximation}. $\bm X_\eps =[\bm x_{1,\eps}, \bm x_{2,\eps},\ldots,\bm x_{n,\eps}]$ be the matrix obtained by applying \textsc{FIFOPush} as described in Algorithm \ref{algo:fifo-push}. We only show the first two cases to see how to represent these matrices in terms of $\bm X_\eps$ and $\bm R_\eps$. For example, in the first case, we have
\begin{align*}
\bm M_{\lambda,\beta} = \left(\frac{\bm D - \bm W}{2\lambda} + \frac{\bm I_n}{2 n} \right)^{-1} =   2\lambda \left( \alpha \bm I_n + {\mc{\bm L}}  \right)^{-1} = 2\lambda \bm X_{\mc{\bm L}} = 2 \lambda (\bm X_\eps + \alpha \bm X_{\mc{\bm L}} \bm R_\eps).
\end{align*}
For the second case, the kernel matrix can be rewritten as $\bm M = 2 n {\bm D}^{-1/2} \bm X_{\bm L} {\bm D}^{1/2}$. By Lemma \ref{lemma:A-1}, we have
\begin{align*}
\bm X_{\bm L} &= {\bm X}_\eps + \bm X_{\bm L} \bm R_\eps  \\
2n \bm D^{-1/2} \bm X_{\bm L} \bm D^{1/2} &= 2n \bm D^{-1/2} ({\bm X}_\eps + \bm X_{\bm L} \bm R_\eps) \bm D^{1/2} \\
\bm M &= 2n \bm D^{-1/2} {\bm X}_\eps \bm D^{1/2} + 2n \bm D^{-1/2} \bm X_{\bm L} \bm R_\eps \bm D^{1/2} \\
\bm M &= 2n \bm D^{-1/2} {\bm X}_\eps \bm D^{1/2} + \bm M \bm D^{-1/2} \bm R_\eps \bm D^{1/2}.
\end{align*}
For all these cases, we have the relationship that $\bm M = \bm M_\eps  + \bm E_\eps$.

\section{Eigenvalues of $\bm M_{\lambda,\beta}$ and quadratic approximation guarantees of $\bm M_\eps$}
\label{sect:appendix:eig-evaluation}

$\bm M_{\lambda,\beta}$ is the matrix we want to approximate. Recall $\bm X_\eps$ be the approximated matrix by \textsc{FIFOPush}. $\bm M_\eps$ be the matrix built upon $\bm X_\eps$ (See 3rd column of Table \ref{tab:graph-kernel-approximation}). In the following, we will construct the relaxation function based on $\bm M_\eps$. The following lemma shows that if $\bm M_\eps$ is a good approximation of $\bm M$ then we can define a relaxation function based $\bm M_\eps$. Before we present the theorem, let us characterize the eigenvalues of $\bm X_{\bm L}$ and $\bm X_{\mc{L}}$.

\begin{lemma}
The eigenvalue functions of these two basic kernel matrices $\bm X_{\bm L}$ and $\bm X_\mc{\bm L}$ satisfy 
\begin{align*}
\lambda(\bm X_{\bm L}) = \lambda\left( \alpha \left( \bm I - (1-\alpha)\bm W \bm D^{-1}\right)^{-1} \right) &\in \left[\frac{\alpha}{2-\alpha}, 1\right], \\
\lambda(\bm X_{\mc{\bm L}}) = \lambda\left( \left( \alpha \bm I + \bm D - \bm W\right)^{-1} \right) &\in \left[ \frac{1}{\alpha + 2 D_{\max}}, \frac{1}{\alpha}\right],
\end{align*}
where $D_{\max}$ is the maximum weighted degree among $\mc V$. For $\bm X_{\bm L}$, we assume $\alpha \in (0,1)$, and for $\bm X_{\mc{\bm L}}$,  $\alpha > 0$.
\label{lemma:eigenvalue-bound}
\end{lemma}
\begin{proof}
Notice that since the magnitude of the eigenvalues of the column stochastic matrix $\bm W \bm D^{-1}$ is bounded by $1$, i.e., $|\lambda(\bm W \bm D^{-1})|\leq 1$, then we have
\begin{align*}
\lambda(\bm I - (1-\alpha)\bm W \bm D^{-1}) &\in \left[\alpha, 2-\alpha\right] \\
\lambda\left( \left( \bm I - (1-\alpha)\bm W \bm D^{-1}\right)^{-1} \right) &\in \left[\frac{1}{2-\alpha},\frac{1}{\alpha}\right] \\
\lambda\left( \alpha \left( \bm I - (1-\alpha)\bm W \bm D^{-1}\right)^{-1} \right) &\in \left[\frac{\alpha}{2-\alpha}, 1\right].
\end{align*}
Hence, we have the first bounding inequality. To show the second inequality, notice that if $\bm x$ is an eigenvalue of $\bm D-\bm W$, then
\begin{align*}
(\bm D - \bm W ) \bm x &= \lambda \bm x \\
\bm x ^\top (\bm D - \bm W ) \bm x &= \lambda \bm x^\top \bm x\\
\sum_{ (u,v)\in \mathcal{E}} w_{u v} (x_u - x_v)^2 &= \lambda \bm x^\top \bm x.
\end{align*}
Hence $\lambda \geq 0$, and the lower bound is achieved when $\bm x = \bm 1$.  On the other hand, a well-known result \cite{anderson1985eigenvalues} of the upper bound is  
\[
\lambda(\bm D - \bm W)\leq \max_{(u,v)\in \mathcal{E}}  D_u + D_v \leq 2 D_{\max}.
\] 
where $D_{\max}$ is the maximum weighted degree among all nodes. It follows that
\begin{align*}
\lambda(\bm D - \bm W) &\in \left[0, 2 D_{\max}\right] \\
\lambda(\alpha \bm I + \bm D - \bm W) &\in \left[\alpha, \alpha + 2 D_{\max}\right] \\
\lambda\left( \left( \alpha \bm I + \bm D - \bm W\right)^{-1} \right) &\in \left[ \frac{1}{\alpha + 2 D_{\max}}, \frac{1}{\alpha}\right].
\end{align*}
\end{proof}

\begin{lemma}
Let $\bm M$ be a symmetric positive definite matrix and $\bm D$ be a positive diagonal matrix. Then, for any nonnegative real matrix $\bm R \in {\mathbb R}_+^{n\times n}$ and $\bm x \ne \bm 0$, we have 
\begin{equation}
\frac{1}{2}\frac{\bm x^\top (\bm M \bm D^{-1/2} \bm R \bm D^{1/2} + \bm D^{1/2} \bm R^\top \bm D^{-1/2} \bm M ) \bm x }{\bm x^\top \bm M \bm x} \leq  \|\bm D^{-1/2} \bm R\bm D^{1/2}\|_2. \label{inequ:r-mat-error-bound}
\end{equation}
\label{lemma:bound-residual-matrix}
\end{lemma}
\begin{proof}
Since $\bm M$ is a symmetric positive definite matrix, one can write $\bm M = \bm B\bm B^\top$ and $\bm B$ are invertible. The decomposition of $\bm M$ is $\bm M = \bm Q \bm \Lambda \bm Q^\top = \bm Q \bm \Lambda^{1/2} \bm \Lambda^{1/2} \bm Q^\top$ where we can let $\bm B = \bm Q\bm \Lambda^{1/2}$ and $\bm Q$ is an orthonormal matrix.  Let $\bm y = \bm B^\top \bm x$, then $\bm x^\top \bm M \bm x = \bm y^\top \bm y$ and $\bm x = (\bm B^\top)^{-1} \bm y$. Let $\bm Z = \bm D^{-1/2} \bm R \bm D^{1/2}$. We have
\begin{align*}
(*) &= \frac{1}{2} \frac{\bm x^\top \left(\bm M \bm D^{-1/2} \bm R \bm D^{1/2} +  \bm D^{1/2} \bm R^\top \bm D^{-1/2} \bm M \right) \bm x}{\bm x^\top \bm M \bm x} \\
&= \frac{1}{2}\frac{\bm x^\top \left(\bm M \bm Z +  \bm Z^\top \bm M \right) \bm x}{\bm y^\top \bm y} \\
&= \frac{1}{2}\frac{\bm y^\top \bm B^{-1} \left(\bm B \bm B^\top \bm Z +  \bm Z^\top \bm B \bm B^\top \right) (\bm B^\top)^{-1}\bm y}{\bm y^\top \bm y} \\
&= \frac{1}{2}\frac{\bm y^\top \left(\bm B^\top \bm Z (\bm B^\top) ^{-1} +  \bm B^{-1} \bm Z^\top \bm B \right)\bm y}{\bm y^\top \bm y} \\
&= \frac{1}{2}\frac{\bm y^\top \left(\bm \Lambda^{1/2} \bm Q^\top \bm Z (\bm Q^\top)^{-1} \bm \Lambda^{-1/2} + {\bm \Lambda}^{-1/2}\bm Q^{-1} \bm Z^\top \bm Q {\bm \Lambda}^{1/2} \right)\bm y}{\bm y^\top \bm y} \\
&\leq \frac{1}{2}\left\| \bm \Lambda^{1/2} \bm Q^\top \bm Z (\bm Q^\top)^{-1} \bm \Lambda^{-1/2} + {\bm \Lambda}^{-1/2}\bm Q^{-1} \bm Z^\top \bm Q {\bm \Lambda}^{1/2}\right\|_2
\end{align*}
where the inequality follows Rayleigh's quotient property, that is, any matrix norm bounds the maximal absolute eigenvalue, and the spectral radius is less than any matrix norm.  Denoting $\mb A = \mb Q^\top \mb Z (\mb Q^\top)^{-1}$, we may write
\begin{eqnarray*}
2(*) &=& \left\| \bm \Lambda^{1/2} \bm Q^\top \bm Z (\bm Q^\top)^{-1} \bm \Lambda^{-1/2} + {\bm \Lambda}^{-1/2}\bm Q^{-1} \bm Z^\top \bm Q {\bm \Lambda}^{1/2}\right\|_2 \\
&=& \max_{u|\|u\|_2=1} |u^T \bm \Lambda^{1/2}\bm A \bm \Lambda^{-1/2}u + u^T{\bm \Lambda}^{-1/2}\bm A {\bm \Lambda}^{1/2} u|_2
\end{eqnarray*}
and denoting $\bm w = \bm \Lambda^{1/2} \bm u, \bm v = \bm \Lambda^{-1/2} \bm u$
\begin{eqnarray*}
|\bm u^\top \bm \Lambda^{1/2}\bm A \bm \Lambda^{-1/2}\bm u + \bm u^\top{\bm \Lambda}^{-1/2}\bm A {\bm \Lambda}^{1/2} \bm u|
&= 2|\bm w^\top\bm A \bm v|  = 2|\tr{\bm v\bm w^\top \bm A}|
&\overset{(\circ)}{\leq} 2|\bm w^\top \bm v|\|\bm A\|_2,
\end{eqnarray*}
where the inequality $(\circ)$ is due to Holder's $p=1$ inequality applied to the matrix singular values. Since $\bm w^\top \bm v= \bm u^\top \bm u=1$, 
$(*)\leq \|\bm Q^\top \bm Z (\bm Q^T)^{-1}\|_2 = \|\bm Z\|_2$.
\end{proof}

The next lemma shows \textsc{FIFOPush} provides good approximations. Here we only show \textbf{Instance 3} and \textbf{Instance 4}.
\begin{lemma}
Let ${\bm X}_\eps$ and ${\bm R}_\eps$ be the approximate and the residual matrix obtained by applying $\textsc{FIFOPush}(\mathcal{G},\eps,\alpha,s)$ $\forall s \in \mathcal{V}$.  Then we have the following inequalities
\begin{align*}
\bm x^\top \left( \frac{\bm M_\eps + \bm M_\eps^\top}{2} \right) \bm x &\geq \bm x^\top \bm M \bm x \left( 1 - \|\bm D^{-1/2} \bm R_\eps \bm D^{1/2}\|_2 \right) \text{ for } \bm M = \frac{2\lambda n}{n+\lambda - \beta n} {\bm D}^{-1/2}\bm X_{\bm L} {\bm D}^{1/2}, \\
\bm x^\top \left( \frac{\bm M_\eps + \bm M_\eps^\top}{2} \right) \bm x &\geq \bm x^\top \bm M \bm x \left( 1 - \|\bm S^{-1/2} \bm R_\eps \bm S^{1/2}\|_2 \right) \text{ for } \bm M = 2\lambda \bm S^{-1/2}\bm X_{\mc L} {\bm D}^{1/2}.
\end{align*}
\label{thm:bounding-error}
\end{lemma}
\begin{proof}
The inequality is trivially true when $\bm x = \bm 0$. In the rest, we assume $\bm x \ne \bm 0$. 
For ease of notation, we simply write $\bm M = \bm M_{\lambda,\beta}$, and
 \[
 \bm M = \bm M_\eps + \bm E_\eps, \qquad \bm X_{\bm L} =  \bm X_\eps + \bm X_{\bm L} \bm R_\eps.
 \] 
We consider two parameterized kernels as follows
\begin{enumerate}
\item $\bm M =  c{\bm D}^{-1/2}\bm X_{\bm L} {\bm D}^{1/2}$ where $c = \frac{2\lambda n}{n+\lambda - \beta n}$. 
Here,
\[
\bm X_{\bm L} = \alpha (\bm I - (1-\alpha) \bm W \bm D^{-1})^{-1}, \qquad \bm M_\eps = c \bm D^{-1/2} {\bm X}_\eps \bm D^{1/2}, \qquad \bm E_\eps = c \bm D^{-1/2} \bm X_{\bm L} \bm R_\eps\bm D^{1/2}.
\]

 We have
\begin{align*}
\bm x^\top \bm M \bm x &= \bm x^\top \bm M_\eps \bm x + \bm x^\top \bm E_\eps \bm x\\
&= \bm x^\top \left( \frac{\bm M_\eps + \bm M_\eps}{2}^\top \right) \bm x + \frac{1}{2}\bm x^\top \left(\bm M \bm D^{-1/2} \bm R_\eps \bm D^{1/2} + \bm D^{1/2} \bm R_\eps^\top \bm D^{-1/2} \bm M \right) \bm x\\
&= \bm x^\top \left( \frac{\bm M_\eps + \bm M_\eps}{2}^\top \right) \bm x + \bm x^\top \bm M \bm x \cdot \frac{\bm x^\top \left(\bm M \bm D^{-1/2} \bm R_\eps \bm D^{1/2} +  \bm D^{1/2} \bm R_\eps^\top \bm D^{-1/2} \bm M \right) \bm x}{2 \bm x^\top \bm M \bm x} \\ 
&\leq \bm x^\top \left( \frac{\bm M_\eps + \bm M_\eps}{2}^\top \right) \bm x + \bm x^\top \bm M \bm x \cdot  \|\bm D^{-1/2} \bm R\bm D^{1/2}\|_2,
\end{align*}
where the last inequality follows from Lemma \ref{lemma:bound-residual-matrix} of \eqref{inequ:r-mat-error-bound}. Rearrange the above; we finish the first case.

\item $ \bm M = 2\lambda {\bm S}^{-1/2}\bm X_{\mc L} {\bm S}^{1/2}$. Here, 
\[
\bm X_{\mc{\bm L}} = (\alpha \bm I + \bm D - \bm W)^{-1}  \qquad \bm M_\eps = 2\lambda\bm S^{-1/2}\bm X_\eps\bm S^{1/2}, \qquad \bm E_\eps = 2\lambda\bm S^{-1/2} \alpha \bm X_{\mc{\bm L}} \bm R_\eps\bm S^{1/2}.
\]
Then
\begin{align*}
\bm x^\top \bm M \bm x &= \bm x^\top  (\bm M_{\eps} + \bm E_\eps) \bm x \\
&= \bm x^\top \left( \frac{\bm M_\eps + \bm M_\eps}{2}^\top \right)\bm x + \frac{\alpha}{2}\bm x^\top \left(\bm M \bm S^{-1/2} \bm R_\eps \bm S^{1/2} + \bm S^{1/2} \bm R_\eps^\top \bm S^{-1/2} \bm M \right) \bm x\\
&= \bm x^\top \left( \frac{\bm M_\eps + \bm M_\eps}{2}^\top \right) \bm x + \bm x^\top \bm M \bm x \cdot  \frac{ \alpha \bm x^\top \left(  \bm M \bm S^{-1/2} \bm R_\eps \bm S^{1/2} +  \bm S^{1/2} \bm R_\eps^\top \bm S^{-1/2} \bm M \right) \bm x}{2 \bm x^\top \bm M \bm x} \\ 
&\leq \bm x^\top \left( \frac{\bm M_\eps + \bm M_\eps}{2}^\top \right) \bm x + \bm x^\top \bm M \bm x \cdot\alpha \|\bm S^{-1/2} \bm R\bm S^{1/2}\|_2,
\end{align*}
by applying Lemma \ref{lemma:bound-residual-matrix} with $\bm D = \bm S^{1/2}$. 
\end{enumerate}
Rearrange the above; we finish the proof.
\end{proof}
\begin{remark}
The above theorem allows us to control the error of $\bm M_\eps$. Next, we show that if $\frac{\bm M_\eps + \bm M_\eps^\top}{2}$ is a positive semidefinite matrix, then we can find an approximate relaxation function.
\end{remark}

\begin{lemma}
Let $\bm G^t = [\bm \nabla_1,\ldots,\bm \nabla_t, \bm 0,\ldots, \bm 0] \in \mathbb{R}^{k\times n}$ with $\|\bm \nabla_t \|_2\leq D$. Let $\bm M_\eps$ be the approximation built upon $\bm X_\eps$ as defined in Table \ref{tab:graph-kernel-approximation} such that $\left( \frac{\bm M_\eps + \bm M_\eps^\top}{2}\right)$ is positive semidefinite. If we provide the following score
\begin{equation}
\bm \psi_{t} = - \frac{\bm G^{t-1}(\bm M_\eps)_{:,t} + \bm G^{t-1}(\bm M_\eps)_{t,:}^\top}{\sqrt{Q_{t-1} + D^2 \sum_{j=t}^n (M_\eps)_{j,j}}}, \nonumber
\end{equation}
where $Q_t \triangleq \sum_{i=1}^k \left(\bm G_i^t\right)^{\top} \left(\frac{\bm M_\eps + \bm M_\eps^\top}{2}\right) \bm G_i^{t}$, then the relaxation function defined 
\begin{equation}
\operatorname{Rel}_t \left(\bm \nabla_{1}, \ldots, \bm \nabla_{t}; \bm M_\eps\right) \triangleq \sqrt{ Q_t + D^{2} \sum_{j=t+1}^{n} (M_\eps)_{j, j}} \nonumber    
\end{equation}
is admissible; that is, for all $t=1,2,\ldots,n$,
\[
\inf_{\bm \psi_{t} \in \mathbb{R}^{k}} \sup_{\|\bm \nabla_{t}\| \leq D}\left\{ \bm \nabla_t^\top \bm \psi_t + \rel_t\left(\bm\nabla_1 \ldots, \bm \nabla_{t};\bm M_\eps\right)\right\} \leq \rel_{t-1}(\bm \nabla_1, \ldots, \bm \nabla_{t-1}; \bm M_\eps)
 \]
 where $\rel_0(\emptyset;\bm M_\eps) = \sqrt{D^2\cdot \tr{\bm M_\eps}}$.
\label{lemma:relaxed-score}
\end{lemma}
\begin{proof}
Define $\bm b_i^t = \left[0,\ldots,\nabla_t(i),0,\ldots,0\right]^\top \in {\mathbb R}^n$. Note that $(\bm G_i^t)^\top = (\bm G_i^{t-1} + \bm b_i^t)^\top$ where $ i = 1,2,\ldots, k$ indexing label id and $t = 1,2,\ldots, n$ indexing node id. We define the following quadratic based on $\bm M_\eps$
\begin{align*}
Q_t &\triangleq \sum_{i=1}^k \left(\bm G_i^t\right)^{\top} \left(\frac{\bm M_\eps + \bm M_\eps^\top}{2}\right) \bm G_i^{t} \\
&= \sum_{i=1}^k \left(\bm G_i^{t-1} + \bm b_i^t\right)^{\top}\left(\frac{\bm M_\eps + \bm M_\eps^\top}{2}\right) \left( \bm G_i^{t-1} + \bm b_i^t \right) \\
&= \sum_{i=1}^k \left\{\left(\bm G_i^{t-1}\right)^\top \left(\frac{\bm M_\eps + \bm M_\eps^\top}{2}\right)\bm G_i^{t-1} + \left(\bm G_i^{t-1}\right)^\top \left(\bm M_\eps + \bm M_\eps^\top\right) \bm b_i^t + (\bm b_t^i)^{\top} \left(\frac{\bm M_\eps + \bm M_\eps^\top}{2}\right) \bm b_t^i \right\} \\
&= Q_{t-1} + \sum_{i=1}^k \left\{ \left(\bm G_i^{t-1} \right)^\top\bm M_\eps \bm b_i^t + (\bm b_i^t)^{\top} \bm M_\eps \bm G_i^{t-1} + \left(\nabla_t(i)\right)^2 (M_\eps)_{t,t} \right\} \\
&= Q_{t-1} + \sum_{i=1}^k \left( \nabla_t(i)((\bm G_i^{t-1})^\top (\bm M_\eps)_{:,t} + (\bm M_\eps )_{t,:}^\top \bm G_i^{t-1}) + (\nabla_t(i))^2 (M_\eps)_{t,t} \right),
\end{align*}
where ${(\bm M_\eps)}_{t,:}^\top$ is the transpose of $t$-th row vector of $\bm M_\eps$ and ${(\bm M_\eps)}_{:,t}$ is $t$-th column vector of $\bm M_\eps$. When $t=1$, we initialize $Q_0 = 0$ and $\bm G^0 = \bm 0_{k\times n}$. Finally, the recursion of the above is,
\begin{align*}
Q_t &= Q_{t-1} + \bm \nabla_t^\top \bm G^{t-1} {(\bm M_\eps)}_{:,t} + \bm \nabla_t^\top \bm G^{t-1} {(\bm M_\eps)}_{t,:}^\top + (M_\eps)_{t,t} \cdot \|\bm \nabla_t\|_2^2,
\end{align*}
where $t=1,2,\ldots,n$. Now we can obtain an upper bound of the relaxation function $\rel_t$ as the following ($t=0, 1,2,\ldots,n-1$)
\begin{align}
\rel_{t} \left(\bm \nabla_{1}, \ldots, \bm \nabla_{t};\bm M_\eps\right) &\triangleq \sqrt{ Q_t + D^{2} \sum_{j=t+1}^{n} (M_\eps)_{j, j} } \nonumber\\
&=\sqrt{  Q_{t-1} + \bm \nabla_t^\top \bm G^{t-1} {(\bm M_\eps)}_{:,t} + \bm \nabla_t^\top \bm G^{t-1} {(\bm M_\eps)}_{t,:}^\top + (M_\eps)_{t,t} \cdot \|\bm \nabla_t\|^2 + D^{2} \sum_{j=t+1}^{n} (M_\eps)_{j, j} } \nonumber\\
&\leq \sqrt{Q_{t-1} + \bm \nabla_t^\top \bm G^{t-1} {(\bm M_\eps)}_{:,t} + \bm \nabla_t^\top \bm G^{t-1} {(\bm M_\eps)}_{t,:}^\top + D^{2} \sum_{j=t}^{n} M_\eps (j, j)} ,\nonumber
\end{align}
where the above inequality step is due to $\|\bm \nabla_t\|_2^2 \leq D^2$. Hence, we have
\begin{small}
\begin{align*}
&\inf_{\bm \psi_{t} \in \mathbb{R}^{k}}
\sup_{\|\bm \nabla_{t}\| \leq D}\left\{ \bm \nabla_t^\top \bm \psi_t + \rel_{n}\left(\bm \nabla_{1}, \ldots, \bm \nabla_{t};\bm M_\eps \right)\right\} \\
&\leq \inf_{\bm \psi_{t} \in \mathbb{R}^{k}} \sup _{\|\bm \nabla_{t}\| \leq D} \left\{\bm \nabla_{t}^{\top} \bm \psi_{t} +  \sqrt{Q_{t-1} + \bm \nabla_t^\top \bm G^{t-1} {(\bm M_\eps)}_{:,t} + \bm \nabla_t^\top \bm G^{t-1} {(\bm M_\eps)}_{t,:}^\top + D^{2} \sum_{j=t}^{n} (M_\eps)_{j, j}  } \right\}\\
&\leq \sup_{\left\|\bm \nabla_{t}\right\| \leq D} \left\{ - \frac{\bm \nabla_t^\top \bm G^{t-1} {(\bm M_\eps)}_{:,t} + \bm \nabla_t^\top \bm G^{t-1} {(\bm M_\eps)}_{t,:}^\top}{\sqrt{Q_{t-1} + D^2 \sum_{j=t}^n M_\eps(j,j)  }} + \sqrt{Q_{t-1}+ \bm \nabla_t^\top \bm G^{t-1} {(\bm M_\eps)}_{:,t} + \bm \nabla_t^\top \bm G^{t-1} {(\bm M_\eps)}_{t,:}^\top + D^{2} \sum_{j=t}^{n} (M_\eps)_{j, j}  } \right\} \\
&\leq \sqrt{Q_{t-1} + D^2 \sum_{j=t}^n M_\eps(j,j)} = \rel_n(\bm \nabla_1,\ldots, \bm \nabla_{t-1};\bm M_\eps),
\end{align*}
\end{small}
where the first inequality is due to the upper bound of $\rel_n$, and the second is that we replace $\bm \psi_t$ by its definition. To see the last inequality, we can prove it in the following way: Letting $\bm v =  \bm G^{t-1} {(\bm M_\eps)}_{:,t} + \bm G^{t-1} {(\bm M_\eps)}_{t,:}^\top, a = Q_{t-1} + D^2 \sum_{j=t}^n M_\eps(j,j)$, we can simplify the above equality as 
\begin{equation}
\sup_{\|\bm \nabla_t\| \leq D} \left\{ h(\bm \nabla_t) := - \frac{\bm \nabla_t^\top \bm v}{\sqrt{a}} + \sqrt{ a + 2 \bm \nabla_t^\top \bm v} \right\}. \label{sup-problem}
\end{equation}
The function $h$ is concave in $\bm \nabla_t$, and setting its gradient to 0 gives
\[
\frac{\bm v}{\sqrt{a}} = \frac{\bm v}{\sqrt{a + \bm \nabla_t^T\bm v}} \iff \bm \nabla_t^T\bm v = 0
\]
which is feasible in the domain $\|\bm \nabla_t\|\leq D$.
In other words, $h(\bm \nabla_t)\leq \sqrt{a}$, for all $\bm \nabla_t \in {\mathbb R}^k$. This upper bound is always achievable by noticing that there exists $\bm \nabla_t$ such that $\bm \nabla_t^\top \bm v = 0$.
\end{proof}
\begin{remark}
The main argument of the above proof follows from \citet{rakhlin2017efficient}. However, this proof differs in a way that unlike in  \citet{rakhlin2017efficient}, we assume $\frac{\bm M_\eps + \bm M_\eps^\top}{2}$ is the one such that all eigenvalues are nonnegative. Even if $\bm M_\eps$ is asymmetric, we can still find a relaxation function accordingly. 
\end{remark}

\section{Regret based on the estimation of $\bm M_{\lambda,\beta}$}
\label{sect:appendix:regret-bound}
Recall we consider the following online learning paradigm on $\mc{G}$: At each time $t$, a learner picks up a node $v$ and makes a prediction $\hat{y}_v \in \mathcal{Y}$; then the true label $y_v$ is revealed with a cost of corresponding 0-1 loss as $\ell(\hat{\bm y}, \bm y) = 1 - {\bm y}^\top \hat{\bm y}$, the goal is to design an algorithm so that the learner makes as few mistakes as possible.  Denote a prediction of $\mc{V}$ as $\widehat{\bm Y} = \left[\hat{\bm y}_1, \hat{\bm y}_2, \ldots, \hat{\bm y}_n \right] \in \mc{F}$ and true label configuration as ${\bm Y} = \left[{\bm y}_1, {\bm y}_2, \ldots, {\bm y}_n \right] \in \mc{F}$ where the set of allowed label configurations $\mc{F} \triangleq \left\{ \bm F \in \{0,1\}^{k\times n} : {\bm F}_{:,j}^\top
\cdot {\bm 1} = 1, \forall j \in \mc{V}\right\}$. Formally, the goal is to find an algorithm $\mathcal{A}$, which minimizes the following regret 
\begin{equation}
\mathop{\regret}_{\mc{A},\bm Y}(\mc{F}) := \mathop{\mathbb{E}}_{\widehat{\bm Y} \sim \mathcal{A}} \sum_{t=1}^{n} \ell(\hat{\bm y}_t, \bm y_t) - \min_{\bm F \in \mc{F}} \sum_{t=1}^{n} \ell(\bm f_t, \bm y_t), \nonumber
\end{equation}
where the graph Laplacian constraint set is defined as
$\mc{F} \triangleq \mc{F}_{\lambda,\bm K} = \left\{ \bm F \in \mc{F}: \sum_{i=1}^k {\bm F}_{i}^\top {\bm K}^{-1} {\bm F}_{i} \leq \lambda \right\}$.

Before we present the main theorem, we shall state the important properties of $\phi(\cdot,\bm y)$ defined in \eqref{equ:surrogate-loss}. We repeat these lemmas and their proofs as the following. In the rest, we denote $\xi$ for $\phi$ when $y_t \notin S(\bm \psi_t))$.

\begin{lemma}{\cite{rakhlin2017efficient}}
If we use the loss $\phi$ defined in \eqref{equ:surrogate-loss}, then we have the regret upper bounded by
\begin{align}
\mathop{\regret}_{\mc{A},\bm Y}(\mc{F}) &\leq \sum_{t=1}^{n} \phi_{\bm \psi_{t}}\left(\bm \psi_{t}, \bm y_{t}\right)-\inf _{\bm f \in \mc{F}} \sum_{t=1}^{n} \phi_{\bm \psi_{t}}\left(\bm f_{t}, \bm y_{t}\right) \nonumber\\
&\leq \sum_{t=1}^{n} \bm \nabla \phi_{\bm \psi_{t}}\left(\bm \psi_t,  \bm y_{t}\right)^{\top}\bm \psi_{t}-\inf_{f\in \mc{F}} \sum_{t=1}^{n} \bm \nabla \phi_{\psi_{t}}\left(\bm \psi_t,  \bm y_{t}\right)^{\top}\bm f_{t}  \triangleq B_n \label{main-inequ}
\end{align}
\label{lemma-4.1}
\end{lemma}
\begin{proof}
For any time $t$, we have $\phi_{\bm \psi_t}(\bm \psi_t, \bm y_t)$ satisfies the following inequality \cite{rakhlin2017efficient}
\begin{align}
&\mathbb{E}_{\widehat{\bm y}_t \sim \bm q_t(\bm \psi_t)}\left[{1}\{\widehat{\bm y}_t \neq \bm y_t\}\right] -\left[\xi\left(\bm f_t, \bm y_t\right) {1}\{\bm y_t \notin \mc{S}(\bm \psi_t)\}+{1}\left\{\bm y \neq \bm f_t\right\} {1}\{\bm y_t \in \mc{S}(\bm \psi_t)\}\right] \nonumber\\
&\quad\leq \phi_{\bm \psi_t}(\bm \psi_t, \bm y_t)-\phi_{\bm \psi_t}\left(\bm f_t, \bm y_t\right),  \nonumber
\end{align}
where $\xi(\cdot,\cdot)$ denote $\phi$ when $\bm y_t \notin S(\bm \psi_t)$
Summing over $t$ from 1 to $n$, we obtain the first inequality of \eqref{main-inequ}. Notice $\phi_\psi(\bm g, \bm y)$ is a convex function over $\bm g$. Specifically, for any $\bm \psi$ and $\bm y$ and any $\bm g, \bm h \in \mathbb{R}^k$, we have that
\begin{equation}
\phi_{\bm \psi}(\bm g, \bm y)-\phi_{\psi}(\bm h, \bm y) \leq \bm \nabla_{\bm g} \phi_{\psi}(\bm g,\bm  y)^{\top}(\bm g-\bm h). \nonumber
\end{equation}
Let $\bm g = \bm \psi_t$ and $\bm h = \bm f_t$, we obtain the following
\begin{equation}
\sum_{t=1}^{n} \phi_{\bm \psi_{t}}\left(\bm \psi_{t}, \bm y_{t}\right)- \sum_{t=1}^{n} \phi_{\bm \psi_{t}}\left(\bm f_{t}, \bm y_{t}\right) \leq \sum_{t=1}^{n} \nabla \phi_{\bm \psi_{t}}\left(\bm \psi_t,  \bm y_{t}\right)^{\top} \left(\bm \psi_{t}-\bm f_{t}\right). \nonumber
\end{equation}
Taking the sup over both sides, we have
\begin{align*}
\sup_{\bm F\in \mc{F}} \left\{ \sum_{t=1}^{n} \phi_{\psi_{t}}\left(\bm \psi_{t},\bm  y_{t}\right)- \sum_{t=1}^{n} \phi_{\bm \psi_{t}}\left(\bm f_{t}, \bm y_{t}\right) \right\} &\leq \sup_{\bm F\in \mc{F}} \left\{ \sum_{t=1}^{n} \bm \nabla \phi_{\bm \psi_{t}}\left(\bm \psi_t, \bm  y_{t}\right)^{\top} \left(\bm \psi_{t}-\bm f_{t}\right) \right\} \\
 \sum_{t=1}^{n} \phi_{\bm \psi_{t}}\left(\bm \psi_{t}, \bm y_{t}\right)- \inf_{\bm F\in \mc{F}} \sum_{t=1}^{n} \phi_{\bm \psi_{t}}\left(\bm f_{t}, \bm y_{t}\right) &\leq \sup_{\bm F\in \mc{F}} \left\{\sum_{t=1}^{n} \bm \nabla \phi_{\psi_{t}}\left(\bm \psi_t,  \bm y_{t}\right)^{\top} \left(\bm \psi_{t}-\bm f_{t}\right) \right\}\\
 \sum_{t=1}^{n} \phi_{\bm \psi_{t}}\left(\bm \psi_{t}, \bm y_{t}\right)- \inf_{\bm F\in \mc{F}} \sum_{t=1}^{n} \phi_{\bm \psi_{t}}\left(\bm f_{t}, \bm y_{t}\right) &\leq  \sum_{t=1}^{n} \bm \nabla \phi_{\bm \psi_{t}} \left(\bm \psi_t,  \bm y_{t}\right)^{\top} \bm \psi_{t} -  \inf_{\bm F\in \mc{F}} \sum_{t=1}^{n} \bm \nabla \phi_{\bm \psi_{t}} \left(\bm \psi_t,  \bm y_{t}\right)^{\top} \bm f_{t}.
\end{align*}
We finish the proof.
\end{proof}
The above lemma tells us that there is a way to use surrogate loss $\phi_\psi$ to possibly obtain a regret if $\sum_{t=1}^n \bm \nabla_t^\top (\bm \psi_t - \bm F_{:,t})$ can be properly bounded. The next lemma tells us that if we can properly choose a method such that $\bm \psi_t$ satisfies the above lemma, then we have the following lemma. Following the notation in \cite{rakhlin2017efficient}, we define $B_n$ as in \eqref{main-inequ}.
\begin{lemma}{\cite{rakhlin2017efficient}}
If the loss is defined in \eqref{equ:surrogate-loss}, we obtain the regret bound
\begin{equation}
\begin{aligned}
\sum_{t=1}^{n} \underset{\widehat{\bm y}_{t} \sim \bm q\left(\bm \psi_{t}\right)}{\mathbb{E}} 1\left\{\widehat{\bm y}_{t} \neq \bm y_{t}\right\} \leq \inf _{\bm F \in \mc{F}} \left\{ 2\left(1-\frac{1}{k}\right) \sum_{t: \bm y_{t} \notin S\left(\bm \psi_{t}\right)}1\left\{\bm f_{t} \neq \bm y_{t}\right\} +\sum_{t: \bm y_{t} \in S\left(\bm \psi_{t}\right)} 1\left\{\bm f_t\neq \bm y_{t} \right\}\right\}+ B_{n}. \nonumber
\end{aligned}
\end{equation}
\end{lemma}
\begin{proof}
From the Lemma \ref{lemma-4.1}, we have
\begin{equation}
\begin{aligned}
\sum_{t=1}^{n} \underset{\widehat{\bm y}_{t} \sim \bm q\left(\bm \psi_{t}\right)}{\mathbb{E}} 1\left\{\widehat{\bm y}_{t} \neq \bm y_{t}\right\} &\leq \inf _{\bm F \in \mc{F}} \left\{\sum_{t: \bm y_{t} \notin S\left(\bm \psi_{t}\right)} \xi\left(\bm f_{t}, \bm y_{t}\right) +\sum_{t: \bm y_{t} \in S\left(\bm \psi_{t}\right)} 1\left\{\bm f_{t} \neq \bm y_{t}\right\}\right\}+B_{n} \nonumber\\
&= \inf_{\bm F \in \mc{F}} \left\{\sum_{t: \bm y_{t} \notin S\left(\bm \psi_{t}\right)} \xi\left(\bm f_{t}, \bm y_{t}\right) \underset{\widehat{\bm y}_{t} \sim \bm q\left(\bm \psi_{t}\right)}{\mathbb{E}} 1\left\{\widehat{\bm y}_{t} \neq \bm y_{t}\right\} +\sum_{t: \bm y_{t} \in S\left(\bm \psi_{t}\right)} 1\left\{\bm f_{t} \neq \bm y_{t}\right\}\right\}+B_{n},
\end{aligned}
\end{equation}
where the equality holds is due to the fact that $\underset{\widehat{\bm y}_{t} \sim \bm q \left( \bm \psi_{t} \right)}{\mathbb{E}} 1\left\{\widehat{\bm y}_{t} \neq \bm y_{t}\right\}=1$ when $\bm y_t \notin S(\bm \psi_t)$.  Furthermore, when $\bm y_t \notin S(\bm \psi_t)$, $\xi(\bm f_t, \bm y_t)\geq 1$ and for any $j \in \mathcal{Y}, \bm y_t \notin S(\bm \psi_t)$, we have
\begin{align*}
\xi\left(\bm f_t, \bm y_t\right) & = \frac{1 + \max_{r:\bm e_r \ne \bm y_t} \left\{\bm f_t^\top \bm e_r - \bm f_t^\top \bm y_t \right\}}{1+1/|S(\bm \psi_t)|} \nonumber\\
&= \frac{2\cdot 1\{\bm f_t \ne \bm y_t\}}{1+1/|S(\bm \psi_t)|} \\
&\leq 2\left(1-\frac{1}{k}\right) 1\left\{\bm f_t \neq \bm y_t\right\},
\end{align*}
where the last inequality is due to $|S(\bm \psi_t)| \leq k-1$ when $\bm y_t \notin S(\bm \psi_t)$.
\end{proof}

\begin{algorithm}[H]
\caption{\textsc{Relaxation}$(\mc{G}, \lambda, \bm K_\beta^{-1})$ \cite{rakhlin2017efficient}}
\begin{algorithmic}[1]
\STATE $T_1 = \tr{(\bm M)}, A_1 = 0, \bm G = [\bm 0, \bm 0, \ldots, \bm 0] \in \mathbb{R}^{k\times n}, \bm M_{\lambda,\beta} = \left( \frac{\bm K_{\beta}^{-1}}{2\lambda} + \frac{\bm I}{2 n}\right)^{-1}$
\FOR{$t = 1,\ldots, n$}
\STATE $\bm \psi_{t}= - \bm G^t \bm M[:,t]  / \sqrt{A_t + D^2 \cdot T_t}$
\STATE Predict $\hat{y}_t \sim q_t(\bm \psi_t)$ and get loss gradient $\bm \nabla_t  = \begin{cases} \frac{\max_{r: \bm e_r \ne \bm y} \left\{ \bm e_r - \bm y \right\}}{1+1 /|\mathcal{S}(\psi)|}  & \text { if } y_{t} \notin \mathcal{S}\left(\psi_{t}\right) \\ \frac{1}{\left|\mathcal{S}\left(\psi_{t}\right)\right|} \mathbf{1}_{\mathcal{S}\left(\psi_{t}\right)}- \bm y_{t} & \text { otherwise }\end{cases}$
\STATE Update $\bm G[:,t] = \bm \nabla_t$
\STATE $A_{t+1} = A_{t} + 2 \bm \nabla_t^\top \bm G^t \bm M[:,t] + m_{tt}\cdot \| \bm \nabla_t \|^2$
\STATE $T_{t+1} = T_t - m_{tt}$
\ENDFOR
\end{algorithmic}
\label{algo:relaxation-repeat}
\end{algorithm}

If we use the above Algorithm \ref{algo:relaxation-repeat}, then we have the following lemmas directly from \cite{rakhlin2017efficient}

\begin{lemma}{\cite{rakhlin2017efficient}}
Let $\bm G = [\bm \nabla_1, \ldots, \bm \nabla_n] \in \mathbb{R}^{k\times n}$. We have the following
\begin{equation}
- \inf_{ {\bm F} \in \bar{\mathcal{F}}_{\lambda,\bm K}} \sum_{t=1}^n {\bm \nabla}_t^\top {\bm f}_t =\sqrt{\sum_{i=1}^{k} {\bm G}_{i}^{\top} \bm M_{\lambda,\beta} {\bm G}_{i}},
\end{equation}
where the pre-computed matrix is $\bm M_{\lambda,\beta} = \left( \frac{{\bm K}_\beta^{-1}}{2\lambda} + \frac{\bm I_n}{2n} \right)^{-1}$. Then we could have
\begin{equation}
- \inf_{ \bm F \in \mc{F}_{\lambda,\bm K}} \sum_{t=1}^n {\bm \nabla}_t^\top \bm f_{:,t} \leq - \inf_{ \bm F \in \bar{\mc{F}}_{\lambda, {\bm K}}} \sum_{t=1}^n \bm \nabla_t^\top \bm F_{:,t}.
\end{equation}
\end{lemma}
The above lemma is from the following fact:
\begin{lemma}
\label{lemma:4.4}
Consider the following optimization problem
\begin{equation}
\min_{\bm F \in \mathbb{R}^{k\times n}} \tr{\bm F \bm Y^\top} \text{ subject to } \tr{\bm F \bm M^{-1} \bm F^\top}\leq 1, \label{equ:trace-opt}
\end{equation}
where $\bm M \in \mathbb{R}^{n\times n}$ is a symmetric positive definite matrix and $\bm Y \in \mathbb{R}^{k\times n}$. \eqref{equ:trace-opt} obtains the optimal at $F^* = - \frac{\bm Y \bm M}{\sqrt{\tr{\bm Y \bm M \bm Y^\top}}}$, that is
\begin{equation}
\min_{\bm F \in \R^{k\times n}} \tr{\bm F \bm Y^\top} \geq - \sqrt{\tr{\bm Y \bm M \bm Y^\top}} \text{ with equality holds at } \bm F^*.
\end{equation}
\end{lemma}
\begin{proof}
A possible Lagrangian can be defined as
\begin{equation}
\mc L(\bm F;\lambda) = \tr{\bm F\bm Y^T} + \lambda(\tr{\bm F\bm M^{-1} \bm F^T}-1)
\end{equation}
and differentiating and setting the gradient to be 0, we have
\begin{align*}
\nabla_F \mc \bm L &=  \bm Y + 2\lambda \bm F \bm M^{-1}  \\
\bm Y &= -2\lambda \bm F\bm M^{-1}.
\end{align*}
Denote $\lambda^* = \frac{\sqrt{\tr{\bm Y \bm M \bm Y^\top}} }{2}, \bm F^* = - \frac{\bm Y \bm M}{2\lambda^*}$. The rest of the proof is to show $(\bm F^*,\lambda^*)$ satisfies the KKT conditions. First of all, $\bm 0_{k\times n} \in \bm \nabla_F \mc{L}$ when $\bm F = \bm F^*$ hence stationarity is satisfied. $\lambda^* > 0$, so it is dual feasible. $\lambda^*\left(\tr{\bm F^* \bm M^{-1} {\bm F^*}^\top} -1\right) = 0$, so it satisfies complementary slackness. Clearly, $\tr{\bm F^* \bm M^{-1} {F^*}^\top}-1 \leq 0$, so it is primal feasible. Hence the problem obtains the optimal at $\bm F^*$.

By applying $\bm Y = \bm G$, we have that $\bm F^* = - \frac{\bm Y_n \bm M}{\sqrt{\tr{\bm Y_n \bm M \bm Y_n^\top}}}$.
\end{proof}
From the above lemma \ref{lemma:4.4}, we have the following upper bound of the defined regret.
\begin{lemma}{\cite{rakhlin2017efficient}}
The regret has the following bound
\begin{align*}
\mathop{\mathbb{E}}_{\widehat{\bm Y} \sim \mathcal{A}} \sum_{t=1}^{n} \ell(\hat{\bm y}_t, \bm y_t) - 2\left(1-\frac{1}{k}\right) \min_{\bm F \in \mc{F}_{\lambda,\bm K}} \sum_{t=1}^{n} \ell(\bm f_t, \bm y_t) \leq \sup_{\bm F\in \mc{F}_{\lambda,\bm K}} \left\{\sum_{t=1}^n \bm \nabla_1 \ell(\bm \psi_t, \bm y_t) ^\top(\bm \psi_t - \bm f_t) \right\}
\end{align*}
\end{lemma}

We are ready to prove our main theorem.
\begin{reptheorem}{thm:regret-analysis-relaxation}[Regret of \textsc{Relaxation} with parameterized Kernel matrix $\bm M_{\beta}^{-1}$]
Let $\hat{\bm Y}$ be the prediction matrix returned by \textsc{Relaxation}, if the input label sequence $\bm Y$ has good pattern, meaning a strong assumption $\lambda = n^\gamma$ with $\gamma \in (0, 1)$, then choosing $\beta = n^{\gamma -1}$ for kernel $\bm K_{\beta}^{-1} = \bm I - \beta \bm D^{-1/2}\bm W {\bm D}^{-1/2}$ and $\beta = 1-\frac{\lambda}{n}$ for $\bm K_\beta^{-1} = \beta \bm I + \bm S^{-1/2} \mathcal{L} \bm S^{-1/2}$, we have the following regret bound
\begin{equation}
\mathop{\mathbb{E}}_{\widehat{\bm Y} \sim \mathcal{A}} \sum_{t=1}^{n} \ell(\hat{\bm y}_t, \bm y_t) - 2\left(1-\frac{1}{k}\right) \min_{\bm F \in \mc{F}_{\lambda,\bm K}} \sum_{t=1}^{n} \ell(\bm f_t, \bm y_t) \leq D \sqrt{2 n^{1+\gamma}}.
\end{equation}
\label{thm:regret-analysis-relaxation-repeat}
\end{reptheorem}
\begin{proof}
Since $\mc{F}_{\lambda,\beta} \subseteq \bar{\mc{F}}_{\lambda,\beta}$, we continue to have an upper bound of the regret as 
\begin{align*}
\sup_{\bm F \in \mc{F}_{\lambda,\beta}} \left\{\sum_{t=1}^n \bm \nabla_1 \ell(\bm \psi_t,\bm y_t) ^\top(\bm \psi_t - \bm f_t) \right\} &\leq \sup_{\bm F \in \bar{\mc{F}}_{\lambda,\beta}} \left\{\sum_{t=1}^n \bm \nabla_1 \ell(\bm \psi_t, \bm y_t) ^\top(\bm \psi_t - \bm f_t) \right\} \\
&= \sum_{t=1}^n \bm \nabla_1 \ell(\bm \psi_t, \bm y_t) ^\top \bm \psi_t - \inf_{\bm F\in \bar{\mc{F}}_{\lambda,\beta}} \left\{\sum_{t=1}^n \bm \nabla_1 \ell(\bm \psi_t, \bm y_t)^\top \bm f_t \right\} \\
&= \sum_{t=1}^n \bm \nabla_t^\top \bm \psi_t - \inf_{ \bm F\in \bar{\mc{F}}_{\lambda,\beta}} \sum_{t=1}^n \bm \nabla_t^\top \bm f_t,
\end{align*}
where we denote $\bm \nabla_1 \ell(\bm \psi_t, \bm y_t)$ as $\bm \nabla_t$ in the last equality. For the term $\sum_{t=1}^n \bm \nabla_t^\top  \bm \psi_t$, from Lemma \ref{lemma:relaxed-score}, we know that if we choose
\begin{equation}
\bm \psi_{t} = - \frac{\bm G^{t-1} \bm M[t,:]}{\sqrt{ \sum_{i=1}^k \left(\bm G_i^{t-1} \right)^{\top} \bm M \bm G_i^{t-1}  + D^2 \sum_{j=t}^n M(j,j)}}, \nonumber
\end{equation}
where relaxation is defined as
\begin{equation}
\rel_t\left(\bm \nabla_{1}, \ldots, \bm \nabla_{t};\bm M\right) =  \sqrt{\sum_{i=1}^k \left(\bm G_i^{t-1} \right)^{\top} \bm M \bm G_i^{t-1}  + D^2 \sum_{j=t}^n M(j,j)} \nonumber
\end{equation}
satisfies
\begin{equation}
\bm \nabla_t^\top \bm \psi_t  \leq \rel_{t-1}(\bm \nabla_1,\ldots,\bm \nabla_{t-1};\bm M) - \rel_t(\bm \nabla_1,\ldots,\bm \nabla_t;\bm M), \quad \forall t = 1,2,\ldots,n. \nonumber
\end{equation}q
Then we continue to have
\begin{align*}
\sum_{t=1}^n \bm \nabla_t^\top \bm \psi_t &\leq \sum_{t=1}^n \left\{\rel_{t-1}(\bm \nabla_1,\ldots,\bm \nabla_{t-1}; \bm M) - \rel_t(\bm \nabla_1,\ldots,\bm \nabla_t; \bm M) \right\} \\
&= \rel_0(\emptyset; \bm M) - \rel_n(\bm \nabla_1,\ldots,\bm \nabla_n; \bm M) \\
&= D \sqrt{\sum_{i=1}^n m_{i i}} - \sqrt{\sum_{i=1}^{k} \left( \bm G_i^n \right)^{\top} \bm M G_i^n}.
\end{align*}
We are ready to provide the whole bound 
\begin{align*}
\regret(\mathcal{F}_{\lambda,{\bm K}}) &\leq \sum_{t=1}^n \bm \nabla_t^\top  \bm \psi_t - \inf_{\bm F\in \bar{\mc{F}}_{\lambda,\bm K}} \left\{\sum_{t=1}^n \bm \nabla_t^\top \bm f_t \right\} \\
&\leq D \sqrt{\sum_{i=1}^n m_{i i}} - \sqrt{\sum_{i=1}^{k} \left( \bm G_i^n \right)^{\top} \bm M \bm G_i^n} - \inf_{\bm F \in \bar{\mc F}_{\lambda,\beta}} \left\{\sum_{t=1}^n \bm \nabla_t^\top \bm f_t \right\} \\
&= D \sqrt{\sum_{i=1}^n m_{i i}} - \sqrt{\sum_{i=1}^{k} \left( \bm G_i^n \right)^{\top} \bm M \bm G_i^n} + \sqrt{\sum_{i=1}^{k} \left( \bm G_i^n \right)^\top \bm M \bm G_i^n} \\
&= D \sqrt{\sum_{i=1}^n m_{i i}}.
\end{align*}
\begin{enumerate}
\item For the first parameterized kernel $\bm K_{\beta}^{-1} = \bm I - \beta \bm D^{-1/2}\bm W {\bm D}^{-1/2}$, we have $\bm M_{\lambda,\beta} = \frac{2\lambda n}{ n + \lambda - n \beta } {\bm D}^{-1/2} \bm X_{\bm L} {\bm D}^{1/2}$ where $1-\alpha = \frac{n \beta}{n+\lambda}$, and the parameter $\beta \in (0,\frac{n+\lambda}{n})$. Note that the eigenvalue of $\bm X_{\bm L}$ be
\begin{equation}
\lambda(\bm X_{\bm L}) = \lambda\left( \alpha \left( \bm I - (1-\alpha)\bm W \bm D^{-1}\right)^{-1} \right) \in \left[\frac{\alpha}{2-\alpha}, 1\right]
\end{equation}
 We have
 \begin{align*}
\lambda(\bm M_{\lambda,\beta}) &= \frac{2\lambda n}{ n + \lambda - n \beta } \lambda(\bm X_{\bm L}) \in \frac{2\lambda n}{ n + \lambda - n \beta } \left[\frac{\alpha}{2-\alpha}, 1\right].
\end{align*}
Notice that when $\lambda = n^\gamma$ with $\gamma\in (0, 1)$, there always exists such $\beta$ so that $\lambda(\bm M_{\lambda,\beta})$ could be bounded by $2n^\gamma$, that is
\begin{align}
\lambda(\bm M_{\lambda,\beta}) &\leq \frac{2 \lambda n}{n+\lambda - n \beta} \\
&= \frac{2 \lambda }{1+\lambda/n - \beta} = \frac{2 n^{\gamma} }{1 + n^\gamma/n - \beta} = 2 n^{\gamma},
\end{align}
where we always choose $\beta = n^{\gamma -1}$. Note $\beta$ is a valid parameter for our problem setting where the kernel matrix requires $\beta \in (0,1)$ and $\beta \in (0,1+\frac{\lambda}{n})$. Notice that the trace of $\bm M$ is then bounded by
\begin{equation}
\tr{(\bm M_{\lambda,\beta})} \leq 2 n^{1+\gamma}.
\end{equation}
Therefore, we have
\begin{equation}
\regret(\mathcal{F}_{\lambda,\bm K}) \leq D \sqrt{\sum_{i=1}^n m_{i i}} \leq D \sqrt{ 2 n^{1+\gamma}}.
\end{equation}

\item Notice that our kernel matrix is $\bm K_\beta^{-1} = \beta \bm I + \bm S^{-1/2}\bm L \bm S^{-1/2}$ and corresponding ${\bm M}_{\lambda,\beta} = 2 \lambda \bm S^{-1/2} \bm X_{\mc L} \bm S^{1/2}$ with $\alpha = \tfrac{n\beta + \lambda}{n}$ (see Table \ref{tab:graph-kernel-presentation}). By Lemma \ref{lemma:eigenvalue-bound}, we have
\begin{align*}
\lambda(\bm X_{\mc{\bm L}}) &= \lambda\left( \left( \alpha \bm I + \bm D - \bm W\right)^{-1} \right) \in \left[ \frac{1}{\alpha + 2 D_{\max}}, \frac{1}{\alpha}\right] \\
\lambda(\bm M_{\lambda,\beta}) &= \lambda\left( 2 \lambda \bm S^{-1/2} \bm X_{\mc L} \bm S^{1/2}\right) \in \left[ \frac{2 \lambda}{\alpha + 2 D_{\max}}, \frac{2 \lambda}{\alpha}\right].
\end{align*}
Whenever $\lambda = n^\gamma$ with $\gamma \in (0, 1)$, we choose $\beta = 1 - \frac{\lambda}{n}$ so that $\alpha = 1$ and $\beta \in (0, 1)$. Hence, the trace of $\bm M_{\lambda,\beta}$ can be bound by
\begin{align*}
\tr{(\bm M_{\lambda,\beta})} = \sum_{i=1}^n \lambda_i(\bm M_{\lambda,\beta}) &\in \left[ \frac{2 n \lambda}{\alpha + 2 D_{\max}}, \frac{2 n \lambda}{\alpha}\right] \\
&\in \left[ \frac{2 n \lambda}{1 + 2 D_{\max}}, 2 n \lambda\right].    
\end{align*}
Therefore, we have
\begin{equation}
\regret(\mathcal{F}_{\lambda,\bm K}) \leq D \sqrt{\sum_{i=1}^n m_{i i}} \leq D \sqrt{ 2 n^{1+\gamma}}. \nonumber
\end{equation}
\end{enumerate}
\end{proof}
Our final step is to prove the approximated bound for \textsc{FastONL}. We state it as in the following theorem. In the rest, we simply define $M_\eps(i,i) = m_{i i}$.

\begin{reptheorem}{thm:regret-fastonl}[Regret analysis of \textsc{FastONL} with approximated parameterized kernel]
Consider \textsc{FastONL} presented in Algo. \ref{algo:fast-onl}. If we call \textsc{FastONL}$(\mc{G},\eps,\bm K_{\beta},\lambda=n^\gamma)$ and $\epsilon$ is chosen such that
\begin{align}
& \left\| \bm D^{-1/2} \bm R_\eps \bm D^{1/2} \right\|_2 \leq 1, \quad \alpha  \left\| \bm D^{-1/2} \bm R_\eps \bm D^{1/2} \right\|_2 \leq 1, \nonumber
\end{align}
then we have the following regret bounded by 
\begin{equation}
\regret := \mathop{\mathbb{E}}_{\widehat{\bm Y} \sim \mathcal{A}} \sum_{t=1}^{n} \ell(\hat{\bm y}_t, \bm y_t) - \frac{2k -1}{k} \min_{\bm F \in \mc{F}_{\lambda,\bm K}} \sum_{t=1}^{n} \ell(\bm f_t, \bm y_t) \leq D\sqrt{(1+k^2) n^{1+\gamma}}
\end{equation}
\label{thm:regret-apprximate-algo-repeat}
\end{reptheorem}

\begin{proof}
Since $\mc{F}_{\lambda,\bm K}$  has been relaxed to $\bar{\mc{F}}_{\lambda,\bm K}$, we have $\mc{F}_{\lambda,\bm K} \subseteq \bar{\mc{F}}_{\lambda,\bm K}$, the surrogate loss has been chosen, we continue to have an upper bound of the regret as 
\begin{align*}
\sup_{\bm F \in \mc{F}_{\lambda,\bm K}} \left\{\sum_{t=1}^n \bm \nabla_1 \ell(\bm \psi_t,\bm y_t) ^\top(\bm \psi_t - \bm f_t) \right\} &\leq \sup_{\bm F \in \bar{\mc{F}}_{\lambda,\bm K}} \left\{\sum_{t=1}^n \bm \nabla_1 \ell(\bm \psi_t, \bm y_t) ^\top(\bm \psi_t - \bm f_t) \right\} \\
&= \sum_{t=1}^n \bm \nabla_1 \ell(\bm \psi_t, \bm y_t) ^\top \bm \psi_t - \inf_{\bm F\in \bar{\mc{F}}_{\lambda,\bm K}} \left\{\sum_{t=1}^n \bm \nabla_1 \ell(\bm \psi_t, \bm y_t)^\top \bm f_t \right\} \\
&= \sum_{t=1}^n \bm \nabla_t^\top \bm \psi_t - \inf_{ \bm F\in \bar{\mc{F}}_{\lambda,\bm K}} \sum_{t=1}^n \bm \nabla_t^\top \bm f_t,
\end{align*}
where we denote $\bm \nabla_1 \ell(\bm \psi_t, \bm y_t)$ as $\bm \nabla_t$ in the last equality. For the term $\sum_{t=1}^n \bm \nabla_t^\top  \bm \psi_t$, from Lemma \ref{lemma:relaxed-score}, we know that if we choose
\begin{equation}
\bm \psi_{t} = - \frac{ \bm G^{t-1}{(\bm M_\eps)}_{:,t} + \bm G^{t-1}{(\bm M_\eps)}_{t,:}^\top}{\sqrt{ \sum_{i=1}^k \left(\bm G_i^{t-1} \right)^{\top} \bm M_\eps \bm G_i^{t-1}  + D^2 \sum_{j=t}^n M_\eps(j,j)}}, \nonumber
\end{equation}
where relaxation is defined as
\begin{equation}
\rel_t\left(\bm \nabla_{1}, \ldots, \bm \nabla_{t};\bm M_\eps\right) =  \sqrt{\sum_{i=1}^k \left(\bm G_i^{t-1} \right)^{\top} \bm M_\eps \bm G_i^{t-1}  + D^2 \sum_{j=t}^n M_\eps(j,j)} \nonumber
\end{equation}
satisfies
\begin{equation}
\bm \nabla_t^\top \bm \psi_t  \leq \rel_{t-1}(\bm \nabla_1,\ldots,\bm \nabla_{t-1};\bm M_\eps) - \rel_t(\bm \nabla_1,\ldots,\bm \nabla_t;\bm M_\eps), \quad \forall t = 1,2,\ldots,n. \nonumber
\end{equation}
Then we continue to have
\begin{align*}
\sum_{t=1}^n \bm \nabla_t^\top \bm \psi_t &\leq \sum_{t=1}^n \left\{\rel_{t-1}(\bm \nabla_1,\ldots,\bm \nabla_{t-1}; \bm M_\eps) - \rel_t(\bm \nabla_1,\ldots,\bm \nabla_t; \bm M_\eps) \right\} \\
&= \rel_0(\emptyset; \bm M_\eps) - \rel_n(\bm \nabla_1,\ldots,\bm \nabla_n; \bm M_\eps) \\
&= \sqrt{D^2 \sum_{j=t}^n M_\eps(j,j)} - \sqrt{\sum_{i=1}^{k} \left( \bm G_i^n \right)^{\top} \bm M_\eps G_i^n}.
\end{align*}
We are ready to provide the whole bound 
\begin{align*}
\regret(\mathcal{F}_{\lambda,\mathcal{K}}) &\leq \sum_{t=1}^n \bm \nabla_t^\top  \bm \psi_t - \inf_{\bm F\in \bar{\mc{F}}_{\lambda,\bm K}} \left\{\sum_{t=1}^n \bm \nabla_t^\top \bm f_t \right\} \\
&\leq D \sqrt{\sum_{i=1}^n M_\eps(i,i)} - \sqrt{\sum_{i=1}^{k} \left( \bm G_i^n \right)^{\top} \bm M_\eps \bm G_i^n} - \inf_{\bm F \in \bar{\mc F}_{\lambda,\beta}} \left\{\sum_{t=1}^n \bm \nabla_t^\top \bm f_t \right\} \\
&= D \sqrt{\sum_{i=1}^n M_\eps(i,i)} \underbrace{- \sqrt{\sum_{i=1}^{k} \left( \bm G_i^n \right)^{\top} \bm M_\eps \bm G_i^n} + \sqrt{\sum_{i=1}^{k} \left( \bm G_i^n \right)^\top \bm M \bm G_i^n}}_{E}.
\end{align*}
Notice that when $E\leq 0$, we have the regret $D \sqrt{\sum_{i=1}^n M_\eps(i,i)}$, which is less than $D \sqrt{\sum_{i=1}^n m_{i i}}$. In the rest, we will assume $E >0$ and show $E$ is not too large when $\bm R_\eps$ is small enough. 
\begin{align}
E &\leq \sqrt{\sum_{i=1}^{k} \left( \bm G_i^n \right)^\top \left( \bm M_{\lambda,\beta} - \bm M_\eps \right) \bm G_i^n} \nonumber\\
&= \sqrt{\sum_{i=1}^{k} (\bm G_i^n)^\top \bm G_i^n \cdot
\frac{ \left( \bm G_i^n \right)^\top \left( \bm M_{\lambda,\beta} - \bm M_\eps \right) \bm G_i^n}{ (\bm G_i^n)^\top\bm G_i^n}} \nonumber\\
&\leq \sqrt{k n D^2\cdot\sum_{i=1}^{k} 
\frac{\left( \bm G_i^n \right)^\top \left( \bm M_{\lambda,\beta} - \bm M_\eps \right) \bm G_i^n}{(\bm G_i^n)^\top\bm G_i^n}}, \label{inequ:last-inequality}
\end{align}
where the first inequality is because $\| \bm \nabla_t\|_2^2 \leq D^2$. Hence, $\| \bm G_i^n\|_2^2 \leq n D^2$. The last inequality is due to the Rayleigh quotient property. Recall from Thm. \ref{thm:bounding-error}, we have
\begin{equation}
\bm x^\top \bm M \bm x - \bm x^\top \bm M_\eps \bm x \leq \bm x^\top \bm M \bm x \cdot \sqrt{\frac{\lambda_{\max}(\bm M)}{\lambda_{\min}(\bm M)}} \|\bm D^{-1/2} \bm R\bm D^{1/2}\|_2 \leq \bm x^\top \bm M \bm x, \nonumber
\end{equation}
which means
\begin{equation}
\sum_{i=1}^k \frac{{\bm G_i^n}^\top \bm M {\bm G_i^n} - {\bm G_i^n}^\top \bm M_\eps {\bm G_i^n}}{{\bm G_i^n}^\top {\bm G_i^n}}  \leq \sum_{i=1}^k \frac{{\bm G_i^n}^\top \bm M {\bm G_i^n}}{{\bm G_i^n}^\top{\bm G_i^n}} \leq k \lambda_{\max}(\bm M) \leq k n^\gamma. \label{inequ:505}
\end{equation}
Combine \eqref{inequ:last-inequality} and \eqref{inequ:505}, we have
\begin{equation}
\regret(\mathcal{F}_{\lambda,\bm K}) \leq \sqrt{D^2\sum_{i=1}^n M_\eps(i,i) + D^2 k^2 n^{1+\gamma}}  \leq D\sqrt{(1+k^2) n^{1+\gamma}}. \nonumber
\end{equation}
\end{proof}

\subsection{Practice Implementation}

In practice, we use $T_0 = k \cdot n^2$ and use $\bm x_t$ to estimate the both $t$-th column and row vector of $(\bm M_\eps + \bm M_\eps^\top)/2$. This upper bound works well in all our experiments. In the following, $m_{t t} =  (M_\eps)_{t t}$. 
\begin{algorithm}[H]
\caption{\textsc{FastONL}($\mc{G},\epsilon,\bm K_\beta^{-1},\lambda, \Gamma$)}
\begin{algorithmic}[1]
\STATE $\bm G^0 = [\bm 0, \bm 0, \ldots, \bm 0] \in \mathbb{R}^{k\times n}$
\STATE $A_1 = 0$ and let $\alpha,\bm M_\eps$ be defined in Table \ref{tab:graph-kernel-approximation}
\STATE $T_1 =\Gamma$ // Big enough
\FOR{$t = 1, 2,\ldots, n$}
\STATE $\bm x_t,\bm r_t = $\textsc{FIFOPush}$(\mathcal{G},\eps,\alpha, t)$
\STATE Compute $(\bm M_\eps)_{:,t}$ on based on Tab. \ref{tab:graph-kernel-presentation}
\STATE $\bm \psi_{t}= - \bm G^t {(\bm M_\eps)}_{:,t} / \sqrt{A_t + k \cdot T_{t}} $
\STATE $\bm \nabla_t  = \begin{cases} \frac{\max_{r: \bm e_r \ne \bm y} \left\{ \bm e_r - \bm y \right\}}{1+1 /|\mathcal{S}(\psi)|}  & \text { if } y_{t} \notin \mathcal{S}\left(\psi_{t}\right) \\ \frac{1}{\left|\mathcal{S}\left(\psi_{t}\right)\right|} \mathbf{1}_{\mathcal{S}\left(\psi_{t}\right)}- \bm y_{t} & \text { otherwise }\end{cases}$
\STATE Update gradient $\bm G^t_{:,t} = \bm \nabla_t$
\STATE $A_{t+1} = A_{t} + 2 \bm \nabla_t^\top \bm v + m_{tt}\cdot \| \bm \nabla_t \|_2^2$
\STATE $T_{t+1} = T_{t} - m_{tt}$
\ENDFOR
\end{algorithmic}
\label{algo:fast-onc-approx}
\end{algorithm}

\section{More experimental details and results}
\label{sect:appendix:experiments}

\subsection{Experimental setups}

We implemented all methods using Python and used the inverse function of scipy library to compute the matrix inverse. All experiments are conducted on a server with 40 cores and 250GB of memory.

\subsection{Dataset Description}
\begin{table*}[ht]
\centering
\caption{Datasets Statistics}
\begin{tabular}[t]{l>{\raggedright}p{0.1\linewidth}>{\raggedright\arraybackslash}p{0.1\linewidth}p{0.1\linewidth}p{0.1\linewidth}p{0.1\linewidth}}
\toprule
& $|\mc{V}|$ & $\left|\mc{L}\right|$ & $|\mc{E}|$ & $|\mc{Y}|$ & Weighted \\
\midrule
Political & 1,222 & 1,222 & 16,717 & 2 & No \\
Cora & 2,485 & 2,485 & 5,069 & 7 & No \\
Citeseer & 2,110 & 2,110 & 3,668 & 6 & No \\
PubMed & 19,717 & 19,717 & 44,324 & 3 & No \\ 
MNIST & 12,000 & 12,000 & 97,089 & 10 & Yes \\ 
BlogCatalog & 10,312 & 10,312 & 333,983 & 39 & No \\\hline
Flickr & 80,513 & 80,513 & 5,899,882 & 195 & No \\
OGB-Arxiv & 169,343 & 169,343 & 1,157,799 & 40 & No \\
YouTube & 1,134,890 & 31,684 & 2,987,624 & 47 & No \\
OGB-Products & 2,385,902 & 500,000 & 61,806,303 & 47 & No \\\hline
Wikipedia & 6,216,199 & 150,000 & 177,862,656 & 10 & No\\
\bottomrule
\end{tabular}
\label{tab:datasets}
\end{table*}

We list all ten graph datasets in Tab. \ref{tab:datasets}. where $|\mc L|$ is the number of available labeled nodes.

\begin{enumerate}
\item Political \cite{adamic2005political}. This political blog graph contains 1,490 nodes and 16,715 edges. Each node represents a web blog on US politics, and the label belongs to either Democratic or Republican. We collect the largest connected component, including 1,222 nodes and 16,717 edges.
\item Cora \cite{sen2008collective}. The Cora graph has 2,708 nodes and 5,278 edges. The label of each node belongs to a set of 7 categories of computer science research areas, including Neural Networks, Rule Learning, Reinforcement Learning, Probabilistic Methods, Theory, Genetic Algorithms, Case Based.
\item Citeseer \cite{sen2008collective}. The Citeseer graph contains 3,312 nodes, including label sets (Agents, IR, DB, AI, HCI, ML). Since it contains many small connected components, we remain the largest connected component with 2,110 nodes and 3,668 edges  as the input graph.
\item PubMed \cite{namata2012query}. The PubMed graph includes 19,717 edges and 44,324 edges. Each node belongs to one of three types of diabetes.
\item MNIST \cite{rakhlin2017efficient}. We downloaded MNIST with background noise images from \url{https://sites.google.com/a/lisa.iro.umontreal.ca/public_static_twiki/variations-on-the-mnist-digits}, which includes 12,000 noisy images from digit 0 to 9. To create edges between these images, we first create a 10-nearest neighbors graph and then create the edge weights by using the averaged $\ell_2$ distance as suggested in \citet{cesa2013random}.
\item BlogCataglog, Flickr, and Youtube datasets are found in \cite{perozzi2014deepwalk}.
\item OGB-Arix and OGB-Products are from OGB dataset \cite{hu2020open}
\item Wikipedia. We download the raw corpus of English Wikipedia from \url{https://dumps.wikimedia.your.org/enwiki/20220820/} until the end of the year 2020. We create the inner-line edges for each Wikipedia article by checking the interlinks from these articles. Thus, we created the Wikipedia graph with 6,216,199 nodes and 177M edges. We then collect labels from \url{https://dbpedia.org/ontology/} where we can get 3,448,908 available node labels (only use first 150,000 nodes in our experiments) from ten categories including, Person, Place, Organisation, Work, MeanOfTransportation, Event, Species, Food, TimePeriod, Device.
\end{enumerate}

\subsection{Baseline Methods}
\begin{figure}
\centering
\includegraphics[width=.7\textwidth]{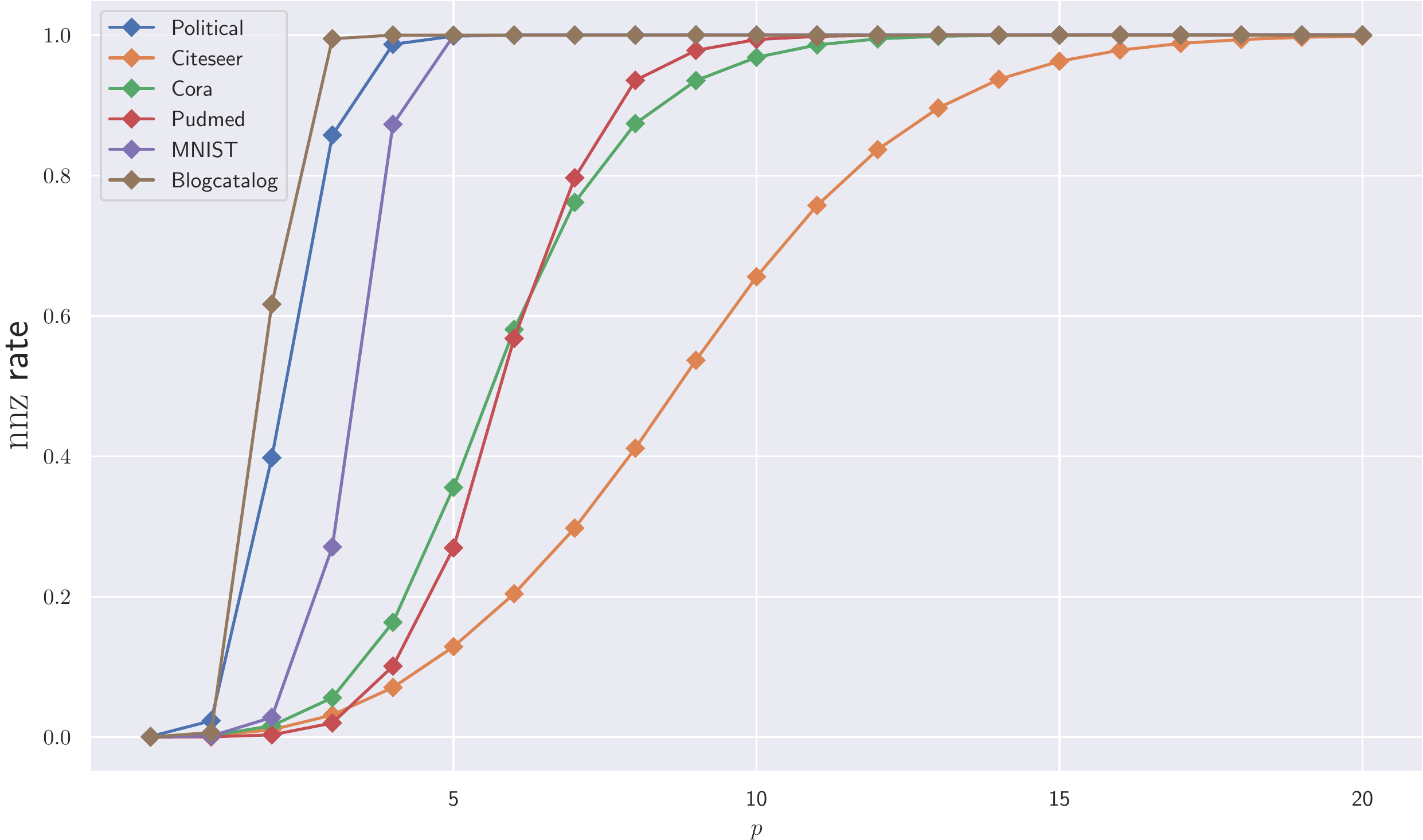}
\caption{The $\nnz$ rate as a function of iteration $p$ for \textsc{Approximate}. The $\nnz$ rate, i.e., the sparsity rate of the approximated matrix, is defined as $\nnz{(\bm M_p)} / n^2$ where $\nnz{(\bm M_p)}$ is the number of nonzero entries in $\bm M_p$ after $p$ iterations.  Most of these approximations become dense after 10 iterations.\vspace{-5mm}}
\label{fig:nnz-rate}
\end{figure}

\begin{figure}
    \centering
    \includegraphics[width=.9\textwidth]{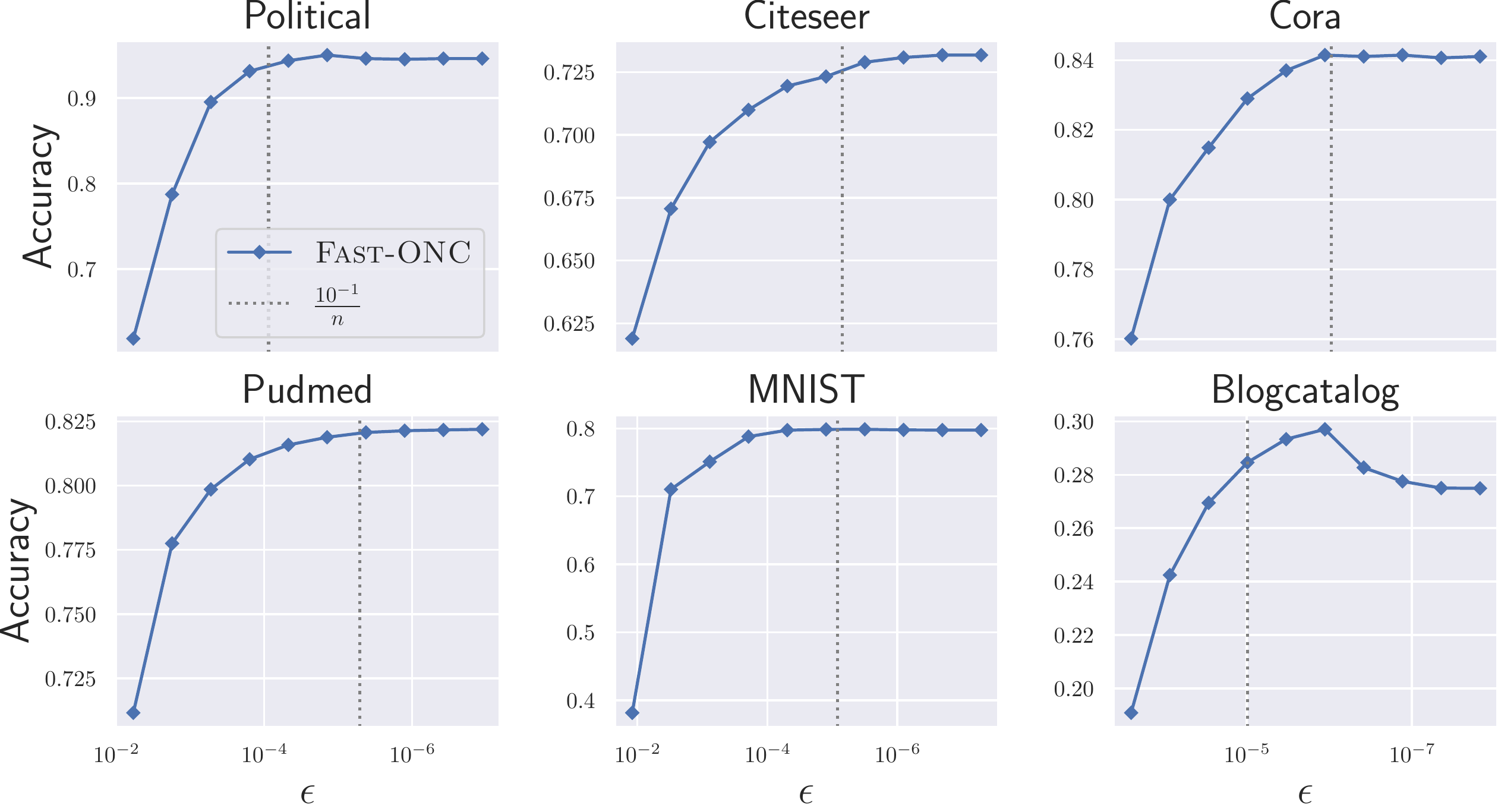}
    \caption{The overall node labeling accuracy as a function of $\eps$ over all six small graphs. The vertical line is when $\eps = \frac{10^{-1}}{n}$ corresponding to the $\eps$ we set in our experiments. This rough parameter estimation is good enough for most of the datasets.\vspace{-5mm}}
    \label{fig:diff-eps}
\end{figure}

We describe all four baseline methods and our method as follows
\begin{enumerate}

\item \textsc{Relaxation} \cite{rakhlin2017efficient}. It has a parameter $\lambda$. To have the best performance, we choose $\bm M_{\lambda,\beta} = 2n \bm D^{-1/2} \bm L \bm D^{1/2} $ as the underlying kernel matrix. In our experiments, we tune the parameter $\lambda$ from $\{0.1 * n,0.2*n,\ldots, 0.9*n\}$.

\item \textsc{Approximate} as defined in \eqref{inequ:approximate-method}. It has parameter $p=5$ in our small graph experiments. The reason that \textsc{Approximate} works well is that we use our \textsc{FastONL} framework to predict labels. In other words, the only difference from our method is that \textsc{Approximate} use \eqref{inequ:approximate-method} to obtain kernel vectors. In contrast, we use \textsc{FIFOPush} to obtain kernel vectors. Similarly, we find the second kernel works great. Hence, we choose the second kernel as the underlying kernel to approximate. We found memory issues when we apply this approximation method to middle-scale graphs. As illustrated in Fig. \ref{fig:nnz-rate}, the approximated matrix becomes dense when $p \geq 10$ for most of the small graphs. For example, the approximated solution for the Blogcatalog dataset becomes dense only after 3 iterations. This suggests that \textsc{Approximate} is unsuitable for large-scale networks.

\item \textsc{WM} is the weighted-majority method. We implemented it as follows: 1) If $u_t$ is the target node to be predicted, then the algorithm first finds all its available neighbors (nodes that have been seen in some previous iterations). Based on its neighbors, one can create a distribution of it. We just the maximal likelihood to choose the prediction label. We randomly select a label from the true label sets if no neighbors are found.
\item \textsc{WTA} is the weighted tree algorithm \cite{cesa2013random}. The essential idea of \textsc{WTA} is that the algorithm first constructs a random spanning tree (in our implementation, we choose to implement Wilson's algorithm described in \citet{wilson1996generating} to generate a random spanning tree. In expectation, the run time is linear to the number of nodes. One important parameter of \textsc{WTA}, we use $s=5$ for all small-scale graphs. Our Python implementation is too slow for middle-scale graphs to finish one graph without several hours. However, we still see a large gap between \textsc{WTA} and ours.
\item \textsc{FastONL} has two important parameter, including the label smoothness parameter $\lambda$ where we choose the same as did in \textsc{Relaxation}, and the precision $\eps$ to control the quality of kernel vectors. Except for the small-scale graphs. In all middle-scale graphs, we choose $\eps = .1/n$, which is good enough for node labeling tasks. To see this $\eps$ works well in general, we fix to use the second kernel and set $\lambda$ based on the best value shown in Fig. \ref{fig:parameter-tuning}. The results are shown in Fig. \ref{fig:diff-eps}. This precision setting is good enough for most of these small datasets. In Fig. \ref{fig:parameter-tuning}, we use the first 4 kernel matrices defined in Tab. \ref{tab:graph-kernel-presentation}. We name these kernels as $\bm K_1, \bm K_2, \bm K_3$, and $\bm K_4$, respectively. For $\bm K_3$ and $\bm K_4$, we directly use the defined $\beta$ for our experiments.  
\end{enumerate}

\paragraph{Run time comparison between \textsc{FastONL} and \textsc{Relaxation}.\quad} We conduct the experiment to compare these two methods. The table below shows the run time (in seconds) where the total time is to predict all $n$ nodes. Recall the size of the graph is between 
$n=1,222$ (Political) to $n=169,343$ (OGB-Arxiv). We fix 
 $\lambda = 0.15 n, \alpha = 1 - \frac{n}{n+\lambda}$, and use $\bm K_2$. The parameter $\eps = 0.1/n$. We perform experiments per data 10 times and take the average. As presented in Tab. \ref{tab:run-time-2} and \ref{tab:acc-2}, it is evident that \textsc{FastONL} is more efficient in most of datasets. Importantly, this increased efficiency does not come with a significant trade-off in accuracy.
 
\begin{center}
\begin{table}[H]
\centering
\caption{Run time of two methods in seconds}
\renewcommand{\arraystretch}{1.5}
\begin{tabular}{llllllll}
\toprule
& Political & Citeseer & Cora & Blogcatalog & MNIST & Pubmed & OGB-Arxiv \\\hline
 \textsc{Relaxation} & \textbf{0.1417} & 	1.0222 & 	1.5480 & 	87.8246	& 135.3061	& 585.3558 & \red{Out of Memory} \\\hline
\textsc{FastONL} &  0.1557	& \textbf{0.5883}	 & \textbf{1.0971}	 & \textbf{5.9510} &	\textbf{60.0465} & \textbf{50.7256} & \textbf{3199.6773} \\\hline
\bottomrule
\end{tabular}
\label{tab:run-time-2}
\end{table}    
\end{center}

\begin{center}
\begin{table}[H]
\centering
\caption{Accuracy of two methods}
\renewcommand{\arraystretch}{1.5}
\begin{tabular}{llllllll}
\toprule
& Political & Citeseer & Cora & Blogcatalog & MNIST & Pubmed & OGB-Arxiv \\\hline
 \textsc{Relaxation} & 0.9493 & 	0.7415 & 0.8404 & 0.2192 & 	0.7930	& 0.8254 & - \\\hline
\textsc{FastONL} &  0.9418 & 0.7404 & 	0.8420 & 	0.2921 & 	0.7960 & 	0.8257	& 0.7089  \\\hline
\bottomrule
\end{tabular}
\label{tab:acc-2}
\end{table}    
\end{center}

\begin{figure}
    \centering
    \includegraphics[width=.9\textwidth]{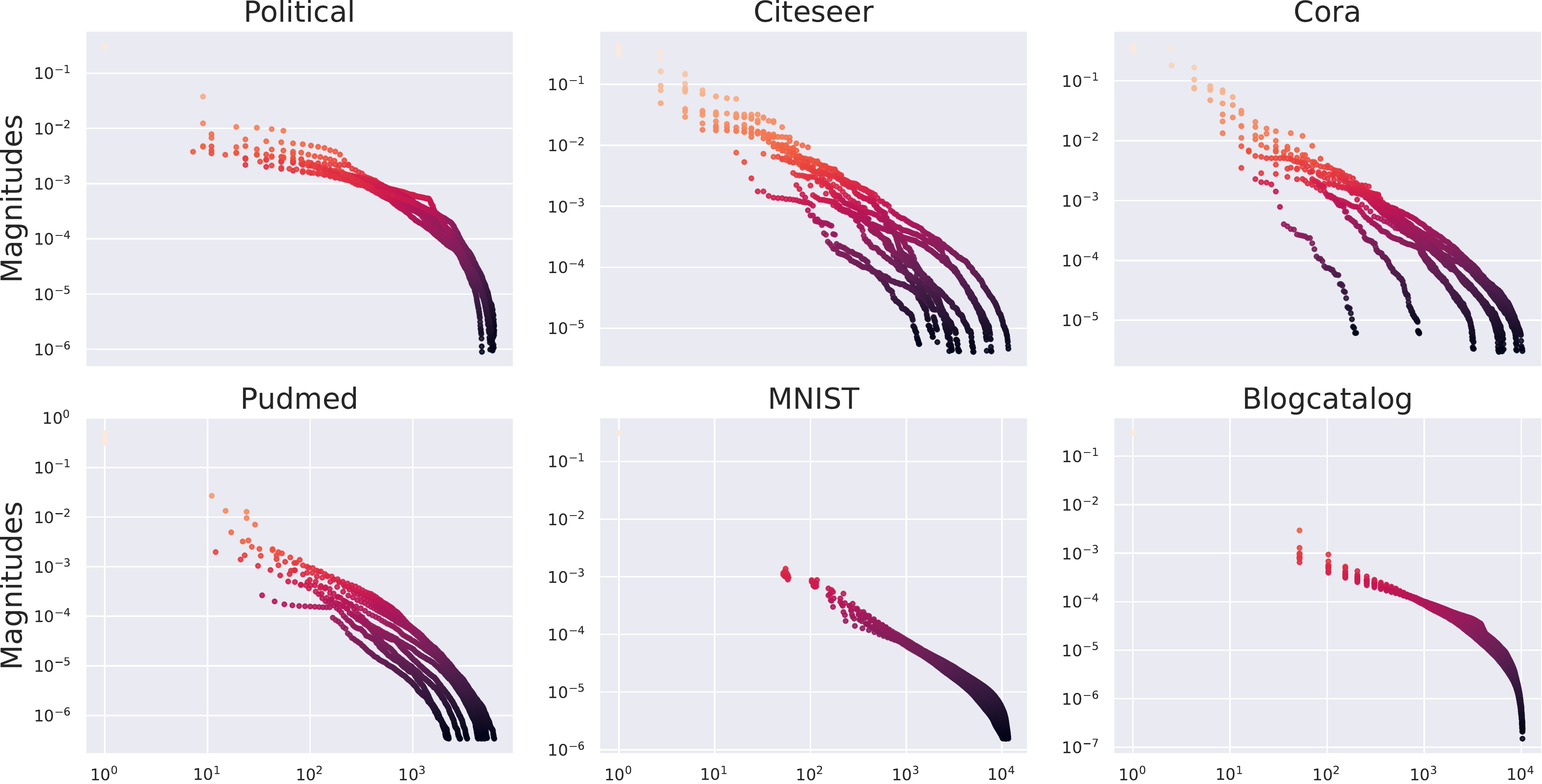}
    \caption{The power law distribution of magnitudes of $\bm X_{\bm L}$ on six small graphs. We set $\epsilon = 10^{-1}/m$ and fix $\alpha=0.2$. We then randomly select 20 nodes corresponding to 20 columns of $\bm X_{\bm L}$, sort the magnitudes of $\bm x_s$, and plot them according to their rankings. These magnitudes follow the power law distribution.}
    \label{fig:power-low-normalized}
\end{figure}

\begin{figure}
    \centering
    \includegraphics[width=.9\textwidth]{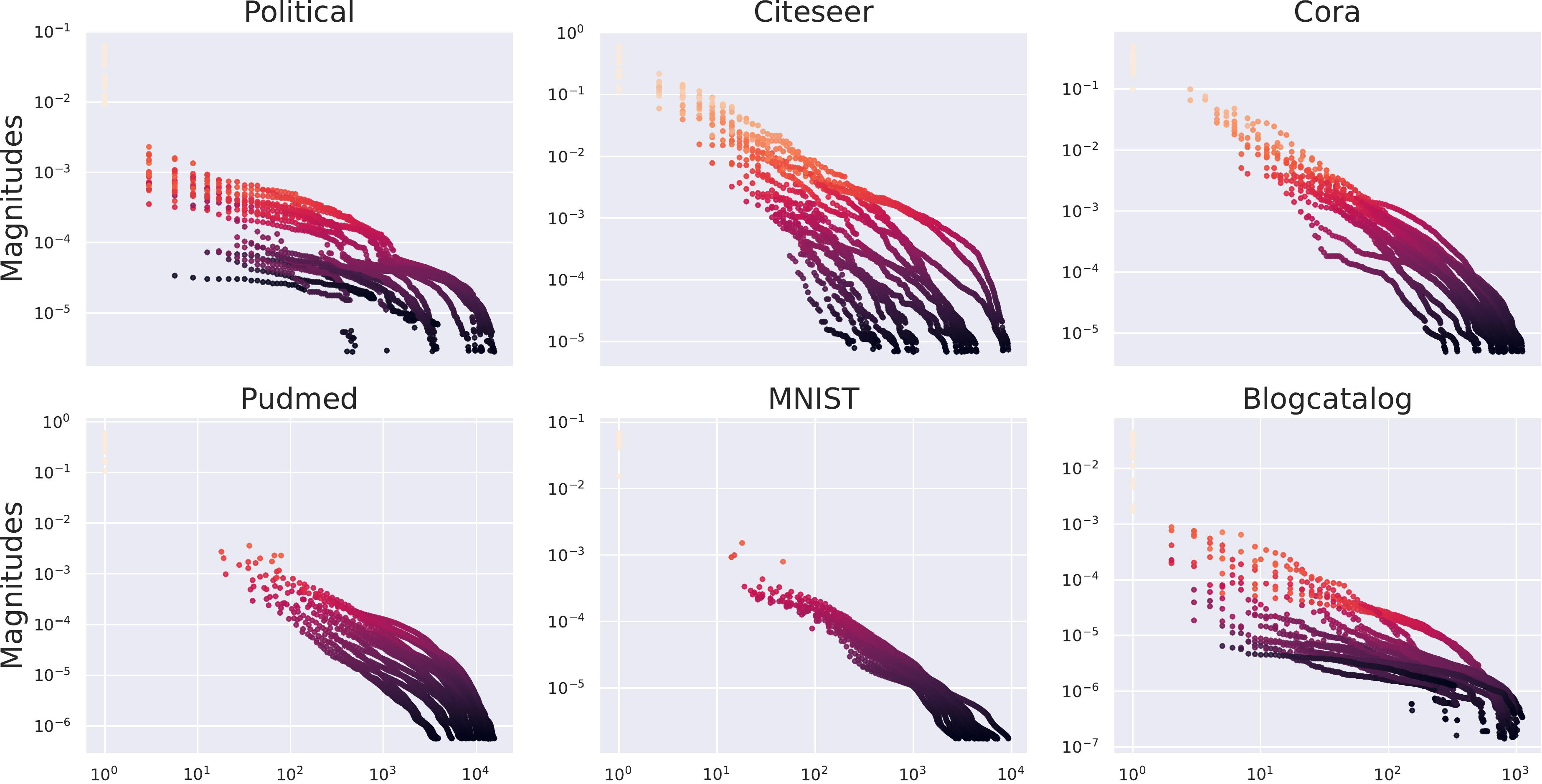}
    \caption{The power law distribution of magnitudes in $\bm X_{\mc L}$ on six small graphs. We set $\epsilon = 10^{-1}/m$ and fix $\alpha=0.2$. We then randomly select 20 nodes corresponding to 20 columns of $\bm X_{\mc L}$, sort the magnitudes of $\bm x_s$, and plot them according to their rankings. These magnitudes follow the power law distribution.}
    \label{fig:power-low-unnormalized}
\end{figure}

\paragraph{Power law of magnitudes of $\bm x_s$.\quad} We close up this section by showing the magnitudes of $\bm x_s$ of both $\bm X_{\mc L}$ and $\bm X_{\bm L}$ follow the power law distribution as we illustrate in Fig. \ref{fig:power-low-normalized} and \ref{fig:power-low-unnormalized}.

\begin{table}[H]
\centering
\renewcommand{\arraystretch}{1.8}
\begin{tabular}{ccccc} 
\toprule
Method & Regret & Per-Iteration & Total-Time & Memory \\
\hline WM & - & $\mathcal{O}\left(d_u\right)$ & $\mathcal{O}(m)$ & $\mathcal{O}(m)$ \\
\hline Perceptron & $\Phi G(\mathbf{y}) \cdot\left(\frac{n}{n^{\dagger} n^{-}}\right)^2$ & $\mathcal{O}(n)$ & $\mathcal{O}\left(n^3\right)$ & $\mathcal{O}\left(n^2\right)$ \\
\hline WTA & $\Phi G(\mathbf{y}) \cdot \log n$ & $\mathcal{O}(s)$ & $\mathcal{O}(s \cdot n \log n)$ & $\mathcal{O}(s \cdot n)$ \\
\hline RELAXATION & $\mathcal{O}(\sqrt{\operatorname{tr}(\mathbf{M})})$ & $\mathcal{O}(n)$ & $\mathcal{O}\left(n^3\right)$ & $\mathcal{O}\left(n^2\right)$ \\
\hline Ours & $\mathcal{O}(\sqrt{n^{1+\gamma}})$ & $\mathcal{O}\left(\frac{\mathcal{S}_T}{\alpha \cdot \eta T} \log ^{3 / 2} n\right)$ & $\mathcal{O}\left(\frac{n \cdot \mathcal{S}_T}{\alpha \cdot \eta T} \log ^{3 / 2} n\right)$ & $\mathcal{O}(\operatorname{supp}(\mathbf{x}) \cdot n)$\\
\bottomrule
\end{tabular}
\caption{Comparison of time and memory complexity between these online node labeling algorithms.}
\label{tab:complexity}
\end{table}

